\documentclass[preprint,12pt]{elsarticle}

\usepackage[Yipsnames]{xcolor}
\usepackage{array}
\usepackage{tabularx}
\usepackage{hyperref}
\usepackage{amsfonts} 
\usepackage{breqn}
\usepackage{bm}
\usepackage{multirow}
\usepackage{pdflscape}
\usepackage{float}
\usepackage{caption}
\usepackage{xcolor}
\usepackage{subfigure}
\definecolor{darkred}{rgb}{0.9,0.1,0.1}
\usepackage{longtable}
\usepackage[ruled,vlined]{algorithm2e}
\hypersetup{
 colorlinks,
 citecolor=black,
 filecolor=black,
 linkcolor=black,
 urlcolor=black
}
\usepackage{amssymb}

\usepackage[top=1.2in, bottom=1.4in, left=1.2in, right=1.2in]{geometry}

\journal{arXiv}

\biboptions{sort&compress}

\begin{document}

\begin{frontmatter}

\title{From PINNs to PIKANs:
Recent Advances in Physics-Informed Machine Learning }

\author[inst1,*]{Juan Diego Toscano}
\author[inst2,*]{Vivek Oommen}
\author[inst2,*]{Alan John Varghese}
\author[inst1]{Zongren Zou}
\author[inst2,inst3]{Nazanin Ahmadi Daryakenari}
\author[inst2]{Chenxi Wu}
\author[inst1,inst2,label2]{George Em Karniadakis}
\affiliation[inst1]{organization={Division of Applied Mathematics, Brown University},
  city={Providence},
  postcode={02912}, 
  state={RI},
  country={USA}}
\affiliation[inst2]{organization={School of Engineering, Brown University},
  city={Providence},
  postcode={02912}, 
  state={RI},
  country={USA}}
  
\affiliation[inst3]{organization={Center for Biomedical Engineering, Brown University},
  city={Providence},
  postcode={02912}, 
  state={RI},
  country={USA}}

\fntext[*]{Co-First authors}
\fntext[label2]{Corresponding author: george\_karniadakis@brown.edu}

\begin{abstract}

Physics-Informed Neural Networks (PINNs) have emerged as a key tool in Scientific Machine Learning since their introduction in 2017, enabling the efficient solution of ordinary and partial differential equations using sparse measurements. Over the past few years, significant advancements have been made in the training and optimization of PINNs, covering aspects such as network architectures, adaptive refinement, domain decomposition, and the use of adaptive weights and activation functions. A notable recent development is the Physics-Informed Kolmogorov–Arnold Networks (PIKANs), which leverage a representation model originally proposed by Kolmogorov in 1957, offering a promising alternative to traditional PINNs. In this review, we provide a comprehensive overview of the latest advancements in PINNs, focusing on improvements in network design, feature expansion, optimization techniques, uncertainty quantification, and theoretical insights. We also survey key applications across a range of fields, including biomedicine, fluid and solid mechanics, geophysics, dynamical systems, heat transfer, chemical engineering, and beyond. Finally, we review computational frameworks and software tools developed by both academia and industry to support PINN research and applications.
\end{abstract}



\end{frontmatter}

\tableofcontents

\section{Introduction}

The finite element method (FEM) has been the cornerstone of Computational Science and Engineering (CSE) in the last few decades but it was viewed with skepticism when the first published works appeared in the early 1960s. Despite their success in academic research and industrial applications, FEM cannot easily assimilate measured data unless elaborate data assimilation methods are employed that render large-scale computations prohibitively expensive. FEM and other conventional numerical methods are effective in solving well-posed problem with full knowledge of the boundary and initial conditions as well as all material parameters. Unfortunately, in practical applications, there are always gaps in such a setting and arbitrary assumptions have to be made, e.g. to assume the thermal boundary conditions at the walls in power electronics cooling applications. This may lead to erroneous results as in such a problem of interest is the highest temperature or the highest heat flux that is typically located at the surface where erroneous assumptions are employed. What may be available instead are a few
sparse thermocouple measurements either on the surface or inside the domain of interest. Unfortunately, current numerical methods like FEM cannot utilize such measurements effectively and hence important experimental information for the system is lost.
On the other hand, neural networks are trained based on data of any fidelity or any modality so data assimilation is a natural process in such settings.

Physics-Informed Neural Networks (PINNs) were developed to address precisely this need, considering different simulation scenarios where there is some knowledge of the governing physical laws but not complete knowledge, and there exist some sparse measurements for some of the state variables but not for all.
Hence, PINNs provide a framework to encode physical laws in neural networks~\citep{raissi2019physics} and resolve the disconnect between traditional physically grounded mathematical models and modern purely data-driven methods. Specifically, PINNs incorporate the governing laws by having an additional `residual' loss term in the objective function that enforces the underlying PDE as a soft constraint. They are effective in solving both forward and inverse problems across all scientific domains. PINNs can incorporate sparse and noisy data, making them effective in scenarios where acquiring accurate measurements can be difficult or expensive.
A key innovation in PINNs is the use of automatic differentiation based on computational graphs that leads to accurate treatment of the differential operators employed in conservation laws but most importantly removes the tyranny of elaborate mesh generation that is time consuming and limits solution accuracy.



Since the original two papers appeared on the arXiv in 2017~\citep{raissi2017physicsI, raissi2017physicsII} and the subsequent publication of a combined paper in 2019~\citep{raissi2019physics}, there has been  great excitement in the CSE community and very important advances on many  aspects of the method have been proposed by research groups from around the world and across all scientific domains. At the time of this writing, there have already been  over 11000 citations of~\citep{raissi2019physics} , with many studies investigating the applicability of PINNs  across different scientific domains while other studies proposing algorithmic improvements aimed at addressing the limitations of the original formulation. In the current review paper, we provide a compilation of most of the major algorithmic developments and present a non-exhaustive list of applications of PINNs across different disciplines. A comprehensive timeline of some of the important papers about PINNs is presented in the Appendix from PINNs~\citep{raissi2017physicsI} to PIKANs~\citep{liu2024kan}.

While existing reviews, such as those by~\citep{cuomo2022scientific, farea2024understanding, ganga2024exploringphysicsinformedneuralnetworks, raissi2024physics} summarize key aspects of PINNs, our paper differentiates itself by providing a more extensive overview of the latest algorithmic developments and by covering a broader range of applications of PINNs across scientific disciplines. Reviews by~\citep{cuomo2022scientific} and~\citep{farea2024understanding} focus primarily on the methodology and applications of PINNs in various domains, with less emphasis on recent algorithmic improvements. The review by~\citep{raissi2024physics} provides a concise overview of PINNs and their extensions, with an example on data-driven discovery of equations, but does not dive deep into applications of PINNs. The review in~\citep{ganga2024exploringphysicsinformedneuralnetworks} includes a discussion of algorithmic developments, but limits the scope of their discussion on applications to thermal management and computational fluid dynamics. Additionally, several reviews focus on specific domains of application. For example,~\citep{chi2024comprehensive} and~\citep{cai2021physics} review the use of PINNs in fluid dynamics, while~\citep{huang2022applications} focuses on applications within power systems. In contrast,~\citep{lawal2022physics} conducted a bibliometric analysis of 120 research articles, highlighting key publication trends, highly cited authors and leading countries in PINN research.


\begin{figure}[!ht]
    \centering
    \includegraphics[width= \linewidth]{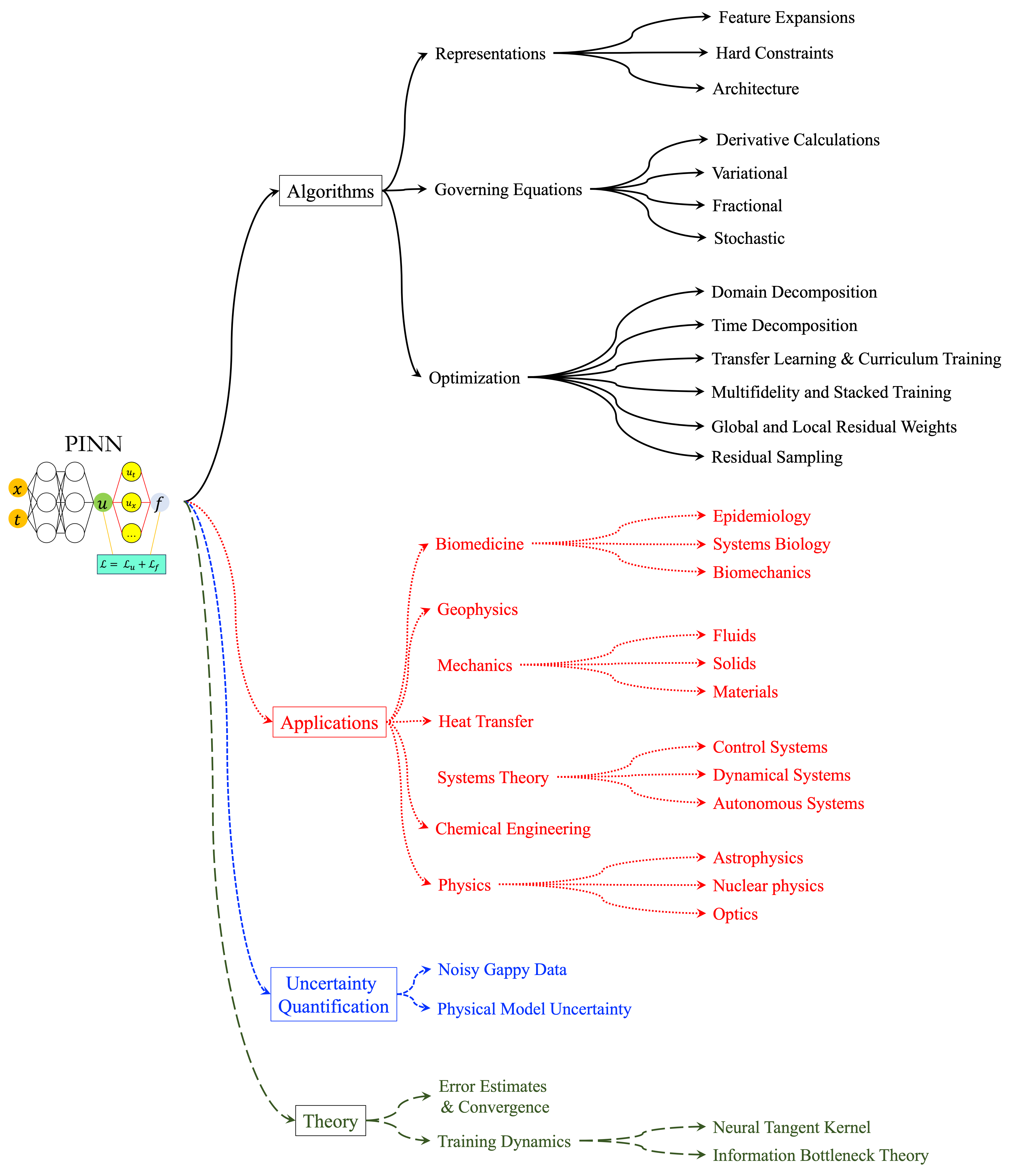}
    \caption{\textbf{Overview of the paper:} Taxonomy of algorithmic developments (Section \ref{algorithmic_developments}), applications (Section \ref{applications}), uncertainty quantification (Section \ref{uncertainty_quantification}), and theory of PINNs(Section \ref{theoretical_advances}) is illustrated here.  }
    \label{fig:schematic}
\end{figure}

The structure of the paper is shown schematically in Fig. \ref{fig:schematic}. 
In Section \ref{framework} we outline the general framework of Physics-Informed Machine Learning. Section \ref{algorithmic_developments} provides a comprehensive summary of the major techniques aimed at improving PINNs. In Section \ref{applications} we provide an overview of the diverse applications of PINNs. Section \ref{uncertainty_quantification} focuses on uncertainty quantification methods in PINNs. In Section \ref{theoretical_advances}, we summarize the developments in the theory behind PINNs. Section \ref{computational_frameworks} reviews the various computational frameworks and software. Finally, in Section \ref{discussion}, we provide a discussion and future outlook.

\section{Physics-Informed Machine Learning (PIML)}
\label{framework}

Physics-Informed Machine Learning (PIML) has emerged as a powerful alternative to traditional numerical methods for solving partial differential equations (PDEs) in both forward and inverse problems.
PIML was first introduced in a series of papers by Raissi, Perdikaris, and Karniadakis~\citep{raissi2017machine} based on Gaussian processes regression (GPR); see also the patent by the same authors~\citep{raissi2021physics}. In this paper, however, we will review the subsequent development of PIML using neural networks and automatic differentiation, starting with the two papers from 2017 on the arXiv~\citep{raissi2017physicsI, raissi2017physicsII}, which 
were combined into a single paper later in~\citep{raissi2019physics}.
It is worth noting that earlier papers by~\citep{dissanayake1994neural,lagaris1998artificial} attempted to solve PDEs (forward problems) but without any data fusion or automatic differentiation. The PIML we present in this paper employs a representation model, namely a multilayer perceptron (MLP) or a Kolmogorov-Arnold Network (KAN)~\citep{liu2024kan}, to approximate the solution of ordinary or partial differential equations (ODEs/PDEs) and match any given data and constraints by minimizing a loss function comprised of multiple terms. In particular, this loss function is designed to fit observable data or other physical or mathematical constraints while enforcing the underlying physics, e.g., conservation laws~\citep{raissi2019physics,shukla2024comprehensive}.

Unlike traditional numerical methods, most PIML models do not rely on predefined grids or meshes, allowing them to handle complex geometries and high-dimensional problems efficiently. By leveraging automatic differentiation, PIML models compute derivatives accurately without discretization, seamlessly integrating governing physical laws with data. This flexibility allows PIML models to approximate solutions from partial information, making them optimal for uncovering hidden parameters~\citep{raissi2019physics}, as well as reconstructing~\citep{cai2021flow} or inferring hidden fields~\citep{raissi2020hidden} from real-world data. Moreover, PIML models are well-suited for handling high-dimensional PDEs~\citep{hu2024tackling}, coupled systems~\citep{jin2021nsfnets, shukla2024neurosem}, stochastic differential equations~\citep{yang2019adversarial}, and fractional PDEs~\citep{pang2019fpinns}, all while maintaining scalability through parallelization on modern hardware such as GPUs~\citep{karniadakis2021physics}. This enables PIML models to efficiently tackle multi-physics problems and large-scale simulations with reduced computational overhead compared to traditional methods.


PIML is agnostic to specific governing laws, so here we consider the following nonlinear ODE/PDE:
\begin{subequations}\label{eq:problem}
  \begin{align}
    \mathcal{F}_\tau[\hat{u}](x)  &= f(x), \quad x \in \Omega, \\
    \mathcal{B}_\tau[\hat{u}](x)  &= b(x), \quad x \in \Omega_B,
  \end{align}
\end{subequations}
where \( x \) represents the spatial-temporal coordinate, \( \hat{u} \) is the solution to the ODE/PDE, \( \tau \) are the parameters of the equation, \( f \) is the source term, \( b \) is the boundary term, and \( \mathcal{F} \) and \( \mathcal{B} \) are general nonlinear differential and boundary operators, respectively. The PIML approach aims to approximate the solution to the problem defined by Eq. \eqref{eq:problem} using a representation model, denoted as:
\begin{align}
  \hat{u}(x) \approx u(\theta, x), x\in\Omega\cup\Omega_B,
\end{align}
where \( u \) is the representation model, and \( \theta \) are its learnable parameters. Since \( u \) is continuous and differentiable, it allows for the computation of the source and boundary terms \( f \) and \( b \) through automatic differentiation~\citep{baydin2018automatic}, expressed as \( \mathcal{F}_\tau[u] \) and \( \mathcal{B}_\tau[u] \)~\citep{raissi2019physics}.

The goal of PIML training is to find the optimal learnable parameters that minimize the cumulative error between the approximated solution and the known components of the true solution, such as the governing equation, boundary conditions, or data residuals. This framework can also be easily extended to ODE/PDE systems by stacking constraints for each approximated solution~\citep{raissi2020hidden}.

In general, when the equation parameters \( \tau \) are known and the boundary conditions are prescribed, the problem is referred to as a \textit{forward problem}, where no observational data within the domain are required~\citep{raissi2019physics, meng2020ppinn}. Conversely, when partial information, such as \( \tau \), boundary conditions, or hidden fields in the ODE/PDE system, is unknown, then the problem is referred to as an \textit{inverse problem}, where the objective is to infer simultaneously the unknown information and the solution from available data or observations~\citep{raissi2020hidden}.
A schematic of the overall PIML framework is shown in Fig. \ref{fig:PIML_compz}.

\section{Algorithmic Developments of PIML}
\label{algorithmic_developments}
\begin{figure}
    \centering
    \includegraphics[width=0.75\linewidth]{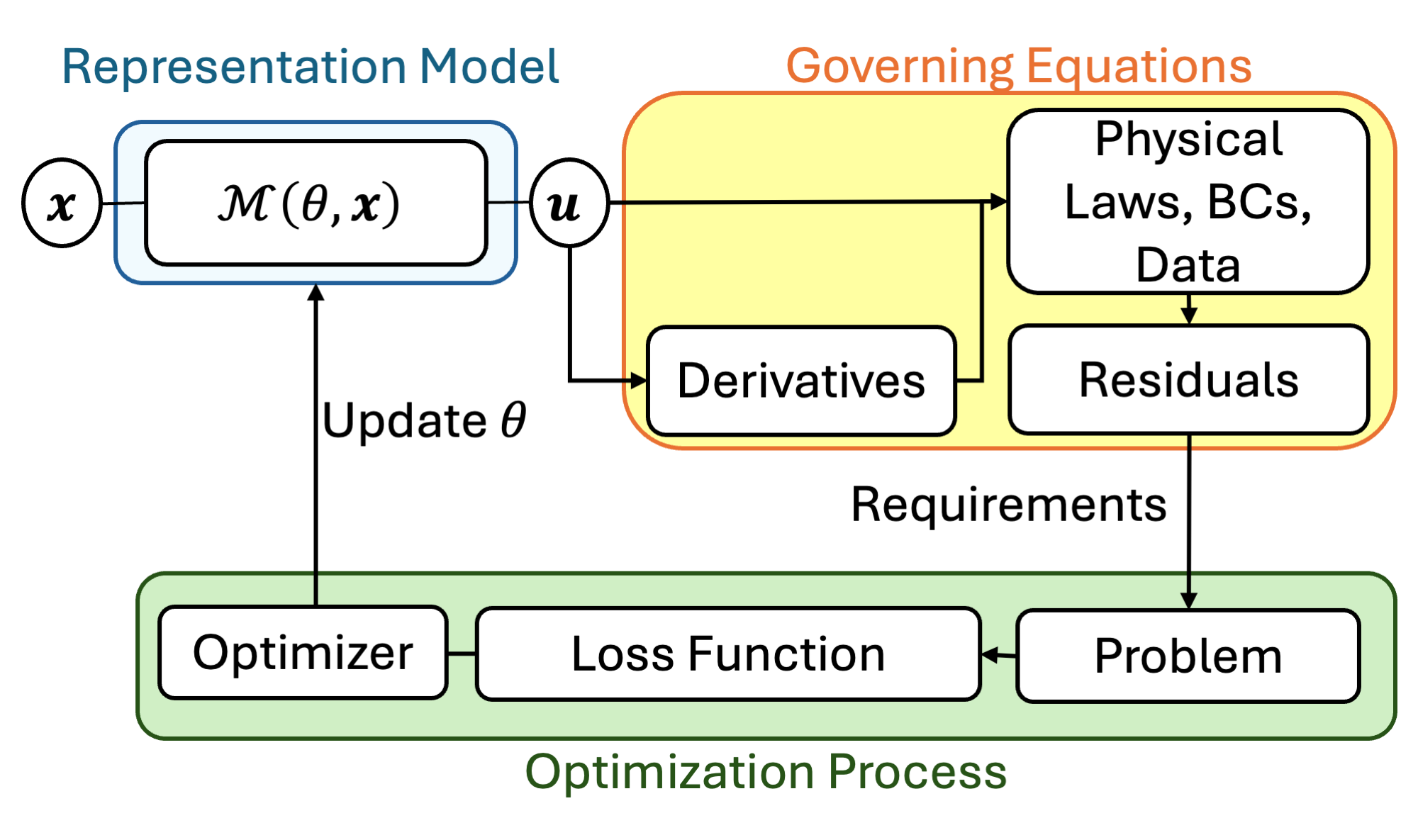}
\caption{\textbf{PIML Training components}. A representation model provides an approximation \( u \) of the ODE/PDE solution \( \hat{u} \). The mismatch (residuals) between the approximated solution and the known components of the true solution, such as physical laws and boundary conditions, is calculated. These governing equations define a set of requirements that generate an optimization problem. This problem is reformulated as a loss function and then sent to an optimizer that updates the model parameters.}

    \label{fig:PIML_compz}
\end{figure}
From the PIML framework outlined in Section~\ref{framework}, we can identify three key components: (1) a representation model to approximate the solution, (2) a governing equation (such as an ODE or PDE), and (3) an optimization process that minimizes a multi-objective loss function to find the optimal learnable parameters,
see Fig. \ref{fig:PIML_compz}. Ongoing research has greatly enhanced PIML's baseline performance through various methods targeting these three areas, namely, modifications to the representation model, advancements in the treatment of the governing equation, and optimization process improvements. 

\subsection{Representation Model Modifications}
\label{Rep_enhancemetns}

When the representation model is defined using an MLP, the PIML formulation is referred to as physics-informed neural networks (PINNs) ~\citep{raissi2019physics}. PINNs utilize MLPs to approximate the solutions of an ODE/PDE system ($\bm{\hat{u}} = \{\hat{u}_1, \ldots, \hat{u}_p\}$)  by leveraging the network's ability to model complex nonlinear functions. The approximated solution ($\bm{{u}} = \{{u}_1, \ldots, {u}_p\}$)  using MLPs can be mathematically expressed as:
\begin{align}
  \bm{u}(\theta, \bm{x})  &= \text{MLP}(\theta, \bm{x})  \\
  &= \sigma\left(W^{(L) } \sigma\left(W^{(L-1) } \ldots \sigma\left(W^{(1) } \bm{x} + b^{(1) }\right)  \ldots + b^{(L-1) }\right)  + b^{(L) }\right) 
  \label{MLP}
\end{align}
\noindent where ($\bm{x}= \{x_1, \ldots, x_n\}) $)  are inputs and $\theta = \{W^{(l) }, b^{(l) }\}$ are the trainable parameters of the network; $W^{(l) }$ and $b^{(l) }$ denote the weights and biases of the $l$-th layer, respectively. The network consists of $L$ layers, and $\sigma$ denotes a suitable activation function. The output of each layer serves as the input for the subsequent layer, culminating in the final output \( u \).

The ability of MLPs to approximate virtually any continuous function on compact subsets of \( \mathbb{R}^n \)  is supported by the Universal Approximation Theorem~\citep{hornik1989multilayer}. This theorem underpins the effectiveness of neural networks in modeling complex nonlinear relationships.

Building on the foundational work in~\citep{raissi2019physics}, many studies have explored enhancing the expressiveness of representation models in PINNs through various strategies. These include input and output normalization, feature expansions, hard constraint encoding, model decompositions, and architectural modifications. Each of these strategies aims to improve the network's ability to capture the underlying physics of the problem more accurately and efficiently.
\begin{figure}[!ht]
    \centering
    \includegraphics[width=\linewidth]{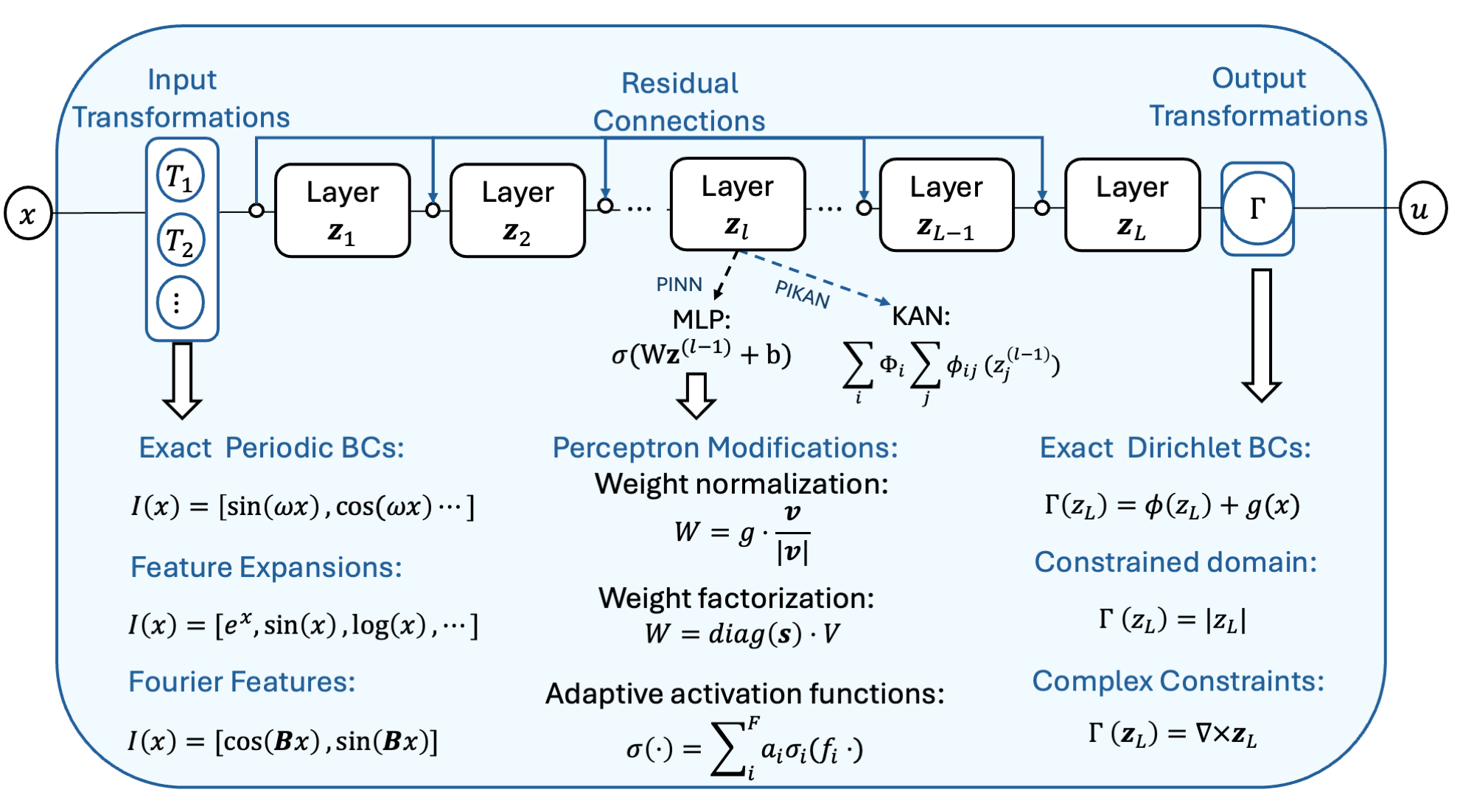}
\caption{\textbf{Representation Model Enhancements}: Input transformations improve the model's ability to impose periodic boundary conditions~\citep{dong2021method}, enhance expression capabilities~\citep{guan2023dimension}, and mitigate spectral bias through feature expansions~\citep{wang2021eigenvector}. Residual connections boost performance and accuracy for high-order derivatives required to enforce the PDE~\citep{wang2024piratenets}. The model architecture typically consists of layers, where each can be a perceptron (PINNs) or a KAN~\citep{liu2024kan} (PIKAN~\citep{shukla2024comprehensive}). PINN performance is further enhanced by modifications like weight normalization~\citep{salimans2016weight, wang2022random} or adaptive activation functions~\citep{jagtap2022deep}. Output transformations enforce solution constraints, such as Dirichlet boundary conditions~\citep{sukumar2022exact} or divergence-free constraints~\citep{wang2021understanding}.}

    \label{RM_enh}
\end{figure}

\subsubsection{Input/Output Transformations}

One of the most straightforward ways to improve the stability and accuracy of a representation model $\mathcal{M}$ is by transforming the model inputs $\bm{x} \in \mathbb{R}^n$ or outputs $\bm{u} \in \mathbb{R}^p$ using suitable mappings, $I(\cdot) $and $\Gamma(\cdot) $ (see Fig.~\ref{RM_enh}). Under this reformulation, the solutions of an ODE/PDE ($\bm{\hat{u}}$)  can be approximated as:

\begin{align}
\hat{\bm{u}}(\bm{x})  \approx \bm{u}(\theta, \bm{x})  = \Gamma(\mathcal{M}(\theta, I(\bm{x}) ) ), 
\label{expanded_MLP}
\end{align}

\noindent where $\bm{u} = \{u_1, \ldots, u_p\}) $ are the approximated solutions from the representation model $\mathcal{M}$, and  $I: \mathbb{R}^n \rightarrow \mathbb{R}^m$ is a mapping that transforms the multivariate inputs and enhances the model’s expressivity. The choice of $I$ often depends on the activation function $\sigma$, as different activation functions are more effective over specific input domains. For example, Cai et al.~\citep{cai2021artificial} proposed normalizing inputs to the range $[-1, 1]$ when using $\sin(\cdot)$ or $\tanh(\cdot)$ activation functions. Similarly, Raissi et al.~\citep{raissi2020hidden} recommended normalizing inputs based on their mean and standard deviation when using the $swish(\cdot)$ activation function.

On the other hand, the function $\Gamma: \mathbb{R}^p \rightarrow \mathbb{R}^q$ is selected based on the specific characteristics of the problem. For instance, Anagnostopoulos et al.~\citep{anagnostopoulos2024residual} used scalars $a_1, \ldots, a_p \in \mathbb{R}$ to create different linear maps to handle various scales in the model outputs, represented as $\bm{u} = \{a_1u_1, \ldots, a_pu_p\}$, which aids in improving convergence. Similarly, $\Gamma$ can be used to constrain the range of predicted outputs. For example, absolute value $\Gamma(u) =|u|$ or exponential $\Gamma(u) =e^{u}$ functions can enforce the non-negativity required for the density field in high-speed flows~\citep {mao2020physics}. Likewise, Zapf et al.~\citep{zapf2022investigating} used the sigmoid function to constrain the range of the predicted diffusivity.

\paragraph{Feature Expansions}

 These modifications aim to address some fundamental weaknesses in the conventional PIML formulation, particularly in learning high-frequency functions, known as spectral bias~\citep{rahaman2019spectral,cao2019towards}, and capturing other complex relations~\citep{wang2021eigenvector}. To mitigate these issues, researchers have modified the baseline model formulation by transforming its input $\bm{x} \in \mathbb{R}^n$ into an expanded input $I(\bm{x})  \in \mathbb{R}^m$, which is then fed into the representation model as shown in Fig.~\ref{RM_enh}. Several expansion maps have been explored, with their selection typically based on the specific problem. For example, Wang et al.~\citep{wang2021eigenvector} proposed using random Fourier features and demonstrated that this modification helps to mitigate spectral bias. Other types of expansions, such as polynomials, exponentials, Chebyshev polynomials, and even gated recurrent units (GRU), have also been explored or proposed in previous studies~\citep{cai2019multi,liu2020multi,wang2020multi,liu2022linearized,ahmadi2024ai,zhang4957859pkan}.

\paragraph{Hard Constraints}

Training in PIML involves minimizing an objective function that often combines multiple constraints, significantly complicating the optimization process~\citep{toscano2024inferring}. Research has shown that improper enforcement of boundary conditions can degrade both the performance and stability of neural network training~\citep{dong2021method,wang2021understanding,chen2020comparison}. Thus, accurately imposing boundary conditions is crucial for improving model reliability and efficiency. In particular, Zeinhofer et al.~\citep{zeinhofer2023unified} theoretically demonstrated that hard constraints can lead to lower error estimates in linear problems. One practical way to address these challenges is by embedding boundary conditions directly into the model's structure, either through input/output transformations or specialized architectures, thereby simplifying the optimization and enhancing the overall performance.

\subparagraph{Dirichlet Boundary Conditions.} Several methods have been developed to enforce Dirichlet boundary conditions exactly in PIML problems. Berrone et al.~\citep{berrone2023enforcing} introduced the Nitsche's method, which applies a variational approach to enforce these conditions. Another systematic technique is the Theory of Functional Connections, which imposes constraints through functional connections, as detailed by Leake et al.~\citep{leake2020deep}. On the other hand, hPINNs utilize penalty methods and the augmented Lagrangian approach to impose hard constraints, providing a flexible framework for handling various boundary conditions~\citep{lu2021physics}. Notably, Sukumar et al.~\citep{sukumar2022exact} introduced Approximate Distance Functions (ADF), which impose boundary conditions through output transformations. Under the ADF framework, the constrained expression for Dirichlet boundary conditions is represented as:
\begin{align}
\hat{\bm{u}}(\bm{x})  \approx \bm{u}(\theta, \bm{x})  = \Gamma(\mathcal{M}(\theta, I(\bm{x}) ) ) =\bm{g}(\bm{x})  + \bm{\phi}(\bm{x}) \mathcal{M}(\theta, I(\bm{x}) ), 
\label{ADF}
\end{align}
\noindent where $\Gamma(\cdot) $ transform the network output in terms of a function that satisfies the solution $\hat{\bm{u}}$ along the boundaries $\bm{g}(\cdot) $  and a composite distance function $\bm{\phi}(\cdot) $ that equals zero when evaluated on the boundaries. If the boundary is composed of $M$ partitions, denoted as $[S_1, \ldots, S_M]$, the composite distance function for Dirichlet boundary conditions can be expressed as $\bm{\phi}(\phi_1, \phi_2, \ldots, \phi_M)  = \prod_{i=1}^M \phi_i$, where $[\phi_1, \ldots, \phi_M]$ are the individual distance functions. Notice that if $\bm{x} \in S_i$, then $\phi(\bm{x})  = 0$, ensuring that the neural network approximation exactly satisfies the boundary conditions, i.e., $\bm{u}(\bm{x})  = \bm{g}(\mathbf{x}) $.

\subparagraph{Periodic Boundary Conditions.} On the other hand, this type of boundary conditions can be strictly enforced as hard constraints by selecting a suitable input transformation \(I(\bm{x}) \)  (see Fig.~\ref{RM_enh}). For instance, the periodic nature of a smooth univariate function \(u(x) \)  can be encoded into a model using a one-dimensional Fourier feature embedding, $I(x)  = [1, \cos(\omega_x x), \sin(\omega_x x), \ldots, \cos(m\omega_x x), \sin(m\omega_x x) ]$. Dong et al.~\citep{dong2021method} demonstrated that any representation model, such as \(u(\theta, I(x) ) \), is periodic along the \(x\)  coordinate when using this Fourier feature embedding. Similar expansion maps have been explored for higher dimensions by Wang et al.~\citep{wang2022respecting}. 

Other methods to impose hard constraints for periodic boundary conditions include using hPINNs, which employ the penalty and augmented Lagrangian methods to enforce such constraints~\cite {lu2021physics}. Additionally, hybrid approaches that combine various techniques for the exact imposition of periodic boundary conditions have been investigated~\cite {barschkis2023exact}.

\subparagraph{Complex Constraints.} Complex constraints can be addressed through specialized architectures that incorporate domain-specific knowledge into the learning process. One such approach is the use of Divergence-Free Networks, which apply an appropriate output transformation to ensure that the learned vector fields satisfy divergence-free conditions (see Fig.\ref{RM_enh}), as required in certain fluid dynamics applications\citep{toscano2024inferring,wang2021understanding,anagnostopoulos2024learning}. Zeinhofer et al.\citep{zeinhofer2023unified} theoretically demonstrated that enforcing divergence-free constraints leads to improved error estimates in linear problems. Another example is SympNets, a specialized architecture for identifying Hamiltonian systems from data, which preserves the symplectic structure through different Jacobian matrix factorization techniques\citep{jin2020sympnets}. Finally \citep{zhou2023flow} proposed a method that uses the theory of functional connections to exactly enforce the data constraints in inverse problems.

\subsubsection{Architectures}

\paragraph{Perceptron Modifications}
The original PIML formulation uses an MLP as the representation model~\citep{raissi2019physics}. The building blocks of an MLP are perceptrons, which define a layer $l$ and can be defined as follows: 

\begin{align} \bm{z}^{(l) }&=\sigma^{(l) }(W^{(l) }\bm{z}^{(l-1) } + b^{(l) }) , \label{perceptron} \end{align} 

\noindent where $\bm{z}^{(l) }=\{z_{0}^{(l) },\cdots,z_{H}^{(l) }\}$ are the outputs of layer $l$, $\sigma^{(l) }$ is a non-linear activation function (e.g., $\text{sigmoid}(\cdot) $, $\tanh(\cdot) $, $\sin(\cdot) $, etc.) , and $W^{(l) }$ and $b^{(l) }$ are trainable parameters that linearly transform the inputs $\bm{z}^{(l-1) }=\{z_{0}^{(l-1) },\ldots,z_{H}^{(l-1) }\}$. Several approaches have been proposed to improve the perceptron’s capabilities. For instance, Jagtap et al.~\citep{jagtap2020adaptive,jagtap2020locally,jagtap2022deep} proposed modifying the activation function in as
$\sigma^{(l) }(\cdot)  = \sum_i a_{i}^{(l) } \sigma(f_{i}^{(l) } \cdot) $, where $a_{i}^{(l) }$ and $f_{i}^{(l) }$ are trainable parameters. The authors showed both theoretically and empirically that these modifications significantly improve model performance. 

Other approaches proposed modifying the weight matrix $W^{(l) }$; for instance,~\citep{salimans2016weight,raissi2020hidden} proposed decomposing $W^{(l) }$ into its magnitude and its direction via weight normalization described as $W^{(l) }=\frac{g^{(l) }}{\lVert\mathbf{v}^{(l) }\rVert_{2}}\mathbf{v}^{(l) }$. This re-parameterization speeds up the model convergence with minimal computational overhead~\citep{salimans2016weight}. Similarly, ~\citep{wang2022random}  proposed  weight factorization $W^{(l) }=diag (\bm{s} ^{(l) })\cdot V^{(l) }$, where $\bm{s}^{(l) }$ are trainable parameters. The authors experimentally and theoretically showed that this reparameterization significantly improves the model performance~\citep{wang2022random}.

\paragraph{Other Representation Models} One natural extension to other representation models, given their similarity to MLPs, is Kolmogorov-Arnold Networks (KANs)~\citep{liu2024kan}, which were introduced as PIKANs in~\citep{shukla2024comprehensive}. Each PIKAN layer can be described as follows:

\begin{equation} \bm{z}^{(l)} = \sum_{i=1}^{H}\Phi_i\left(\sum_{j=1}^{H}\phi_{i,j}(z^{(l-1)}_{j})\right),\label{KAN_net} \end{equation}

\noindent where $\bm{z}^{(l-1) }=\{z_{0}^{(l-1) },\cdots,z_{H}^{(l-1) }\}$ is the multivariate input, $H$ denotes the number of neurons and $\Phi_{i,j}$ are the outer and $\phi_{i,j}$ are the inner univariate functions. The specific form of $\phi(\cdot)$ and $\Phi(\cdot)$ defines the types of KAN architectures. Among these variations,~\citep{shukla2024comprehensive} introduced cPIKANs, which use Chebyshev polynomials as inner and outer univariate functions defined as $ \phi(\zeta,\theta) =w_n\sum_n^d c_nT_n(\tanh(\zeta)) $, where $\zeta$ are the inputs, $\theta=(w_n,c_n) $ are trainable parameters, $d$ is the degree and $T_n$ is the $n$-th order Chebyshev polynomial, defined recursively as $T_n(\zeta) =2\zeta T_{n-1}(\zeta)  + T_{n-2}(\zeta) $~\citep{karniadakis2005spectral}. The authors found that this stable representation is more robust to noise~\citep{shukla2024comprehensive} and can lead to improved performance with fewer parameters~\citep{toscano2024inferring,shukla2024comprehensive} than MLPs. Subsequent studies extended the KAN framework and developed new architectures for PIML problems~\citep{howard2024finite,rigas2024adaptive,shuai2024physics,zhang4957859pkan, wang2024kolmogorovarnoldinformedneural,guilhoto2024deeplearningalternativeskolmogorov,koenig2024kan,patra2024physics,toscano2024inferring,wang2024expressiveness,liu2024kan2}.

Other representation models have also been explored, including convolutional neural networks (CNNs)~\citep{gao2021super}, Hermite spline CNNs~\citep{wandel2022spline}, generative adversarial neural networks (GANs)~\citep{yang2020physics,bullwinkel2022deqgan}, spiking neural networks~\citep{zhang2023artificial}, transformers~\citep{zhao2023pinnsformer}, long short-term memory (LSTM)  networks~\citep{cho2022lstm,nathasarma2023physics}, information-botleneck inspired architectures~\citep{guo2024ib} and reinforcement learning models~\citep{banerjee2023survey,ramesh2023physics,radaideh2021physics}.

\paragraph{Residual Connections } Another approach to improve the PIML architecture performance is by adding residual connections (see Fig.~\ref{RM_enh}), which enable accurate calculation of high-order derivatives. The general formulation for a single additive skip connection is defined as follows: \begin{align} \bm{z}^{(l) }(\bm{z}^{(l-1) },\theta) &=\mathcal{M}^{(l) }(\bm{z}^{(l-1) },\theta) +\bm{z}^{(l-1) }, \label{Residual} \end{align}

\noindent where $\bm{z}^{(l) }=\{z_{0}^{(l) },\ldots,z_{H}^{(l) }\}$ are the outputs of layer $l$, $\mathcal{M}^{(l) }$ is a layer of the representation model (e.g., MLP layer, KAN layer), and $\bm{z}^{(l-1) }=\{z_{0}^{(l-1) },\ldots,z_{H}^{(l-1) }\}$ are the inputs to layer $l$. Several studies have explored the advantages of incorporating skip connections via addition or multiplication. Wang et al.~\citep{wang2021understanding} pioneered this approach by introducing a method referred to as modified MLP, where two single-layer MLPs project the model inputs ($\bm{x}$)  into a high-dimensional feature space, which is then used to update the remaining hidden layers via element-wise multiplication and addition. Other approaches introduce multiplicative connections between every layer~\citep{jiang2024densely}. Finally, the modified MLP was further improved by incorporating adaptive residual connections that incorporate a new learnable parameter that controls the contribution of the deeper layers with respect to the input~\citep {wang2024piratenets}. This improved architecture enables the use of deeper networks without compromising the accuracy of PIML problems.

\paragraph{Model Decomposition}
Further improvements can be achieved by splitting the network into several components. Cho et al.~\citep{cho2024separable} proposed separable PINNs (sPINNs), which utilize separate sub-networks to approximate the desired solution.Under this formulation, the solution of a PDE is approximated as $\hat{u}(x_1,\cdots,x_d)  = \sum_{j=1}^p \prod_{i=1}^d \mathcal{M}_{j}(x_i,\theta_i)$
, where $\{\mathcal{M}_j(x_i,\theta_i) \}_{j=1}^p$ are the outputs univariate representation models that encodes each dimension $x_i$ into a $p$-dimensional space. This decomposition addressed the 
curse-of-dimensionality and was introduced to the deep learning community as tensor neural networks in~\citep{wang2022tensor_integral}. While~\citep{wang2022tensor_integral} focused on improving accuracy through enhanced numerical integration methods,~\citep{cho2024separable} prioritized computational efficiency, achieving up to 60-times speedups for high-dimensional problems, such as the 3D Helmholtz Equation and the 4D Navier-Stokes Equation. This formulation has been further explored and improved in subsequent studies~\citep{wang2024tensor,vemuri2024functional,hu2024tackling}.

Other types of decompositions have also been explored for inverse problems~\citep{toscano2024invivo,zhou2023flow}. For instance,~\citep{toscano2024invivo} extended the negative log-likelihood (NLL)  framework~\citep{lakshminarayanan2017simple} for linear PDEs in PIML, enabling the quantification of aleatoric uncertainty and improving model performance. The authors assumed that the observed data was corrupted by noise, so they decomposed the desired solution into mean fields and fluctuations as $u=\bar{u}+u'$. Then they used independent models to learn the data-driven mean fields \(\bar{u}\) and their corresponding fluctuations, standard deviations \(u'\)  by using the NLL criterion~\citep{quinonero2006machine,lakshminarayanan2017simple,cawley2005estimating}. On the other hand,~\citep {zhou2023flow} proposed a method to simultaneously reconstruct flow states and determine particle properties from Lagrangian particle tracking (LPT) using a neural network as a flow model and a data-constrained polynomial as a particle model. 

\subsection{Governing Equations}  

\begin{figure}
    \centering
    \includegraphics[width=\linewidth]{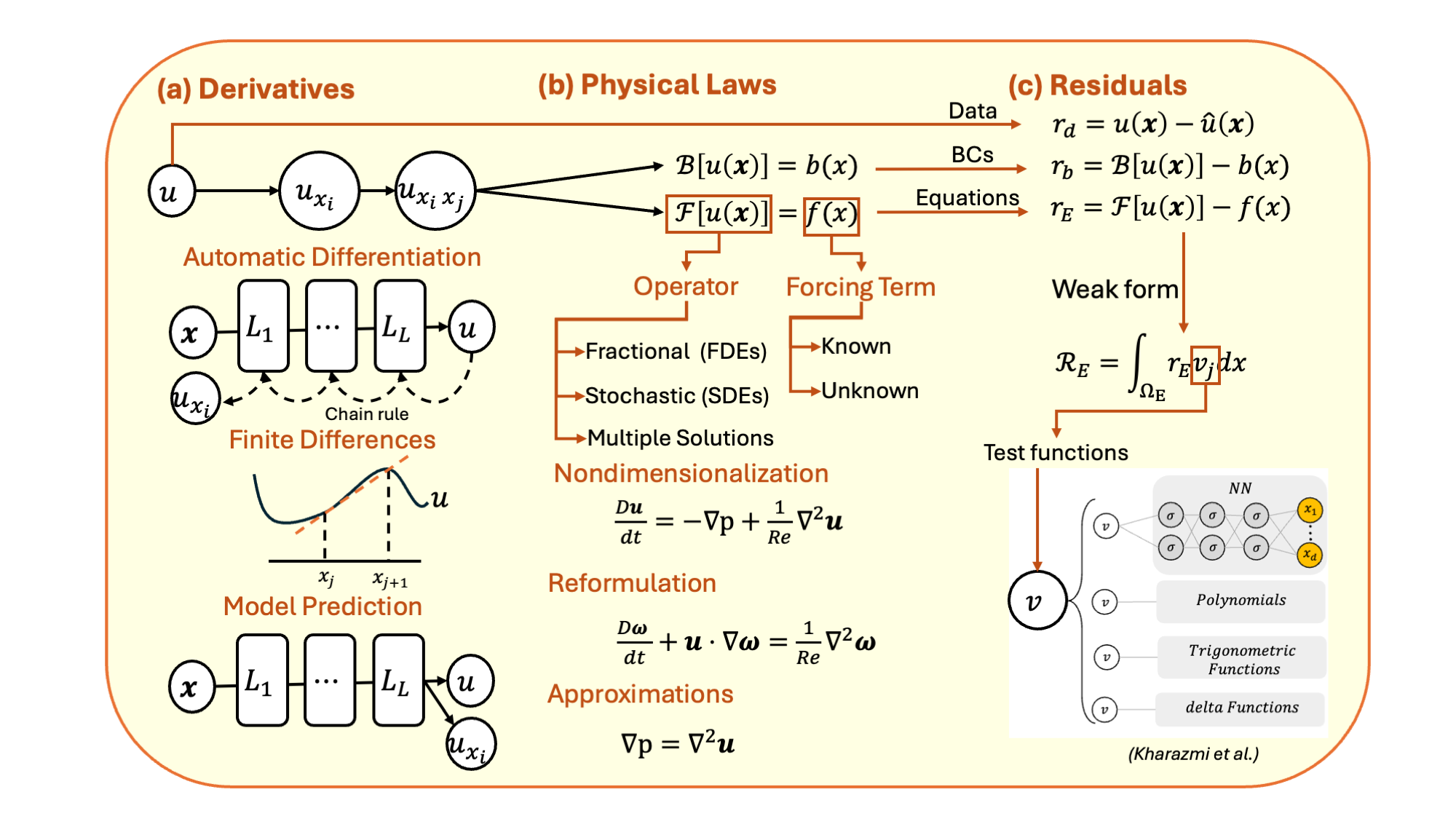}
    \caption{\textbf{Governing Equations}. (a) Derivatives are required to enforce the ODE/PDE, boundary conditions (BCs), and possibly data. The most common method to obtain derivatives is through automatic differentiation~\citep{baydin2018automatic}. However, other methods have been proposed, such as finite differences~\citep{lim2022physics}, estimating derivatives as outputs~\citep{gladstone2022fo}, or stochastic dimension gradient descent~\citep{hu2024tackling}. (b) The physical laws are imposed using various operators, including fractional~\citep{pang2019fpinns}, stochastic~\citep{yang2020physics}, and even those allowing multiple solutions~\citep{huang2022hompinns}. Model performance can be enhanced through non-dimensionalization~\citep{wang2022and}, reformulations~\citep{jin2021nsfnets}, or approximations~\citep{toscano2024invivo}. (c) Residuals can be enforced either in their strong form~\citep{raissi2019deep} or weak form~\citep{kharazmi2021hp}. In the weak form, several types of test functions have been explored.}
    \label{Govern_Eqn}
\end{figure}

PIML aims to obtain a representation model \( u \) that adheres to the governing equations. In the original study by~\citep{raissi2019physics}, the ODE/PDE and boundary conditions (BCs) are enforced by iteratively minimizing the strong-form residuals from the governing equations. The ODE/PDE residuals \( r_E(u, \theta) \) and boundary conditions (or initial conditions) \( r_B(u, \theta) \) are defined as:
\begin{align}
\label{PDE_res}
    r_E(x,\theta) &= \mathcal{F}_{\tau}[u](x,\theta) - f(x), \quad x \in \Omega_E, \\
\label{bcs_res}
    r_B(x,\theta) &= \mathcal{B}_{\tau}[u](x,\theta) - b(x), \quad x \in \Omega_B.
\end{align}
\noindent These residuals quantify the extent to which the approximation \( u \) satisfies the ODE/PDE and boundary constraints specified in Eq.~\ref{eq:problem}. If \( r_E = 0 \) and \( r_B = 0 \), the approximated solution  satisfies the PDE and BCs exactly.

For inverse problems, it is necessary to incorporate additional observations within the domain. The disagreement between the observations and predictions can be quantified using a data residual \( r_D(x,\theta) \), defined as:
\begin{align}
\label{data_res}
    r_D(x,\theta) &= u(x,\theta) - \hat{u}(x), \quad x \in \Omega_D,
\end{align}
\noindent where $\Omega_D\subseteq\Omega$ is the data domain. The objective of PIML is to find a solution that adheres to both the physical laws and the observational data.

\subsubsection{Derivative Calculation}

To enforce governing laws, it is necessary to compute the spatial and temporal derivatives of the approximated solution in order to construct and penalize PDE residuals. In the original formulation by~\citep{raissi2019physics}, these derivatives were computed exactly using automatic differentiation (AD). AD leverages the fact that all numerical computations are ultimately compositions of a finite set of elementary operations, for which derivatives are known~\citep{baydin2018automatic,raissi2024forward}. However, AD significantly increases computational cost due to the need for calculating and multiplying gradients at each layer, which can become inaccurate for higher-order derivatives~\citep{wang2024piratenets} and infeasible for fractional operators or high-dimensional problems~\citep{hu2024tackling}. To address these challenges, several studies have explored alternatives to or enhancements of backpropagation.

\paragraph{Alternative Differentiation Methods}
~\citep{lim2022physics} proposed approximating derivatives using finite differences, which speeds up computation; however, this method relies on a predefined grid, limiting its broader applicability. Other approaches~\citep{gladstone2022fo,buzaev2024hybrid} involve predicting derivatives as additional network outputs and learning the relationship through an auxiliary loss function. To handle fractional derivatives, several studies have employed Monte Carlo methods~\citep{pang2019fpinns,mehta2019discovering,ren2023class,wang2024gmc,sivalingam2024physics,hu2024tackling_fractional}.

\subparagraph{High-Dimensions}
One of the main advantages of PIML methods is their ability to handle high-dimensional problems~\citep{raissi2024forward,wang20222}. However, computing derivatives with AD becomes particularly challenging in such cases since the requirements for derivative calculation increase with the number of dimensions. Alternative approaches to AD have been proposed, particularly for high-dimensional problems. For instance,~\citep{he2023learning} introduced a Gaussian-smoothed model with Stein’s identity to parameterize PINNs, bypassing backpropagation and accelerating convergence. Additionally,~\citep{hu2024tackling} proposed Stochastic Dimension Gradient Descent (SDGD), a method that decomposes the gradient of PDEs and PINN residuals into components corresponding to different dimensions. During each training iteration, a subset of these dimensional components is randomly sampled, resulting in a highly efficient approach that enables solving PDEs with up to 100,000 dimensions.

\subsubsection{ODE/PDE Reformulations}
To enhance the performance of PIML models, several studies have proposed reformulations of the ODE/PDEs.

\paragraph{Non-dimensionalization}
As discussed in~\citep{wang2023expert}, one of the simplest and most effective ways to improve model performance is through non-dimensionalizing the governing equations. In this approach, the inputs and predicted outputs are scaled using characteristic units, helping to identify important non-dimensional numbers (e.g., Reynolds, Peclet, Prandtl, Rayleigh) that characterize the solution's behavior. Additionally, choosing appropriate characteristic units can control the magnitude of the inputs and outputs, which is crucial for stabilizing the training process. Several studies have successfully approximated solutions to PDEs in their non-dimensional form~\citep{raissi2020hidden,jin2021nsfnets,cai2021artificial,cai2021flow,toscano2024inferring,toscano2024invivo,wang2021understanding,wang2023solution}.

\paragraph{Equivalent and Auxiliary Formulations}
Another approach to improving model performance involves transforming the governing equations into an equivalent form that simplifies the optimization problem. For example, Wang et al.~\citep{wang2021understanding} and subsequent studies~\citep{anagnostopoulos2024learning} solved the Navier-Stokes equations using the streamfunction formulation, which inherently satisfies the conservation of mass, thereby reducing the number of constraints to optimize. Similarly, Jin et al.~\citep{jin2021nsfnets} reformulated the Navier-Stokes equations into their vorticity formulation, achieving better performance. Basir et al.~\citep{basir2022investigating} introduced an auxiliary vorticity variable, which lowered the order of the Stokes equations, further simplifying the problem. In some cases, these reformulations are necessary to achieve an acceptable solution. For instance, Toscano et al.~\citep{toscano2024inferring} reformulated the Rayleigh-Bénard equations using the vorticity formulation, eliminating the pressure dependence and enabling the inference of temperature from sparse turbulent velocity data. Similarly, Wang et al.~\citep{wang2023solution} reformulated the Navier-Stokes equations with an entropy-viscosity method, allowing for the approximation of solutions at high Reynolds numbers.

\subsubsection{Differential Operator Variations}  
 The PIML approach is flexible enough to solve several types of problems even with multiple solutions; for instance, Huang et al.~\citep{huang2022hompinns} proposed homotopy physics-informed neural networks (HomPINNs) for solving multiple solutions of nonlinear elliptic differential equations. This flexibility allows handling residuals in their strong or weak form and obtaining approximate solutions from different types of operators, leading to various PIML extensions.

\paragraph{Variational Methods}  
 Several studies have proposed weakly enforcing the PDE and boundary constraints by solving the problem in its variational form. This approach, known as variational PINNs (vPINNs), was introduced by~\citep{kharazmi2019variational}. Similar to the Deep Ritz Method~\citep{yu2018deep}, vPINNs compute weighted integrals of the residuals by projecting them onto a suitable space of test functions \( V \)~\citep{kharazmi2021hp}. In the variational form, the residuals are represented as:
\begin{align*}
    \mathcal{R}_{e,j}(u) &= \int_{\Omega} r_e(x,\theta) v_j \, dx, \\
    \mathcal{R}_{b,j}(u) &= \int_{\Omega_B} r_b(x,\theta) v_j \, dx, 
\end{align*}
where \( v_j \) represents a chosen test function. Ideally, the exact solution is obtained when all residuals are identically zero~\citep{kharazmi2021hp}. Various types of test functions have been proposed, including Dirac-delta functions~\citep{raissi2019deep}, global~\citep{kharazmi2019variational}, piece-wise~\citep{khodayi2020varnet} polynomials, and non-overlapping functions~\citep{kharazmi2021hp}. The vPINN formulation has also been explored in the context of mesh-free methods~\citep{berrone2024meshfree}, variable coefficients~\citep{miao2023vc}, volume-weighted methods~\citep{song2024vw}, and has been optimized for computational efficiency~\citep{ghose2024fastvpinns,anandh2024efficient}.

\paragraph{Fractional Differential Equations}  
Another extension involves fractional operators, giving rise to the fractional PINNs (fPINNs) framework, first introduced by~\citep{pang2019fpinns}. Several studies have explored and expanded upon the fPINN formulation~\citep{mehta2019discovering, ren2023class, guo2022monte, wang2024gmc, sivalingam2024physics, hu2024tackling_fractional}. For example,~\citep{wang2024gmc} proposed a Monte Carlo-based method to solve fractional partial differential equations on irregular domains. Furthermore,~\citep{sivalingam2024physics} provided a theoretical analysis to estimate the training and generalization errors for the \(\psi\)-Caputo type fractional PDE. Lastly,~\citep{hu2024tackling_fractional} extended the fPINN framework to overcome the curse of dimensionality in fractional and tempered fractional PDEs.

\paragraph{Stochastic Differential Equations} Stochastic Differential Equations (SDEs) have also been explored within the PIML framework. For instance,~\citep{yang2020physics} utilized GANs as a representation model and applied automatic differentiation to encode the governing laws for solving SDEs. Similarly,~\citep{zhang2020learning} combined spectral dynamically orthogonal (DO) and dynamically biorthogonal (BO) methods with PIML to develop two novel PINN approaches for solving time-dependent SDEs. Furthermore,~\citep{raissi2024forward} extended the PIML formulation to solve high-dimensional SDEs; see also~\citep{hu2024score} for more recent PINN algorithms for high-dimensional Fokker-Planck and Hamilton-Jacobi-Bellman equations.

\subsection{Optimization Process}
\label{optim_proces}

Training a PIML model involves solving an optimization problem that enables a representation model to approximate the solution of a governing equation. The model parameters are learned by optimizing a loss function (i.e., objective function), which minimizes the residuals (i.e., Eqns. \ref{PDE_res},~\ref{bcs_res}, and~\ref{data_res}) of the governing equations, boundary conditions, and data observations. This loss function is minimized using an optimization algorithm, often referred to as an ``optimizer." Based on this description, the optimization process in PIML can be broken down into three main subcomponents: the optimization problem, the loss (i.e., objective) function, and the optimizer.

\subsubsection{Optimization Problem}
\label{optim_problem}

\begin{figure}[!ht]
    \centering
    \includegraphics[width=\linewidth]{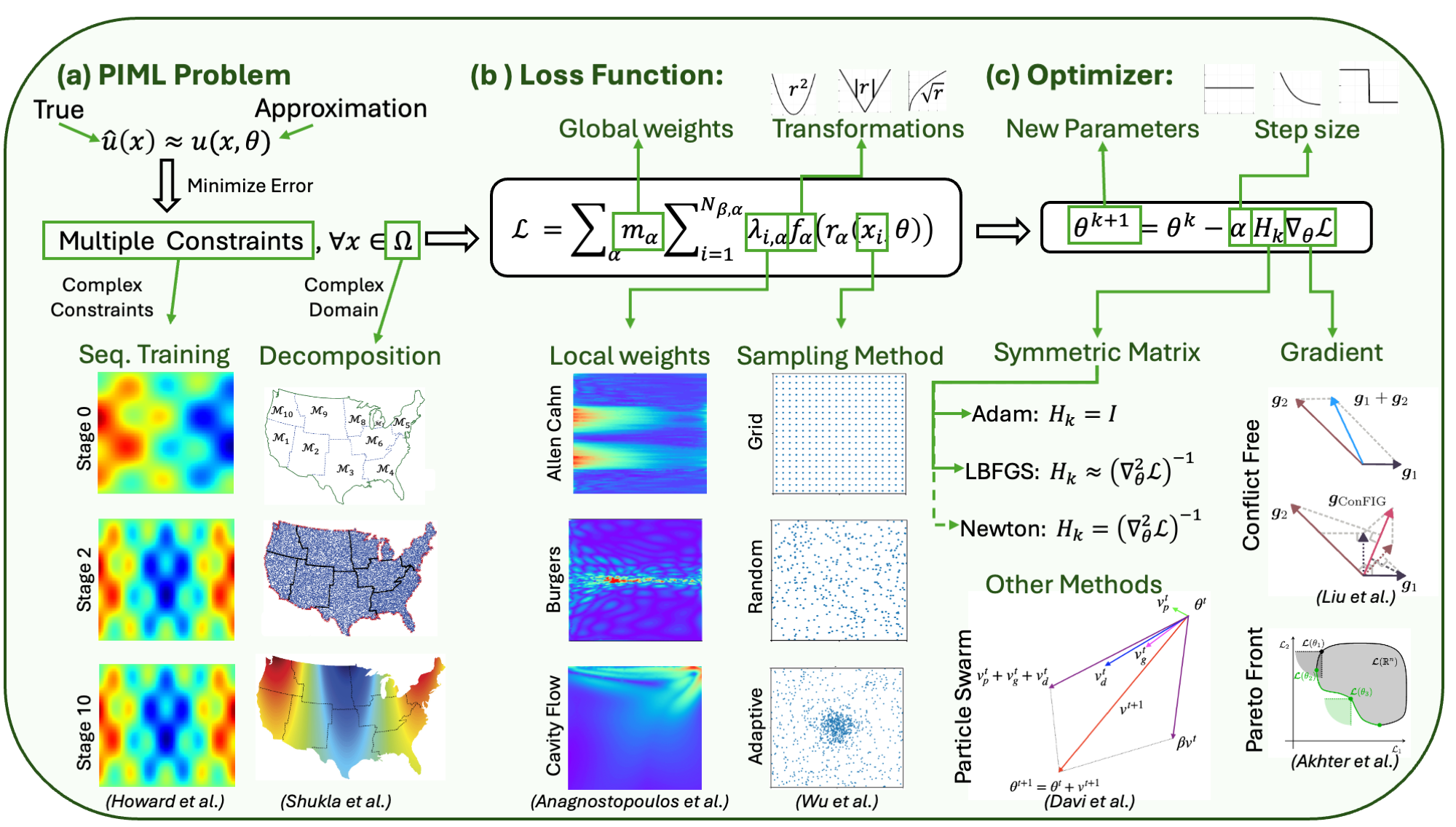}
\caption{\textbf{Optimization Process Enhancements}. The PIML problem involves satisfying multiple constraints in a given domain. For complex constraints, the problem can be simplified using sequential training or stacked training~\citep{howard2023stacked}. When dealing with complex domains (e.g., large, irregular), domain decomposition can be employed~\citep{shukla2021parallel}. (b) The loss function measures the cumulative error in the domain and, once discretized, comprises global weights~\citep{wang2021understanding}, local weights~\citep{anagnostopoulos2024residual}, a sampling method~\citep{wu2023comprehensive}, and a transformation function. (c) The optimizer updates the parameters using a line search method~\citep{urban2024unveiling}. The choice of the symmetric matrix defines the optimization method, such as gradient descent, Adam, L-BFGS, or Newton’s method. Several approaches have improved base performance, ensuring conflict-free updates~\citep{liu2024config}, or by using alternative methods like non-dominated genetic algorithms~\citep{akhter2024common}, or particle swarm optimization~\citep{davi2022pso}.}

    \label{Optim_Enh}
\end{figure}
The optimization problem in PIML can be summarized as minimizing a multi-objective loss function that encourages the model to satisfy constraints related to boundary conditions, governing equations, and potentially observational data. To simplify this problem, several studies have sought to reduce the number of constraints. For instance, in systems of PDEs,~\citep{shukla2024neurosem} proposed using numerical solvers to provide easily accessible high-fidelity data, using the PIML model primarily to uncover hidden fields that are difficult to recover with traditional methods. Other approaches aim to simplify the problem through domain decomposition, learning from low-fidelity data, and sequential training strategies.

\paragraph{Domain Decomposition}  
The baseline approach for domain decomposition was introduced in the extended PINNs (XPINNs) framework~\citep{jagtap2020extended} and can be summarized as follows. First, the training domain \( \Omega \) is divided into \( N \) smaller subdomains \( \Omega_i \subset \Omega \). Then, \( N \) PIML submodels are trained to approximate the solution on each subdomain \(\Omega_i \). Various studies have explored methods for optimally partitioning \( \Omega_i \) and ensuring communication between the submodels. For example,~\citep{jagtap2020extended,hu2021extended,de2024error} proposed incorporating an additional loss term for interface conditions to ensure that submodel predictions and residuals align at the subdomain boundaries. Conservative PINNs (cPINNs)~\citep{jagtap2020conservative} follow a similar approach, ensuring the continuity of fluxes across subdomains, akin to traditional numerical methods. Additionally,~\citep{hu2023augmented} introduced a soft domain decomposition method using a shared subnetwork to route inputs into different submodels. Similar methods have also been published in~\citep{meng2020ppinn, shukla2021parallel, moseley2023finite, dolean2024multilevel, liu2023cv, kharazmi2021hp, nguyen2022efficient, kopanivcakova2024enhancing}, achieving accelerated training, easy parallelization, and improved accuracy.

\paragraph{Sequential Training}  
Sequential training can be viewed as a form of ``problem decomposition," where the model learns or satisfies objectives sequentially. Ahmadi et al.~\citep{CMINNs} proposed a novel method for addressing specific challenges in inverse problems, where the values of constant parameters change at specific times and the system experiences abrupt spikes due to sudden input changes. In this case, it is essential to train the model on sequential intervals, which are strategically chosen to capture the spikes observed in the system. For example, in the CMINNs method, the tumor growth model following drug administration exhibits spikes that cannot be effectively represented without decomposition. Furthermore, domain decomposition is necessary for the inference of piecewise-constant parameter values over continuous intervals, enhancing the ability to capture variations in drug efficacy and providing insights into tolerance phenomena in multi-dose administration.

\subparagraph{Time Decomposition}  
For time-dependent problems, several studies have proposed training the model over a short time interval before gradually expanding the training window until the entire time domain is covered~\citep{wight2020solving,krishnapriyan2021characterizing,mattey2022novel,haitsiukevich2023improved,wang2022respecting, chen2024leveraging2}.
A unified approach of the various proposed methods for wave propagation problems was presented in~\citep{penwarden2023unified}.

\subparagraph{Transfer Learning and Curriculum Training}  
These strategies involve initializing the model parameters by pretraining on a simpler problem and then retraining on the target problem. Once the model is pre-trained, transfer learning approaches~\citep{desai2021one,liu2023adaptive,xu2023transfer,chen2024enhancing,chen2024leveraging} demonstrate that fine-tuning only the final layers can significantly improve model performance. Curriculum training, by contrast, progressively increases the problem complexity and has proven useful for systems of PDEs~\citep{krishnapriyan2021characterizing,zhang2024meshless,wang2023expert,wang2024piratenets} as well as inverse problems~\citep{ahmadi2024ai,CMINNs,toscano2024invivo,toscano2024inferring}. For instance,~\citep{wang2023expert} solved the 2D Navier-Stokes equations at high Reynolds numbers (Re) by gradually increasing 
the Reynolds number during training. Notably,~\citep{wang2024piratenets} proposed initializing the last layer (i.e., linear) using least squares and then employing an adaptive residual connection that progressively includes deeper layers as necessary. Other curriculum training methods include pre-training on simulated data or theoretical models, alternating between learned fields~\citep{zhang2024meshless}, or adjusting the loss function to focus on refining the solution~\citep{toscano2024invivo,toscano2024inferring}. For inverse problems that require inferring hidden fields,~\citep{toscano2024inferring} suggested first fitting the data and boundary conditions, then learning a theoretical representation of the hidden field, and finally incorporating the full physics. Using this approach, the authors successfully inferred hidden temperature fields from sparse experimental turbulent velocity data.

\subparagraph{Multi Fidelity and stacked training}
Multifidelity and stacked training methods have been employed to improve the prediction accuracy of PINNs.
Mutlifidelity PINNs proposed in~\citep{meng2020composite} provide a framework for integrating low- and high-fidelity data.
This model involves four neural networks: the first approximates low-fidelity data, the second and third learn linear and nonlinear correlations between the low- and high-fidelity data, and the final network encodes the underlying PDEs.
The multifidelity approach has been extended to Bayesian networks, quantifying uncertainties in the prediction~\citep{meng2021multi}.
Another approach to learning from multi-fidelity data has been proposed in~\citep{regazzoni2021physics}, involving a NN to approximate the low-fidelity data and a subsequent NN to learn the correction term using the high-fidelity data and a physics-informed loss.
Multifidelity approaches have been successfully applied to solve the inverse-water wave problem governed by Serre-Green-Naghdi equations~\citep{jagtap2022deep}. Stacked training techniques, presented in~\citep{howard2023stacked,wang2024multi}, is another approach to improve the prediction accuracy in PINN. 
By stacking networks sequentially, the output from each step serves as low-fidelity input for subsequent stages, allowing the model to refine predictions progressively.
This approach has been further improved in~\citep{heinlein2024multifidelity} by combining multi-fidelity stacking with domain decomposition methods, making it practical for multiscale time-dependent problems. A multi-stage neural network was introduced in~\citep{wang2024multi} to tackle spectral bias by dividing the training into stages, where each stage involves training a new NN to fit the residue from the previous stage. Also,~\citep{penwarden2022multifidelity} presented an approach to combine different neural networks of varying fidelities by exploiting their low-rank structure.


\subsubsection{Loss Function Modifications}
\label{optim_loss}

The loss function (\( \mathcal{L} \)) quantifies the disagreement between the approximation provided by the representation model and known information from the PDE solution, such as boundary conditions, ODE/PDE residuals, and data. Training in PIML involves iteratively minimizing the loss subcomponents (\( \mathcal{L}_\alpha \)) over their respective domains \( \Omega_\alpha \). In general, \( \mathcal{L} \) can be roughly represented as:
\begin{align}
\mathcal{L}(\theta) &= \sum_{\alpha \in C} m_\alpha \int_{\Omega_{\alpha}} f_\alpha(r_\alpha(x, \theta)) dx,
\label{loss_cont}
\end{align}
\noindent where \( C = \{D, B, E, \dots \} \) is an index specifying the loss groups, e.g., data (\( \mathcal{L}_D \)), boundary (\( \mathcal{L}_B \)), and equation (\( \mathcal{L}_E \)). The function \( f_\alpha: \mathbb{R} \rightarrow \mathbb{R}^{+} \) is a positive, preferable convex, function applied to the residuals \( r_\alpha \) of each subcomponent (i.e., Eqns.~\ref{bcs_res},~\ref{PDE_res}, and~\ref{data_res}), and \( m_\alpha \) are scalar weights that balance the contributions of each term, often referred to as global weights. 

To enable computation, this expression is typically discretized and computed iteratively over finite subdomains \( X_{\beta,\alpha} \subset \Omega_\alpha \) as follows:
\begin{align}
\mathcal{L}(\theta, X_{\alpha,\beta}) &= \sum_{\alpha \in C} m_\alpha \sum_{i=1}^{N_{\beta,\alpha}} \lambda_{\alpha,i} f(r_\alpha(x_i, \theta)), \quad \text{where } x_i \in X_{\alpha,\beta},
\label{loss_dic}
\end{align}
\noindent where \( N_{\beta,\alpha} \) is the batch size, representing the number of points \( x_i \) in the subset \( X_{\beta,\alpha} \), sampled from a probability density function \( p_\alpha \) over the domain \( \Omega_\alpha \). Similar to weighted Monte Carlo methods~\citep{liu2001monte}, the discrete residuals \( f(r_\alpha(x_i, \theta)) \) are scaled using pointwise multipliers, referred to as local weights \( \lambda_{\alpha, i} \).

Defining a suitable \( \mathcal{L} \) is crucial to ensure that model predictions align with the PDE solution, and ongoing research has focused on refining four key components: the global weights (\( m_\alpha \)), the local weights (\( \lambda_{\alpha,i} \)), the choice  of function (\( f_\alpha \)), and the sampling strategy, which is indirectly based on the probability density function (\( p_\alpha \)) and the number of points (\( N_\alpha \)).

\paragraph{Global Weights}
The global weights (\( m_\alpha \)) balance the contribution of each loss subterm and ensure that all constraints are satisfied. These weights can be either fixed, as in the original PIML framework~\citep{raissi2019physics,raissi2020hidden}, or dynamic, adjusting their magnitude during training. In particular, Wang et al.~\citep{wang2021understanding} proposed a learning rate annealing algorithm that dynamically adjusts the global weights based on back-propagated gradients. This approach improved performance and was successfully applied in diverse applications~\citep {jin2021nsfnets,boster2023artificial,cai2021artificial}. Similarly, self-adaptive loss balancing methods, such as those proposed by~\citep{xiang2022self}, also dynamically adjust global weights during training. Liu et al.~\citep{liu2021dual} developed a dual-dimer method for training PINNs with a minimax architecture, optimizing the global weights to manage complex multi-objective problems. Additionally, Basir et al.~\citep{basir2022investigating} explored failure modes in PINNs and refined global weight strategies to enhance model robustness. Finally, specific forms of sequential training can be seen as iteration-based weight adjustments, where the model’s learning process evolves. For example, Wang et al.~\citep{wang2022respecting} introduced a causal parameter for time-dependent problems, which forces the model to learn sequentially, adjusting the weights based on time steps to capture causality. Overall, global weights—whether static, dynamic, or iteration-based—play a critical role in ensuring the effectiveness of PIML models, and ongoing research continues to refine these weights to improve performance. In particular~\citep{wang2022and} extended the Neural Tangent Kernel(NTK) to PIML and proposed a novel adaptive global weighting training strategy that significantly enhance the trainablity and predictive accuracy of PIML models.

\paragraph{Local Weights}

One of the main challenges in PIML is that residuals at key points can be underrepresented in the overall summation of the objective function (Eq.~\ref{loss_dic}). As a result, despite a decrease in total loss during training, certain spatial or temporal characteristics might not be fully captured. This issue becomes particularly pronounced in multiscale problems, where regions of interest may lack detail, and important information from the initial and boundary conditions may not propagate effectively through the domain, thereby hindering convergence~\citep{anagnostopoulos2024residual}. To address this issue, researchers have proposed assigning local weights \( \lambda_{\alpha,i} \) to balance the contribution of each residual point, thus increasing focus on the challenging regions in both space and time dimensions.

McClenny et al.~\citep{mcclenny2023self} introduced a self-adaptive (SA) approach, where individual loss weights are adjusted through adversarial training. Building on this idea, Zhang et al.~\citep{zhang2023dasa} proposed a differentiable adversarial self-adaptive (DASA) weighting scheme, which uses a subnetwork to optimize the local multipliers. Basir et al.~\citep{basir2022physics} developed the physics and quality-constrained artificial neural network (PECANN), which calculates local weights based on the residuals of constraints, such as initial and boundary conditions, using the augmented Lagrangian method. PECANN has since been expanded with adaptive versions, such as PECANN-AL, which incorporate global Lagrange multipliers~\citep{basir2022investigating,basir2023adaptive}. Similarly, Son et al.~\citep{son2023enhanced} introduced an augmented Lagrangian relaxation method for PINNs (AL-PINNs), where the initial and boundary conditions act as constraints to optimize the PDE residual.

Anagnostopoulos et al.~\citep{anagnostopoulos2023residual} introduced residual-based attention (RBA) weights, where \( \lambda_{\alpha,i} \) is computed using the exponentially weighted moving average of the residuals. Since residuals contain information about regions with high error, this method proved to be highly effective, outperforming previous approaches with minimal computational cost. These techniques have been further developed and refined in subsequent studies~\citep{song2024loss,shukla2024comprehensive,ramirez2024residual,chen2024self}. Notably, Chen et al.~\citep{chen2024self} extended both SA and RBA methods using the Neural Tangent Kernel~\citep{wang2022and} and analogies to traditional numerical methods, resulting in a robust and improved algorithm for calculating \( \lambda_{\alpha,i} \).

\paragraph{Sampling}

As shown in Eq.~\ref{loss_dic}, PIML models are iteratively optimized on \( N_\alpha \) points from a discrete subdomain \( X_{\beta,\alpha} \), which is sampled from \( \Omega_{\alpha} \) using a suitable probability density function \( p_\alpha \). In this context, the sampling method refers to the specific choice of \( N_\alpha \) and \( p_\alpha \), which define the number and distribution of the training points.

Sampling methods in PIML can be categorized based on uniformity, adaptability, and selection criteria. In terms of uniformity, these methods are divided into uniform and non-uniform sampling techniques. Early approaches employed simple methods like equispaced grids and uniformly random sampling~\citep{lu2021deepxde, wu2023comprehensive}. Later, more sophisticated non-adaptive uniform sampling techniques were introduced, such as Latin Hypercube Sampling (LHS), Halton, Hammersley, and Sobol sequences.

However, recent studies have demonstrated that uniform sampling methods are often insufficient, particularly for solving PDEs with sharp gradients~\citep{wu2023comprehensive}. This has led to the development of adaptive sampling methods, which dynamically adjust the sampling of residual points based on certain criteria. Broadly, adaptive sampling can be classified into two main strategies: (1) \textbf{Resampling} (adaptive \( p_\alpha \)), where all residual points are resampled after a fixed number of iterations according to the specified criteria, and (2) \textbf{Incremental sampling} (adaptive \( N_\alpha \)), where an initial set of residual points is sampled, and additional points are incrementally added during training, guided by either the same or different criteria.

Adaptive sampling methods employ various selection criteria. While uniform sampling methods may serve as a baseline, non-uniform criteria are more commonly used. The most prevalent approach defines \( p_\alpha \) to be proportional to the PDE residual, concentrating the sampling in regions where the residual is large~\citep{wu2023comprehensive, daw2022rethinking, gao2023active, tang2021deep, peng2022rang, zeng2022adaptive, hanna2022residual, subramanian2023adaptive, nabian2021efficient, zapf2022investigating, toscano2024inferring,daw2022mitigating}. This strategy enhances the distribution of residual points by focusing on areas that contribute most significantly to the PDE loss. Alternatively, other methods prioritize high-error regions without directly relying on the PDE residual. For example,~\citep{toscano2024inferring} proposed using pre-computed local multipliers \( \lambda_{\alpha,i} \), which encode historical information about high-error regions, to update \( p_\alpha \) at every training iteration with negligible computational cost. Another approach, failure-informed adaptive sampling~\citep{gao2023failure, gao2023failure2}, defines a failure probability function and adds new residual points (i.e., increases \( N_\alpha \)) in regions where this probability exceeds a predefined threshold.

\paragraph{Function Selection}

As shown in Eq.~\ref{loss_cont}, \( f_\alpha: \mathbb{R} \rightarrow \mathbb{R}^{+} \) is a positive, preferable convex function applied to the residuals \( r_\alpha \) (i.e., Eqns.~\ref{bcs_res},~\ref{PDE_res}, and~\ref{data_res}). This function aims to transform the residuals and give $\mathcal{L}$ the required characteristics to be optimizable. The most prevalent choice of this function is  $f_\alpha(z,p)=|z|^p$, which transforms $\mathcal{L}$ into the the sum of $L^p$ norm of the residuals for each loss $\mathcal{C}$ subcomponent,($\mathcal{L}=\sum_{\alpha\in\mathcal{C}}\|r_{\alpha}(x)\|^p_p$). Notice that, as described in~\citep{wang20222}, by setting $p=2$ and discretizing with $\lambda_{\alpha,i}=(1/N)$, we recover the mean-squared error as introduced in the first PINN study~\citep{raissi2019physics} and broadly adopted in the PIML community. However, the $L^2$ norm tends to be sensitive to outliers and diminish small values, so it struggles capturing small details; thus, several studies propose training by a combination of $L^1$ and $L^2$ using a sequential approach~\citep{toscano2024inferring} or adaptively via the Huber Loss~\citep{chen2024enhancing,bullwinkel2022deqgan}. On the other hand, Wang et al.,~\citep{wang20222} investigated the relationship between the loss function and the approximation quality of the learned solution and proved that for general $L^p$ loss, several types of equations are stable only if $p$ is sufficiently large and developed a noble algorithm to minimize the $L^{\infty}$ loss.  Other types of functions have also been explored. For instance, ~\citep{urban2024unveiling} experimentally showed that using $f_\alpha(z)=log(z)$ or $f_\alpha(z)=\sqrt{|z|}$ can significantly improve the baseline model performance. Other studies~\citep{toscano2024invivo} proposed using the negative-log-likelihood (NLL), introduced in~\citep{quinonero2006machine,lakshminarayanan2017simple,cawley2005estimating}, for inverse problems in PIML enabling obtaining the aleatoric uncertainty and improving the model performance. Similarly,~\citep{zhou2023stochastic} proposed the stochastic particle advection velocimetry method, which introduces a statistical data loss that improves the accuracy of inverse problems in fluid dynamics. This method is based on an explicit particle advection model that predicts particle positions over time as a function of the estimated velocity field.

\subsubsection{Optimizer}
\label{optim_omptimizer}
The goal of the optimizer is to find the optimal parameters \( \theta \) so that the approximated solution matches as close as possible to the true solution. This process is performed iteratively by gradually minimizing \( \mathcal{L} \) until a desired accuracy is reached. As described in~\citep{urban2024unveiling}, the most common optimizers in PIML fall under the general family of Line Search Methods~\citep{nocedal1999numerical}, where \( \theta \) are updated as follows:
\begin{align}
\label{General_optimizer}
\theta^{k+1}&=\theta^{k}+\alpha^{k}p^{k},\\
p^{k}&=-H_k\nabla_{\theta}\mathcal{L}(\theta^k).
\label{step_optimizer}
\end{align}

Here, \( \alpha^k \) is the step size at iteration \( k \), and \( p^k \) is the step direction, which depends on the loss gradient \( \nabla_{\theta}\mathcal{L}(\theta^k) \) and a symmetric matrix \( H^k \). Under this formulation, linear methods such as gradient descent or ADAM~\citep{kingma2014adam} are recovered by setting \( H_k = I \). On the other hand, quasi-Newton methods, such as L-BFGS~\citep{liu1989limited}, can be recovered by setting \( H^k \) to an approximation of the Hessian matrix of \( \mathcal{L} \), which requires only first-order derivatives and helps achieve superlinear convergence~\citep{urban2024unveiling}.

Due to their computational efficiency and mini-batch flexibility, linear methods are widely used in the PIML community, and several studies have focused on improving their performance~\citep{lu2023nsga,liu2024config,zhou2023generic,yao2023multiadam,fang2023ensemble}. For instance,~\citep{cyr2020robust} proposed adopting an adaptive basis viewpoint of neural networks, which led to novel initialization and a hybrid least squares/gradient descent optimizer. Similarly, \citep{ainsworth2021plateau} proposed an iterative training method, the Active Neuron Least Squares, characterized by explicitly adjusting the activation pattern at each step, designed to enable a quick exit from a plateau. Also, \citep{ainsworth2022active} proposed augmenting the standard gradient descent direction by including search vectors, which are chosen to explicitly adjust the activation patterns of the neurons, which improved the performance of two-layer rectified neural networks. On the other hand,~\citep{lu2023nsga} proposed using Non-dominated Sorting Genetic Algorithms (NSGA) to help traditional stochastic gradient optimization algorithms escape local minima. Similarly~\citep {akhter2024common} analyzed the Pareto front, highlighted the most common pitfalls for multi-objective problems, and compared the standard methods with NSGA-II.

Other studies have focused on improving gradients in multi-task learning~\citep{liu2024config, zhou2023generic, yao2023multiadam}. For example,~\citep{liu2024config} introduced a conflict-free update algorithm to handle multi-objective optimization, ensuring a positive dot product between the final update and each loss-specific gradient. Conversely,~\citep{davi2022pso} proposed replacing gradient descent with particle swarm optimization, which not only improves performance but also allows for the computation of related uncertainties. {Finally,~\citep{jnini2024gauss} propose Gauss-Newton's method in function space for the solution of the Navier-Stokes equations. Upon discretization, this yields a natural gradient method that mimics the function space dynamics and allows the authors to achieve close to single-precision accuracy in the relative $L^2$ norm~\citep{jnini2024gauss}.

In some applications, quasi-Newton methods such as BFGS or L-BFGS~\citep{liu1989limited} can be used to achieve better performance with fewer iterations, though they are more prone to getting trapped at saddle points~\citep{urban2024unveiling}. To mitigate this, some studies recommend using Adam during the initial stages of training, followed by L-BFGS for fine-tuning~\citep{raissi2019physics, jin2021nsfnets, urban2024unveiling}. Given the effectiveness of quasi-Newton methods, recent research has aimed at further enhancing their performance.  For instance,~\citep{urban2024unveiling} proposed a modified BFGS algorithm, demonstrating that by selecting an improved optimization method and incorporating modifications to the loss function, the accuracy of solutions comparable to finite-difference schemes can be achieved in specific examples. On the other hand,~\citep{rathore2024challenges} theoretically analyzed PIML ill-conditioning and introduced a novel second-order optimizer that significantly improves PIML performance. Similarly~\citep{muller2023achieving} proposed energy natural gradient descent with respect to a Hessian-induced Riemannian metric as an optimization
algorithm for PINNs yielding highly accurate solutions for shallow networks, outperforming ADAM and LBFGs. This study was extended to deeper networks and further improved in\citep{dangel2024kronecker}. Finally,~\citep{lee2024two} introduced a two-level overlapping additive Schartz preconditioner strategy that can be combined with any optimizer to accelerate the training of PIML problems.


\section{Applications of PIML}
\label{applications}
Numerous studies have demonstrated the success of PIML across a wide range of fields. Here, we provide a selective yet comprehensive review of PIML applications in biomedicine, mechanics, geophysics, dynamical systems, control and autonomy, heat transfer, physics, chemical engineering, and other miscellaneous areas.

\begin{figure}[!ht]
    \centering
    \includegraphics[width= \linewidth]{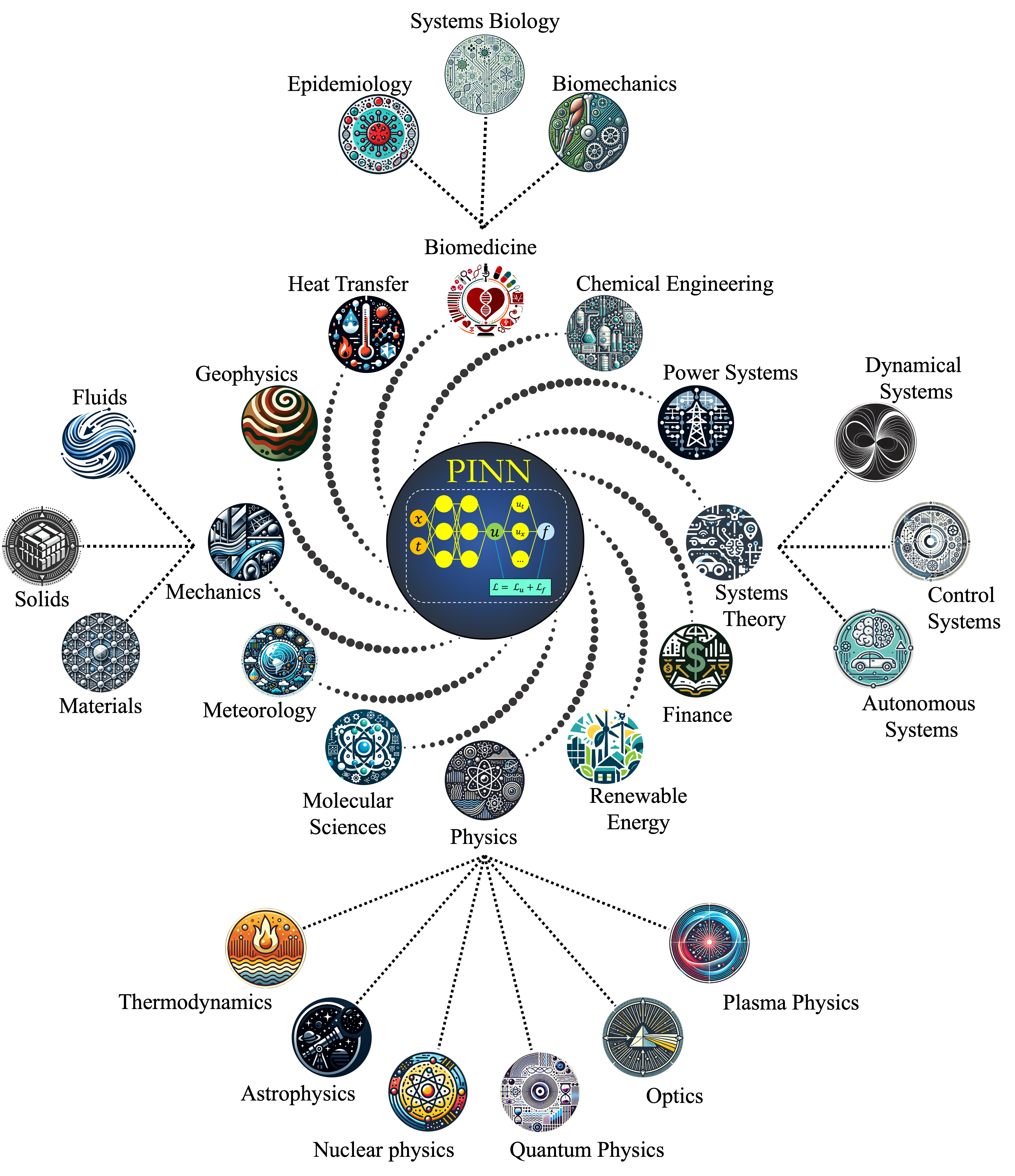}
    \caption{\textbf{Applications of PINNs.} An illustration of the diverse applications where PINNs have been employed by incorporating the underlying physical laws. (The individual logos were created using DALL.E~\cite{OpenAI2024DALLE})}
    \label{fig:pinn_applications}
\end{figure}

    



\subsection{Applications in Biomedicine}

The potential of PINNs have been demonstrated across a wide range of biomedical applications, including systems biology, systems pharmacology, biomechanics, and epidemiology. By integrating biophysical laws with data-driven learning, PINNs offer a powerful framework for solving complex problems in these fields.

In systems biology, for example, PINNs have been employed to model dynamic biological processes. Notably, the work by~\citep{yazdani2020systems} and~\citep{daneker2022systems} successfully applied PINNs to model complex biochemical reactions and conduct parameter identification for system-level biological processes. Building on this, the AI-Aristotle framework~\citep{ahmadi2024ai} has further demonstrated the adaptability of PINNs in both systems biology and systems pharmacology gray-box discovery, showing how these models can handle the intricate dynamics of biological systems~\citep{ahmadi2024ai}. Beyond these applications, PINNs have also been leveraged to explore cellular signaling pathways. For instance,~\citep{jo2024density} introduced Density-PINNs to infer transduction-time distributions in cellular signaling, providing insights into how response times contribute to cell-to-cell heterogeneity. This understanding has the potential to inform more effective disease treatment strategies. 

Transitioning to cardiac health, PINNs have made significant strides in diagnosing and treating atrial fibrillation. They have been used to create detailed activation maps for diagnosis and estimate cardiac fiber orientation from electroanatomical data, both of which are essential for personalized treatment and procedural planning~\citep{sahli2020physics, ruiz2022physics}. Additionally, in the area of blood coagulation—a critical process in hemostasis—PINNs address challenges in parameter estimation due to the difficulty of measuring reaction rates. The introduction of Coagulo-Net~\citep{QIAN2024106732} demonstrated how PINNs can infer unknown parameters and dynamics from sparse and noisy data, offering a robust solution to modeling blood clotting. Similarly, TGM-Net has been proposed in~\citep{chen2023tgm} to facilitate tumor growth forecasting. Finally, in electrophysiology, the precise simulation of action potentials and the estimation of electrophysiological tissue properties are essential for diagnosing and treating arrhythmias such as atrial fibrillation.~\citep{herrero2022ep} presented EP-PINNs, which offer highly accurate simulations and parameter estimation from sparse datasets, paving the way for better treatment outcomes.

In the field of systems pharmacology, PINNs solve the compartment model to predict the assimilation of drugs in the human body, enhancing the understanding of pharmacokinetics for better therapeutic outcomes~\citep{goswami2021study, ahmadi2024ai}. In this study~\citep{podina2024learning}, PINNs are employed to uncover unknown components in differential equations modeling chemotherapy pharmacodynamics. Additionally, a pharmacodynamic study~\citep{math12081195} explores the effectiveness of the Verhulst and Montroll models in simulating tumor cell growth through the application of PINNs. In a recent work~\citep{CMINNs}, Compartment Model Informed Neural Networks (CMINNs) was proposed as a means of transforming compartmental modeling, with the objective of enhancing pharmacokinetic (PK) and integrated pharmacokinetic-pharmacodynamic (PK-PD) modeling. This approach incorporates fractional calculus and  time-varying parameters, combined with constant or 
piecewise-constant parameters, providing insights into drug dynamics in cancer and revealing the dynamics of cancer cell death in multi-dose administrations, which may exhibit resistance, persistence, and tolerance. Recent research demonstrates that using PINNs enables comprehensive mathematical modeling in systems toxicology, providing a deeper understanding of the effects of pharmaceutical substances on cardiac health~\citep{soukarieh2024hypersbinn}. The characterization of drug effects on cardiac electrophysiology by~\citep{chiu2024characterisation} using PINNs marks a pivotal development in predictive healthcare, enabling better management of arrhythmic disorders.

In the field of biomechanics, PINNs are revolutionizing diagnostics and treatment planning by seamlessly integrating physical laws with clinical data. PINNs have proven particularly effective in inferring blood flow dynamics from non-invasive imaging techniques, such as 4D flow MRI, enabling detailed analyses of cardiovascular conditions like intracranial aneurysms and stenosis~\citep{raissi2020hidden, kissas2020machine, sun2020surrogate, arzani2021uncovering, daneker2024transfer}. These networks are adept at handling noisy, sparse data, facilitating the calibration of conventional models, and predicting flow fields without extensive simulation data, thus enhancing diagnostic accuracy and efficiency~\citep{kissas2020machine, sun2020surrogate, arzani2021uncovering}. Moreover, PINNs contribute to the non-invasive inference of thrombus material properties, facilitating the estimation of permeability and viscoelastic moduli, which are vital for assessing treatment options~\citep{yin2021non}. 
This technology also enhances the quality of medical imaging through super-resolution and denoising of 4D flow MRI, making it more reliable for clinical applications~\citep{fathi2020super, gao2021super}. 
~\citep{liu2020generic} extended PINNs to model the mechanical behavior of soft biological tissues, providing crucial insights into tissue mechanics, which are essential for medical simulations and prosthetic design. Similarly,~\citep{lagergren2020biologically} utilized these networks to derive mechanistic insights from sparse experimental data, enhancing the modeling of biological phenomena. A novel method proposed in another study~\citep{SAINZDEMENA2024104092} employed PINNs to fit DCE-MRI data, incorporating contrast agent diffusion while ensuring compliance with mass conservation equations from the pharmacokinetic model, resulting in enhanced predictive accuracy.
Recent research has demonstrated the versatility of PINNs in modeling various aspects of tissue behavior, including thermal properties, biomechanical responses, and elasticity~\citep{Awojoyogbe2024,caforio2024physics, wu2024identifying}. Notably,~\citep{Ragoza2023} applied PINNs to tackle the inverse problem of tissue elasticity reconstruction in Magnetic Resonance Elastography, showcasing the potential of these methods for understanding and engineering biological tissues. Additionally,~\citep{movahhedi2023predicting} introduced a hybrid PINN that reconstructs 3D tissue dynamics from sparse 2D images by integrating fluid dynamics with soft tissue modeling for clinical diagnostics. 

Applications in cardiovascular engineering are also profound, as evidenced by studies on intraventricular flow mapping~\citep{ling2024physics}, cuffless blood pressure estimation~\citep{sel2023physics}, and modeling aortic blood flow~\citep{du2023evaluation} using PINNs.~\citep{jagtap2023coolpinns} demonstrated the utilization of PINNs in modeling active cooling within vascular systems, indicating the broad utility of PINNs in thermoregulation studies within biological contexts.
Overall, PINNs are setting new standards in medical diagnostics and personalized medicine by providing a robust framework that integrates computational models with real-world medical data~\citep{raissi2020hidden, kissas2020machine, sun2020surrogate, arzani2021uncovering, goswami2021study, yin2021non, ruiz2022physics, cai2021artificial, fathi2020super, gao2021super}.

PINNs have been adapted to address a range of epidemiological models, including the susceptibles-infected-recovered frameworks and their extensions. For instance, bi-objective optimization has been employed to train PINNs on the Susceptible-Vaccinated-Infected-Hospitalized-Recovered (SVIHR)  model, which includes compartments for leaky-vaccinated and hospitalized populations, demonstrating a sophisticated approach to managing the trade-off between data fit and model fidelity~\citep{heldmann2022pinn}.
In a practical application, PINNs were used to model COVID-19 infection and hospitalization scenarios based on data from Germany, proving their effectiveness in comparison to traditional finite difference methods~\citep{treibert2022physics}. A recent work~\citep{epiii} developed a PINNs model for estimating temporal changes in the SIR model to analyze transmission dynamics during outbreaks. Furthermore, the concept of Disease-Informed Neural Networks (DINNs)  extends PINNs to predict the spread of various infectious diseases, showcasing the flexibility of PINNs in handling epidemiological data~\citep{shaier2022data}.
The Susceptible-Exposed-Infected-Recovered (SEIR)  model has been augmented to incorporate environmental pathogen concentrations. PINNs have been employed to address both forward and inverse problems, thereby improving the model’s effectiveness in analyzing COVID-19 dynamics that involve intricate interactions between humans and pathogens~\citep{nguyen2022modeling}. Lastly, PINNs have been integrated with the Extreme Theory of Functional Connections (X-TFC)  to dynamically estimate parameters across several compartmental models, illustrating their potential in refining epidemiological predictions and parameter discovery~\citep{schiassi2021physics}. This study~\citep{kharazmi2021identifiability} explores multiple epidemiological models using PINNs to identify time-varying parameters and fractional differential operators. Various extensions of the classic SIR model, including fractional-order and time-dependent parameters, are examined. By applying these methods to COVID-19 data from New York, Rhode Island, Michigan, and Italy, the work simultaneously infers both unknown parameters and unobserved dynamics. The research highlights the identifiability of model parameters and uncertainties related to neural networks and control measures, offering insights into pandemic forecasting.
These applications demonstrate the capability of PINNs to provide a deeper understanding of disease dynamics, making them invaluable in the ongoing efforts to manage public health crises.

\begin{figure}[!ht]
    \centering
    \includegraphics[width=1 \linewidth]{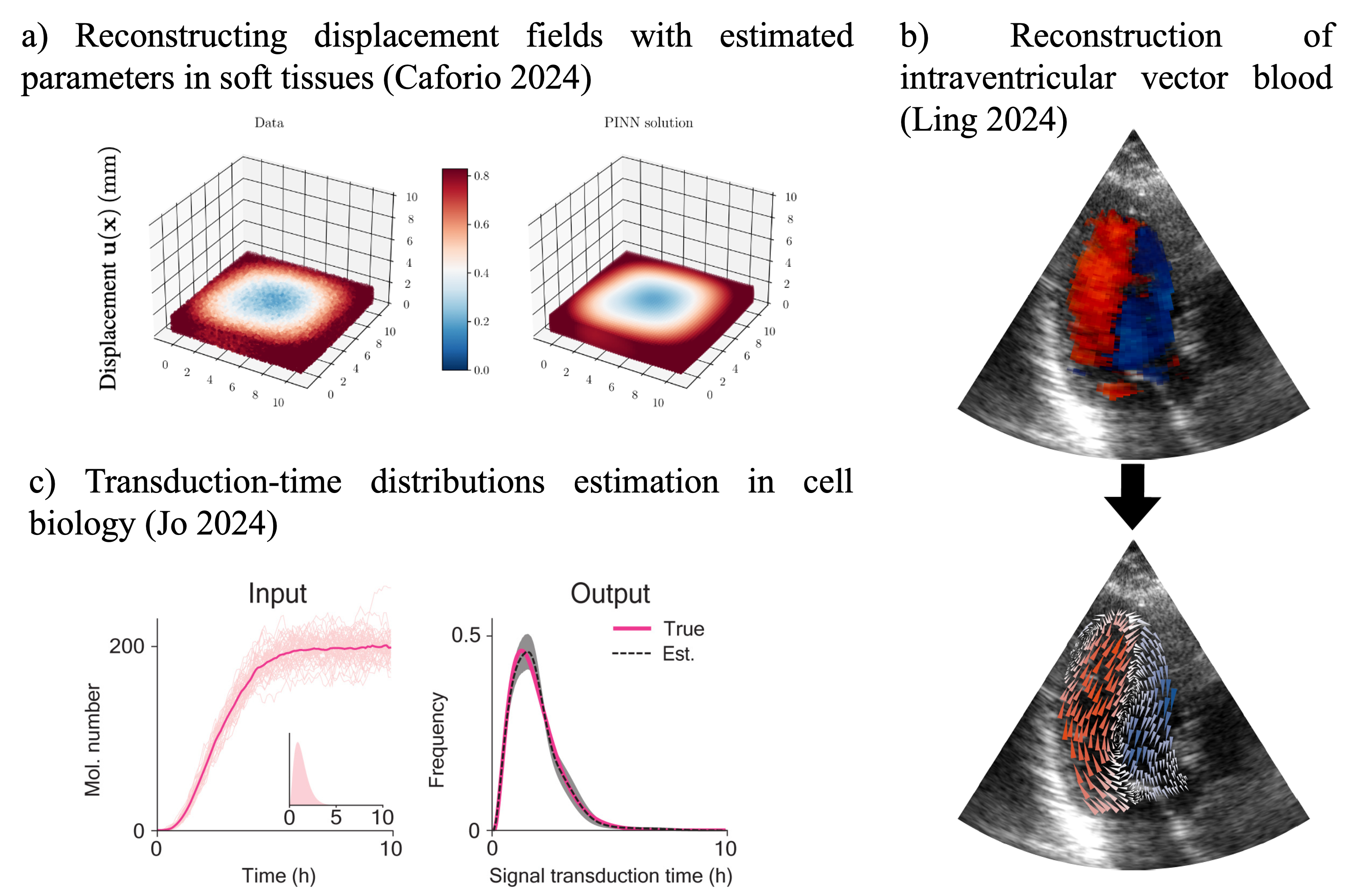}
    \caption{\textbf{Applications in Medicine.}
    PINNs are highly effective at reconstructing displacement fields, mapping intraventricular vector flows, and uncovering cell heterogeneity in signal transduction from sparse data. a)~\citep{caforio2024physics} introduce a novel method that combines PINNs with 3D nonlinear biomechanical models of soft tissue, enabling the reconstruction of displacement fields and the estimation of patient-specific biophysical properties, including stress and strain. b)~\citep{ling2024physics} propose an innovative approach to intraventricular vector flow mapping (iVFM) by replacing the traditional optimization method with a combination of PINNs and a physics-guided nnU-Net model, enhancing color Doppler analysis in cardiac imaging. c)~\citep{jo2024density} proposed Density-PINNs to infer transduction time distribution from observed final stress responses, providing insights into signal pathway dynamics such as speed and accuracy.}
    \label{fig:BME}
\end{figure}

\subsection{Applications in Mechanics}

\subsubsection{Fluid Mechanics}
PINNs  have made significant strides across a wide array of applications in fluid mechanics, showcasing their versatility and robust capability to solve complex problems by integrating physical laws with machine learning.

In imaging and flow visualization, PINNs have been pivotal in inferring complex fluid dynamics such as the 3D velocity and pressure fields from temperature data in tomographic background oriented Schlieren imaging, exemplified by studies like the flow over an espresso cup~\citep{cai2021flow}.~\citep{cai2024physics} reconstructs the velocity field in a turbulent jet from sparse observations available from Particle Tracking Velocimetry (PTV). This technique also extends to biological flows, where PINNs accurately reconstruct pressure fields around swimming fish from Particle Image Velocimetry (PIV) data, providing insights that surpass traditional methods~\citep{calicchia2022reconstructing}.

For turbulent and high-speed flows, PINNs address both forward and inverse problems. They handle high-speed flows by approximating the Euler equations, solving problems involving moving discontinuities and oblique waves, and inferring flow properties from sparse data in supersonic environments~\citep{mao2020physics, jagtap2022physics}. In two-dimensional turbulence, PINNs aid in predicting flow quantities and improving the understanding of small-scale turbulence~\citep{kag2022physics}.

PINNs also explore complex fluid behavior in non-Newtonian fluids, learning the viscosity from velocity measurements, which helps in modeling the flow between parallel plates—a challenging scenario for traditional models~\citep{reyes2021learning}. Similarly, in the context of vortex dynamics and scalar mixing, they provide new approaches to inferring lift and drag forces on bluff bodies and enhancing the understanding of scalar mixing in turbulent flows~\citep{raissi2019vortex, raissi2019deep}.

Further, PINNs have been instrumental in solving inverse problems such as reconstructing Rayleigh-Bernard flows from temperature data and modeling rarefied-gas dynamics under the Bhatnagar-Gross-Krook approximation, showing potential in regimes where data is sparse or incomplete~\citep{di2023reconstructing, de2021physics}. Their application extends to modeling viscoelastic materials, where the ViscoelasticNet framework aids in understanding the complex stress and pressure fields in fluids~\citep{thakur2022viscoelasticnet}.

In structural applications, PINNs facilitate the design of offshore structures by solving the Serre-Green-Naghdi equations for water waves, predicting future states of water surfaces and velocities crucial for engineering applications~\citep{jagtap2022deep}. They also enhance large-eddy simulations of fluid flows by incorporating physical priors into the modeling process, improving predictions across varying Reynolds numbers~\citep{yang2019predictive}.

Moreover, in acoustics, PINNs solve frequency-domain equations for isotropic media, avoiding the high computational costs associated with traditional methods and eliminating numerical dispersion~\citep{song2021solving}. Their utility in predicting fluid flow behaviors in complex geometries is also demonstrated in the modeling of thermal creep flows and vortex-induced vibrations~\citep{lou2021physics, erichson2019physics}.

\begin{figure}[!ht]
    \centering
    \includegraphics[width= \linewidth]{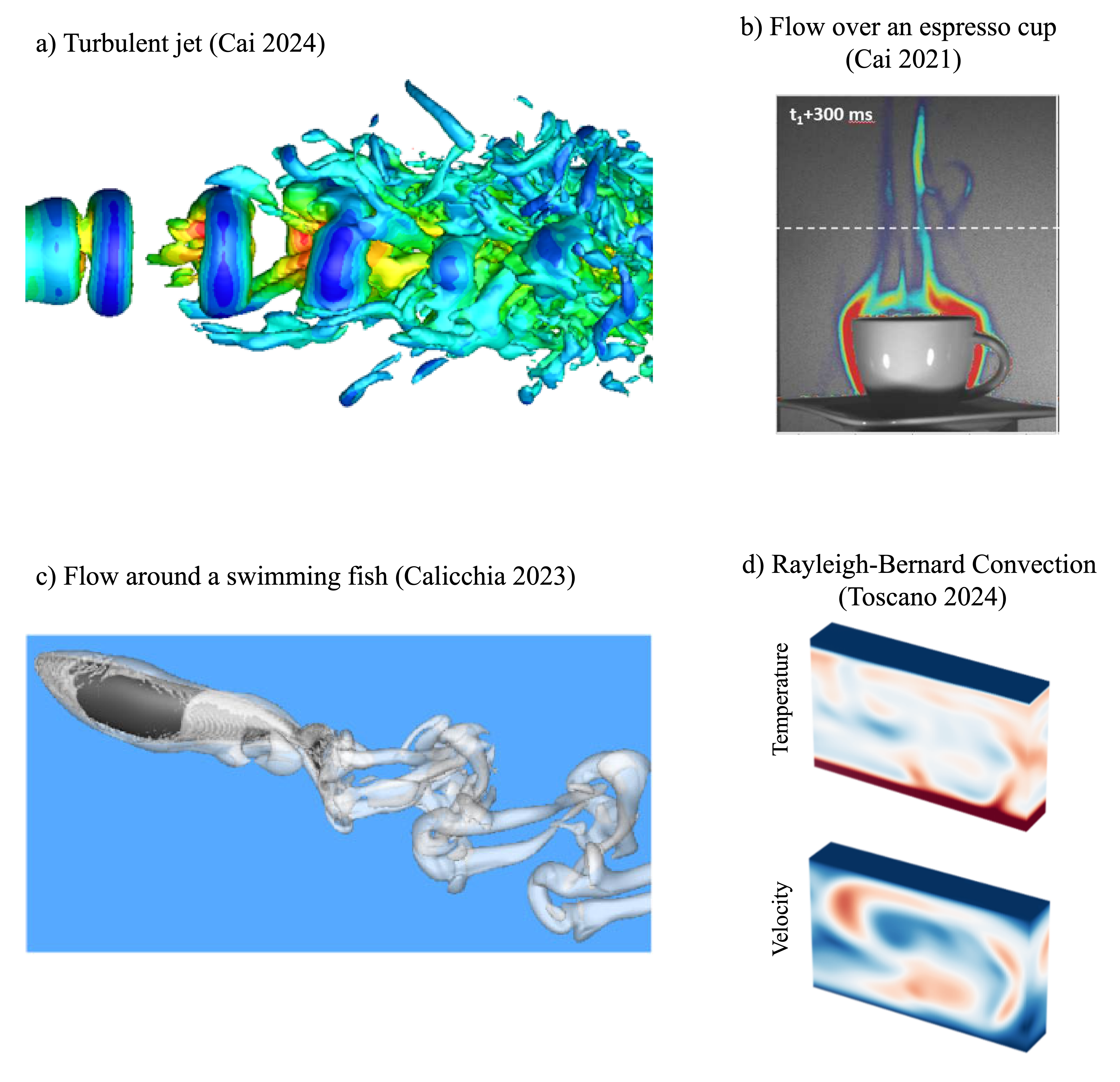}
    \caption{\textbf{Applications in Fluids.} PINNs effectively reconstruct velocity, pressure, and temperature fields from sparse observations.
    a)~\citep{cai2024physics} use PINNs to reconstruct the velocity field from sparse velocity observations from PIV/PTV of a turbulent jet. The iso-surfaces of the vorticity magnitude reconstructed by PINNs are shown here.
    In b)~\citep{cai2021flow} and c)~\citep{calicchia2023reconstructing}, PINNs is used to reconstruct the temperature and pressure fields around an espresso cup and a swimming fish, respectively. 
    In d)~\citep{toscano2024inferring} PIKANs are used to reconstruct the velocity and temperature fields in Rayleigh-Bernard convection from sparse observations.}
    \label{fig:fluids}
\end{figure}

\subsubsection{Solid Mechanics and Material Science}

PINNs address a diverse range of complex problems across different material behaviors and structural forms.
In the domain of nondestructive testing and evaluation, PINNs are effectively utilized to identify and characterize surface-breaking cracks in metal plates. By estimating the speed of sound, which varies due to the presence of cracks, PINNs facilitate precise crack detection from ultrasonic data, highlighting their potential in structural health monitoring~\citep{shukla2020physics}. Similarly, PINNs are deployed to quantify microstructural properties in polycrystalline nickel, solving inverse problems to infer material properties like compressibility and stiffness from ultrasonic data, demonstrating their capability in material characterization~\citep{shukla2021physics}.

For complex fluid behavior in materials, PINNs extend their application to model non-Newtonian fluids, accurately simulating fluids with time and deformation-dependent properties. This includes generalized Newtonian, viscoelastic, and thixotropic fluids, where PINNs recover velocity and stress fields effectively from sparse measurements~\citep{mahmoudabadbozchelou2022nn}. In the realm of solid mechanics, PINNs are applied to infer internal structures and defects in materials, accurately predicting the size, shape, and mechanical properties of voids or inclusions from stress-displacement data, demonstrating versatility across different material types~\citep{zhang2022analyses}.

The modeling of elastic plates using PINNs showcases a comparison between data-driven, PDE-based, and energy-based approaches, emphasizing the efficacy of PINNs in capturing finite deformations governed by complex equations like the Föppl–von Kármán equations~\citep{li2021physics}. Additionally, in the context of shell structures, PINNs solve for the small-strain response of curved shells, showing high accuracy when PDE loss is enforced in the weak form~\citep{bastek2022physics}.

PINNs also enhance the phase-field modeling of fracture, where they predict crack paths in materials by minimizing the variational energy of the system. Transfer learning is employed to improve efficiency, avoiding the need to retrain the network from scratch for each load step~\citep{goswami2020transfer}. Moreover, in the digital realm, PINNs predict the deformation of digital materials under load, applying energy-based formulations and innovative loss functions to prevent erroneous learning of deformation gradients~\citep{zhang2021physics}.

For complex nonlinear problems in computational mechanics, the Integrated Finite Element Neural Network (I-FENN)  combines PINNs with finite element methods to accelerate nonlinear solutions, showcasing the integration of machine learning with traditional numerical methods for enhanced computational performance~\citep{pantidis2022integrated}. Additionally, in elasticity imaging, PINNs are used for inverse identification of nonhomogeneous mechanical properties from displacement measurements under loading, demonstrating their utility in biomechanical applications~\citep{zhang2020physics}.

Lastly, Physics-informed multi-LSTM networks are introduced for the metamodeling of nonlinear structures. These networks incorporate physical laws into LSTM models to learn dynamics from limited data, significantly enhancing their learning efficiency and extrapolation capabilities~\citep{zhang2020physicsLSTM}.

Overall, these applications underscore the broad and impactful role of PINNs in advancing mechanics and material sciences, offering robust solutions to traditionally challenging problems across various material behaviors and structural analyses.

\begin{figure}[!ht]
    \centering
    \includegraphics[width= \linewidth]{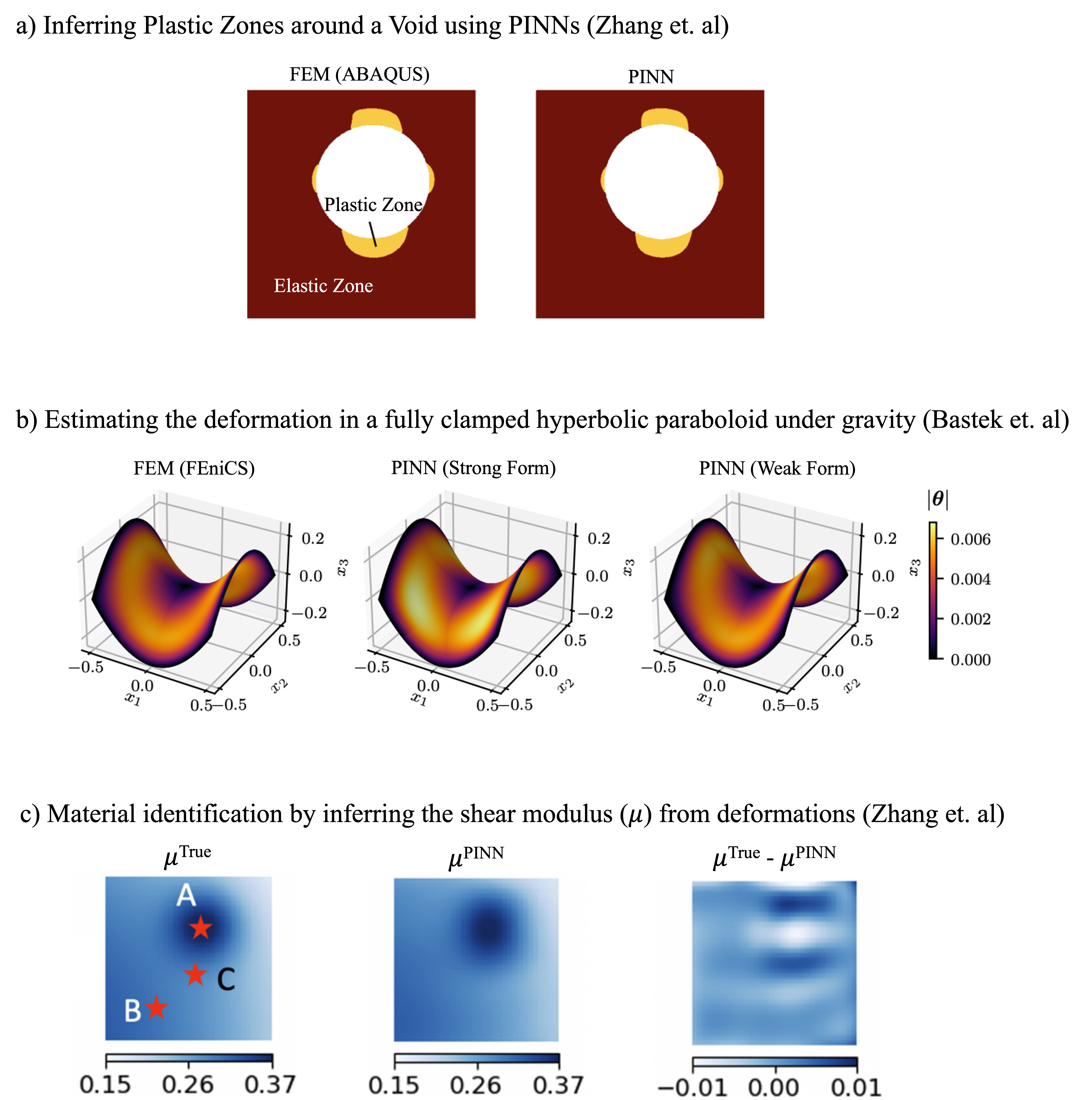}
    \caption{\textbf{Applications in mechanics and materials science.}
    a) Comparing the plastic zones around a defect in material estimated by PINNs and FEM~\citep{zhang2022analyses}.
    b)~\citep{bastek2022physics} uses PINNs based on strong and weak formulations to estimate the deformation in a fully clamped hyperbolic paraboloid subject to gravitational loading. 
    c) Material identification~\citep{zhang2020physics} by inferring the non-homogeneous shear modulus from the deformations. }
    \label{fig:materials}
\end{figure}

\subsection{Applications in Geophysics }

PINNs are demonstrating significant advancements in geophysics. PINNs effectively solve the eikonal equation for seismic wave traveltimes, enhancing applications like source localization and seismic inversion~\citep{bin2021pinneik}. This flexibility allows them to incorporate complex constraints like medium anisotropy and free-surface topography, which are challenging for traditional methods~\citep{ross2021hyposvi}. Additionally, in earthquake hypocentre inversion, PINNs, combined with Stein Variational Inference, rapidly approximate posterior distributions, showing effectiveness with real seismic data~\citep{ross2021hyposvi}.

In compositional modeling, PINNs perform flash calculations to determine phase composition, achieving significantly lower error rates than traditional deep neural networks by adhering to thermodynamic constraints~\citep{ihunde2022application}. For hydrological modeling, PINNs act as surrogate models for water flows in river channels, efficiently handling inverse parameter estimation and outperforming conventional models~\citep{nazari2022physics}. Moreover, they tackle complex wave propagation and full waveform inversion problems, automatically satisfying absorbing boundary conditions and providing efficient solutions compared to conventional solvers~\citep{rasht2022physics}.

In geostatistical modeling, a physics-informed semantic inpainting approach using PINNs incorporates indirect measurements into groundwater flow models, demonstrating how PINNs can enhance the understanding of subsurface phenomena~\citep{zheng2020physics}. These diverse applications highlight the broad potential of PINNs to advance geophysical investigations, offering more accurate, flexible, and efficient methodologies for understanding and predicting geological and hydrological processes.

PINNs have proven to be highly effective in addressing complex transport problems in porous media, bringing significant advances across various applications. PINNs are adept at solving partial PDEs related to hyperbolic transport problems such as the Buckley-Leverett and Burgers equations, demonstrating accuracy comparable to traditional numerical methods like Lagrangian-Eulerian and Lax-Friedrichs~\citep{abreu2021study}. In modeling multiphase flow through porous media, such as the drainage of gas in water-saturated media, PINNs not only estimate water saturation effectively but also solve inverse problems to infer flow parameters, showing superior performance over traditional data-driven methods. especially when data is sparse~\citep{almajid2022prediction}.

Further expanding their application, PINNs have been employed to infer key hydraulic properties and transport phenomena in subsurface transport scenarios. They outperform classical neural networks by accurately predicting hydraulic conductivity and concentration fields using coupled equations like Darcy's law and the advection-dispersion equation~\citep{he2020physics, tartakovsky2020physics}. These networks handle both saturated and unsaturated flow conditions efficiently, often surpassing state-of-the-art methods in parameter estimation from noisy data.

PINNs also address the coupled flow and deformation in porous media, solving complex poroelastic problems using stress-split sequential training methods. 
They efficiently tackle benchmark problems in poroelasticity, proving their robustness in simulations involving both single-phase and multiphase flows~\citep{haghighat2022physics}. 
In non-isothermal multiphase poromechanics, PINNs facilitate inverse modeling, identifying parameters in thermo-hydro-mechanical processes, and applying these capabilities to classic consolidation and injection-production problems~\citep{amini2022inverse}.

Overall, PINNs represent a transformative approach in the study of porous media transport, offering a powerful tool for simulating and understanding the complex interactions within porous materials under various physical conditions.
Their ability to integrate and solve multiphysics problems through a unified framework highlights their potential to reshape traditional approaches in geosciences and engineering~\citep{amini2022physics}.

\begin{figure}[!ht]
    \centering
    \includegraphics[width= \linewidth]{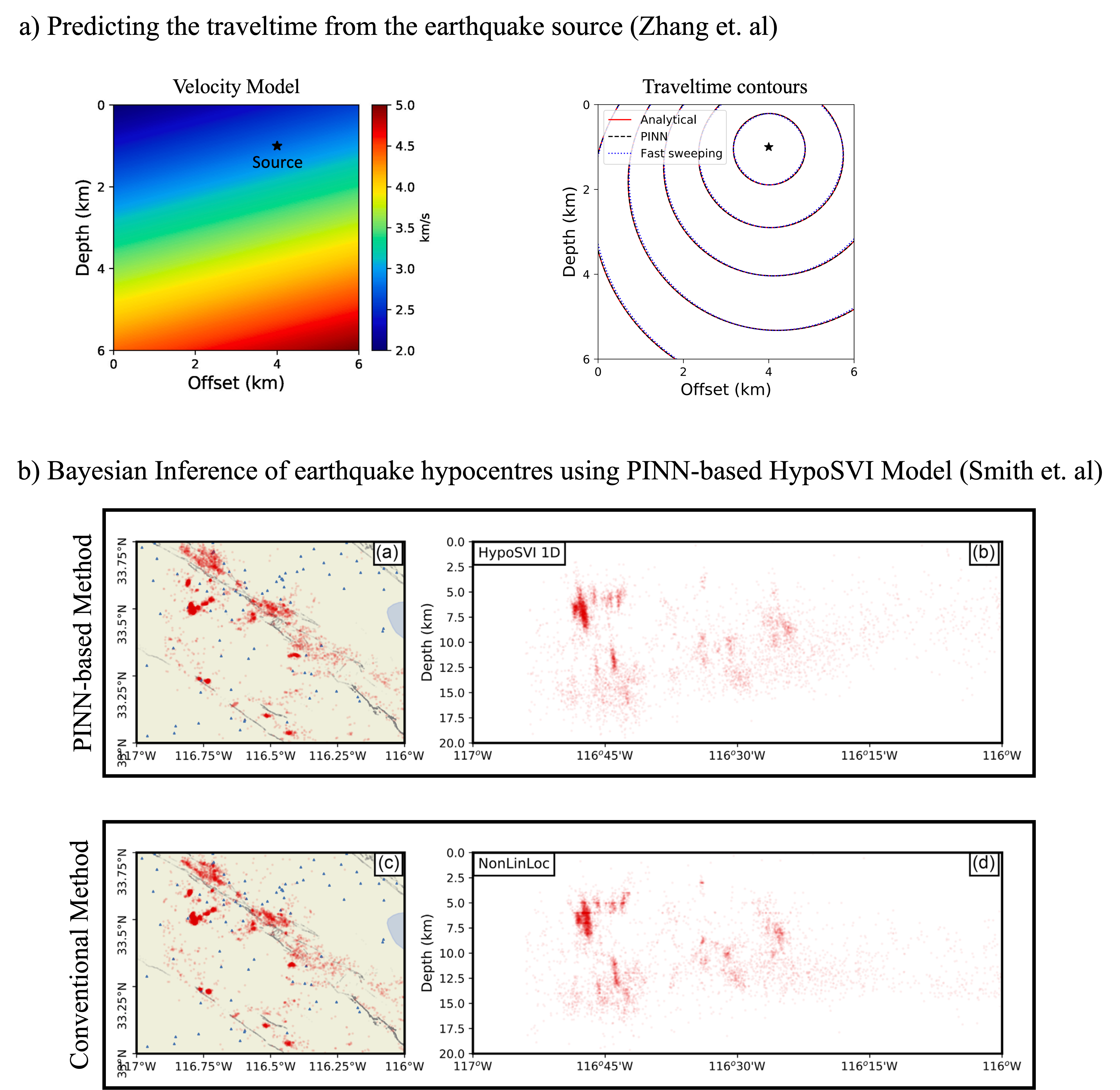}
    \caption{\textbf{Applications in Geophysics.}
    a) Comparing the plastic zones around a defect in material estimated by PINNs and FEM~\citep{zhang2022analyses}.
    b) The earthquake hypocentres are estimated using HypoSVI that blends PINNs with Stein Variational Inference~\citep{ross2021hyposvi}. }
    \label{fig:geophysics}
\end{figure}

\subsection{Applications in Dynamical systems, Control and Autonomy}

PINNs are making impactful advancements in the fields of dynamical systems, control, and autonomy, providing innovative solutions to complex modeling and control challenges across various applications.

\begin{figure}[!ht]
    \centering
    \includegraphics[width= \linewidth]{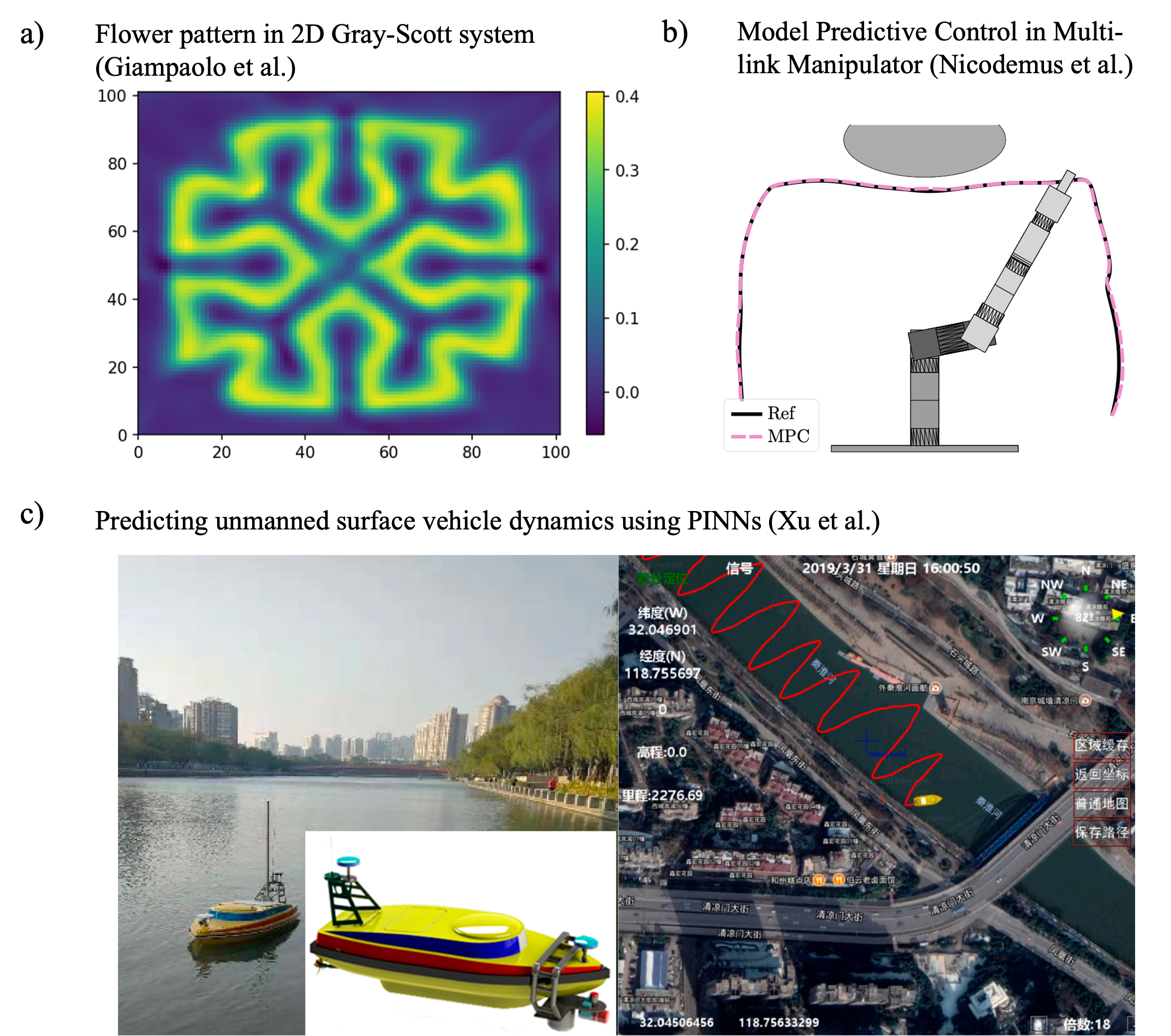}
    \caption{\textbf{Applications in dynamical systems, control, and autonomy:} (a)  Flower-shaped Turing pattern obtained by solving the 2D Gray-Scott system using PINNs~\citep{giampaolo2022physics}. (b)  The reference trajectory and the PINN-based model predictive control trajectory for the tracking problem in a multi-link manipulator~\citep{nicodemus2022physics}. (c)  The image on the left shows a picture of the Deepsea Warriors Uboat (DW-Uboat) in the Qin Huai River, China, from the experiments conducted by~\citep{xu2022physics}. The inset shows a profile of the DW-Uboat. A PINN is used to predict the surge, sway, and rotation velocities of the unmanned surface vehicle using the data collected from zig-zag trajectories (red curve in the image on the right). }
    \label{fig:dynamical_sys}
\end{figure}

In dynamical systems, PINNs are effectively used to solve and understand the behavior of complex models like the 1D and 2D Gray-Scott systems, overcoming challenges related to local minima in the loss function by integrating data from finite element methods~\citep{giampaolo2022physics}. For control applications, PINNs have been extended to handle control variables and enhance extrapolation capabilities, as demonstrated in controlling Van der Pol and four-tank systems, showing significant improvements over traditional PINNs~\citep{antonelo2021physics}.
PINNs are applied to predict and control unmanned surface vehicle dynamics, integrating real trajectory data to enhance model generalization and performance~\citep{xu2022physics}. Furthermore, PINNs are integrated into robust adaptive models predictive control frameworks, such as RAMP-Net, which significantly reduces tracking errors in quadrotor flight by encoding both known physics and data-driven insights~\citep{sanyal2022ramp}. PINNs have also been used to learn the dynamics in quadrotors, demonstrating a better generalization capacity compared to linearized mathematical models~\citep{gu2024physics}

PINNs have also been effectively used for control applications. For example,~\citep{liu2024physics} combines a dynamics model learned using PINNs with model-predictive control for motion prediction and trajectory tracking for rigid and soft continuum robots. Additionally, PINNs are used for nonlinear model predictive control in multi-link manipulators, where they efficiently approximate nonlinear dynamics, outperforming traditional numerical methods in terms of speed and accuracy~\citep{nicodemus2022physics}. The work by~\citep{wang2024pinn} uses PINNs to model complex deformation in a soft robotic gripper with data assimilation, achieving better accuracy when compared to FEM. Furthermore, the paper by~\citep{ni2023progressive} introduces a PINN-based framework that progressively improves the efficiency of motion planning in dynamic environments, enhancing both adaptability and computational performance in real-time applications. In car-following models, PINNs leverage existing physics-based models to enhance prediction accuracy, outperforming traditional data-driven methods, particularly in sparse data scenarios~\citep{mo2021physics}. Lastly, in risk-aware autonomous driving, PINNs model complex wheel-ground interactions and utilize latent features to develop advanced control frameworks, improving performance under variable conditions~\citep{kim2022physics}.

These applications highlight the versatility of PINNs in adapting to and solving multifaceted problems in dynamical systems, control, and autonomy. They showcase their potential to transform traditional approaches with enhanced predictive and control capabilities across various engineering and technological domains.

\subsection{Applications in Heat Transfer}

PINNs are proving to be transformative in handling heat transfer problems across diverse applications, demonstrating their effectiveness in both simulation and optimization. NVIDIA's SimNet framework exemplifies this by solving various multiphysics problems, such as optimizing heat sink designs for electronic components and modeling blood flow dynamics in medical applications~\citep{hennigh2021nvidia}. In traditional heat transfer contexts, PINNs solve forced and mixed convection problems, even when thermal boundary conditions are unknown, which is crucial for real-world applications like electronic chip cooling and heat sink efficiency~\citep{cai2021physics}.

In metal additive manufacturing, PINNs are utilized to predict the temperature and melt pool dynamics, enhancing the understanding and control of manufacturing processes under limited data conditions~\citep{zhu2021machine}. Moreover, in advanced manufacturing scenarios involving convective heating, PINNs offer solutions that integrate convection boundary conditions, outperforming conventional finite element methods in both speed and adaptability~\citep{zobeiry2021physics}.
Additionally, PINNs are shown to be effective in modeling microscale heat conduction in double-layered thin films exposed to ultrashort-pulsed lasers~\citep{bora2021neural, bora2022neural}, extensively used in thermal processing of materials, structural monitoring of thin metal films and laser processing in thin-film deposition.

PINNs also address the thermochemical curing processes in composite-tool systems, modeling the intricate dynamics of temperature and chemical transformations during manufacturing, which are critical for optimizing product quality and material properties~\citep{niaki2021physics}. These applications highlight the extensive potential of PINNs to revolutionize heat transfer modeling, providing robust, efficient, and highly accurate solutions that integrate seamlessly with engineering workflows and lead to improved designs and processes in a variety of industrial contexts.

\subsection{Applications in Physics}

 PINNs are facilitating advancements in fundamental physics. 
 In aerodynamics-thermodynamics, PINNs handle hyperbolic systems with shocks better through a new space-time control volume scheme, addressing traditional challenges in shock-related problems~\citep{patel2022thermodynamically}.
 PINNs have been effectively used to model plasma dynamics in magnetic confinement fusion devices, revealing electric potentials and fields from sparse electron data by solving the drift-reduced Braginskii equations~\citep{mathews2021uncovering}. Another application involves reconstructing 3D structures from 2D diffraction patterns, demonstrating real-time capabilities that outperform traditional numerical methods~\citep{stielow2021reconstruction}.
 In magnetostatics and micromagnetics, PINNs solve both forward and inverse problems, enhancing our understanding and control over magnetic materials~\citep{kovacs2022magnetostatics}.
PINNs have also been employed in the study of electrochemistry for predicting voltammetry across various configurations, solving both forward and inverse problems in cycling voltammetry~\citep{chen2022predicting}.

Furthermore, PINNs are adept at solving the time-dependent Schrödinger equation, providing insights into quantum dynamics~\citep{shah2022physics}. They are also utilized in solving the nonlinear Schrödinger equation, accurately predicting rogue waves, and parameter discovery~\citep{wang2021data}.
One notable study employed an improved PINN method to tackle the derivative nonlinear Schrödinger equation, showcasing the versatility of PINNs in obtaining detailed wave solutions~\citep{pu2021solving}. Lastly, PINNs are used in multiscale mode-resolved phonon transport, optimizing thermal management in microelectronics by solving the phonon Boltzmann transport equation efficiently and accurately~\citep{li2021physicsphonon}.

\begin{figure}[!ht]
    \centering
    \includegraphics[width=\linewidth]{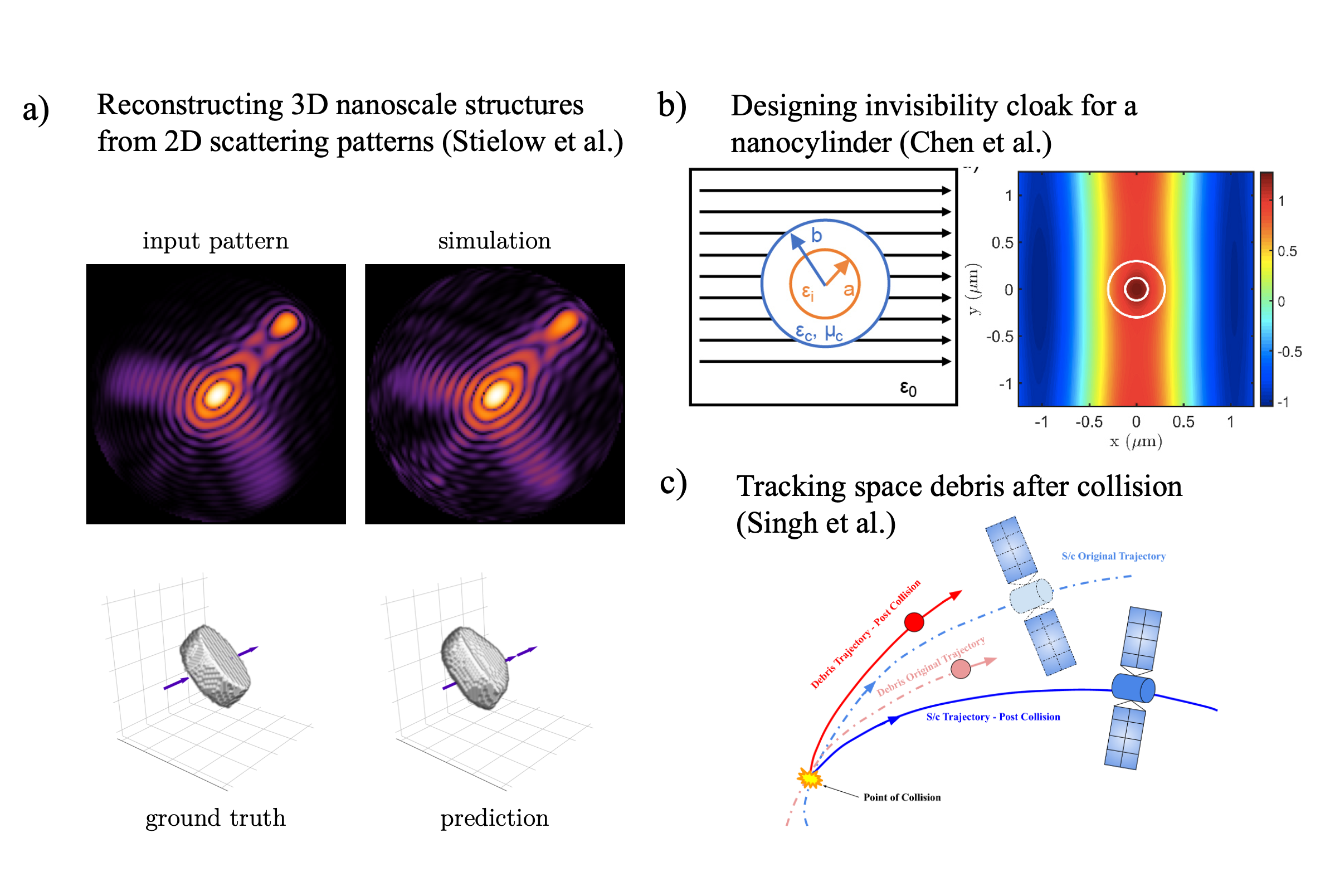}
    \caption{\textbf{Applications in physics:} (a) Input 2D wide-angle scattering pattern, the reconstructed \& ground truth nanoscale structure and the simulated scattering pattern using the predicted structure~\citep{stielow2021reconstruction}. (b)  A schematic from~\citep{singh2024tracking}, where a PINN is used to track the trajectory of space debris after inelastic collision with a satellite. (c)  A schematic of a nanocylinder with constant permittivity coated with a cloaking material to zero out the scattering~\citep{chen2020physics}. The corresponding electric field distribution with the coating layer permittivity predicted using a PINN is also shown.}
    \label{fig:phys_app}
\end{figure}
PINNs enable the reconstruction of electric permittivity distributions from synthetic scattering data in nano-optics and metamaterials, effectively addressing challenges such as cloaking in nanocylinders~\citep{chen2020physics}. In fiber optics, PINNs successfully model nonlinear dynamics governed by the nonlinear Schrödinger equation, enhancing capabilities in optical fiber communications through precise modeling of dispersion and nonlinear effects~\citep{jiang2022physics}.
Additionally, PINNs are used to predict dynamic processes and parameters for vector optical solitons in birefringent fibers, showcasing their ability to perform inverse estimations of dispersion coefficients and nonlinearity coefficients with resilience to noise~\citep{wu2021predicting}. In the context of fiber-optic communication systems, PINNs improve digital backpropagation by incorporating trainable filters, reducing computational complexity while enhancing performance, thus aligning well with practical digital signal processing applications~\citep{hager2020physics}. These examples highlight the transformative impact of PINNs in advancing fundamental research in physics, providing innovative solutions that integrate physical laws with machine learning techniques.

In the study of Earth’s radiation belts, PINNs have been adeptly used to solve the inverse problem associated with the Fokker-Planck equation, using data from the Van Allen Probes to enhance our understanding of electron transport mechanisms~\citep{camporeale2022data}. Additionally, PINNs play a critical role in optimizing planar orbit transfers by learning optimal control actions, adhering to the conditions set by the Pontryagin minimum principle, thus offering a promising avenue for space navigation~\citep{schiassi2022orbit}.
Further applications include the development of closed-loop optimal guidance and control policies in aerospace systems, where PINNs integrate with the Hamilton-Jacobi-Bellman equation and the theory of functional connections to tackle complex control problems in Newtonian mechanics~\citep{furfaro2022physics}. Similarly, PINNs have been used alongside the Extreme Theory of Functional Connections (X-TFC)  to solve constrained optimal control problems, proving effective in scenarios like the constrained minimum time-energy optimal Halo-Halo orbit transfer and fuel optimal landing problems~\citep{d2021physics}. Furthermore, PINNs have been used to predict the trajectory of untracked space debris after inelastic collision with a satellite~\citep{singh2024tracking}.

Reinforcement learning applications in space missions are enhanced by PINN-based gravity models, which simulate environmental dynamics more rapidly than traditional methods, aiding in the development of safer spacecraft behaviors around irregularly shaped bodies and facilitating the discovery of periodic orbits~\citep{martin2022reinforcement}. This approach is extended in gravity field modeling, where PINNs tackle challenges in modeling gravity fields of small celestial bodies, significantly improving robustness to noise and modeling accuracy~\citep{martin2022physics}.
The discovery of periodic orbits is further optimized through PINNs, which reduce search times dramatically by integrating with orbital element space computations, a shift from conventional Cartesian space methods~\citep{martin2022periodic}. Lastly, in the field of radiative transfer, PINNs are employed to solve both forward and inverse problems, enabling the estimation of absorption coefficients and radiative intensity with enhanced accuracy, also providing insights into the generalization error associated with such models~\citep{mishra2021physics}.
These applications demonstrate the transformative potential of PINNs in astronomy and aerospace engineering, providing sophisticated tools for navigating, modeling, and understanding space phenomena and engineering challenges.

\subsection{Applications in Chemical Engineering}

Chemical engineering applications involve a broad range of subjects, including transport phenomena, heat transfer, optimal control, species separation, unit operations, thermodynamics, and reaction kinetics. The understanding of these areas is grounded in the fundamental conservation laws of momentum, energy, and mass, as well as reaction kinetics, such as the Arrhenius equation and rate laws, which quantify reactant consumption and product formation rates. The most precise representation of these conservation laws and reaction kinetics is achieved through PDEs. As multiple studies demonstrated the success of PINNs in modeling multiphysics systems, PINNs have become valuable tools in dealing with chemical engineering problems. Wu et al.~\citep{wu2023application} provide a comprehensive overview of PINNs' applications in chemical engineering, including process modeling, surrogate construction, and design optimization. Zhu et al.~\citep{zhu2022review} review the current use of machine learning in chemical engineering and categorize PINNs as domain knowledge-informed machine learning models, emphasizing their unique capability to integrate prior knowledge of physics, mathematical laws, chemical mechanisms, and boundary conditions as constraints. Since previous sections have covered PINNs in transport and heat transfer, this section focuses on their applications in mass transfer, chemical reactions and reactor design, separation and unit operations, and surrogate model development within chemical engineering.

Mass transfer is central to many separation and reaction processes. Chen et al.~\citep{chen2022predicting} applied PINNs to model 1D and 2D cyclic voltammetry in electrochemical systems, incorporating mass diffusion equations and electrochemically consistent boundary conditions. Batuwatta-Gamage et al.~\citep{batuwatta2023novel} proposed physics-informed neural networks for mass transfer (PINN-MT) to effectively predict cellular-level mass loss and subsequent moisture variations during low-temperature drying. Additionally, Xuan et al.~\citep{xuan2023physics} combined physics-informed learning with an optimization method to predict mass flow rate, pressure, and velocity in refrigerant filling, further demonstrating the effectiveness of PINNs in mass transfer modeling. 

PINNs have been used to address various challenges in chemical reactions and reactor design, focusing on several key aspects. One important application is handling stiff chemical kinetics, as demonstrated by Ji et al.~\citep{ji2021stiff}, De et al.~\citep{de2022physics}, and Weng et al.~\citep{weng2022multiscale}. Additionally, PINNs are applied to identify missing or unknown kinetic information in complex reaction models. For instance, Ngo et al.~\citep{ngo2021solution} used inverse PINN methods to determine the effectiveness factor in a nonlinear reaction rate model for catalytic $CO_{2}$ methanation in an isothermal fixed-bed reactor, while Cohen et al.~\citep{cohen2024data} applied symbolic regression with PINNs to derive kinetic models in dynamic plug flow reactors. Bibeau et al.~\citep{bibeau2024physics} also employed PINNs to predict the kinetics of biodiesel production in microwave reactors. Moreover, PINNs have also been applied to model complex convection-diffusion-reaction systems. Hou et al.~\citep{hou2023pinn} used PINNs to solve a variety of cases, such as gas-solid adsorption, forward-reacting flows under different Péclet numbers, and inverse problems with missing chemical information. Sun et al.~\citep{sun2023physics} introduced PINNs to solve two-dimensional convection-diffusion-reaction equations for nonlinear reacting flows. 
PINNs have been applied to a wide range of reactor types and applications. For continuous flow reactors, such as plug flow reactors (PFRs) and continuous stirred-tank reactors (CSTRs), Choi et al.~\citep{choi2022physics} assessed the feasibility of using PINNs to model a CSTR with a van de Vusse reaction, while Patel et al.~\citep{patel2023optimal} utilized PINNs to optimize temperature trajectories in PFRs. Ngo et al.~\citep{ngo2022forward} developed forward PINNs based on the one-dimensional design equations for PFRs. Furthermore, PINNs have been applied to nuclear reactor design, as demonstrated by Elhareef et al.~\citep{elhareef2023physics} and Schiassi et al.~\citep{schiassi2022physics}.

Modeling multiphysics problems in chemical engineering using PDEs often leads to high degrees of nonlinearity and nonconvexity, which increases the complexity of the optimization process. To address this, surrogate models are used to balance model accuracy with computational efficiency. Many studies have made initial attempts to apply PINNs to construct surrogates of complex systems in chemical engineering. 
For example, Liu et al.~\citep{liu2023multi} proposed a multi-fidelity surrogate modeling method based on PINNs, combining high-fidelity simulation data with low-fidelity governing equations described by differential equations. This approach was applied to simulate the startup phase of a CSTR, demonstrating strong extrapolation performance. Antonello et al.~\citep{antonello2023physics} presented the use of PINNs as surrogate models for accidental scenarios simulation in Nuclear Power Plants (NPPs). Liu et al.~\citep{liu2023surrogate} developed a field-resolving surrogate modeling framework using PINNs for a parameterized two-dimensional methane-air jet combustion system.

PINNs have also been used in species separation and unit operations. For example, they have been applied to model chromatography columns~\citep{zou2024parameter,soderstrom2022physics,tang2023physics}. Li et al.~\citep{li2024unit} examined the predictive capabilities and generalizability of PINNs in modeling unit operation processes (UOPs) for urban water treatment, including continuous stirred-tank reactors, activated sludge reactors, and fixed-bed granular adsorption reactors.

\begin{figure}[!ht]
    \centering
    \includegraphics[width=\linewidth]{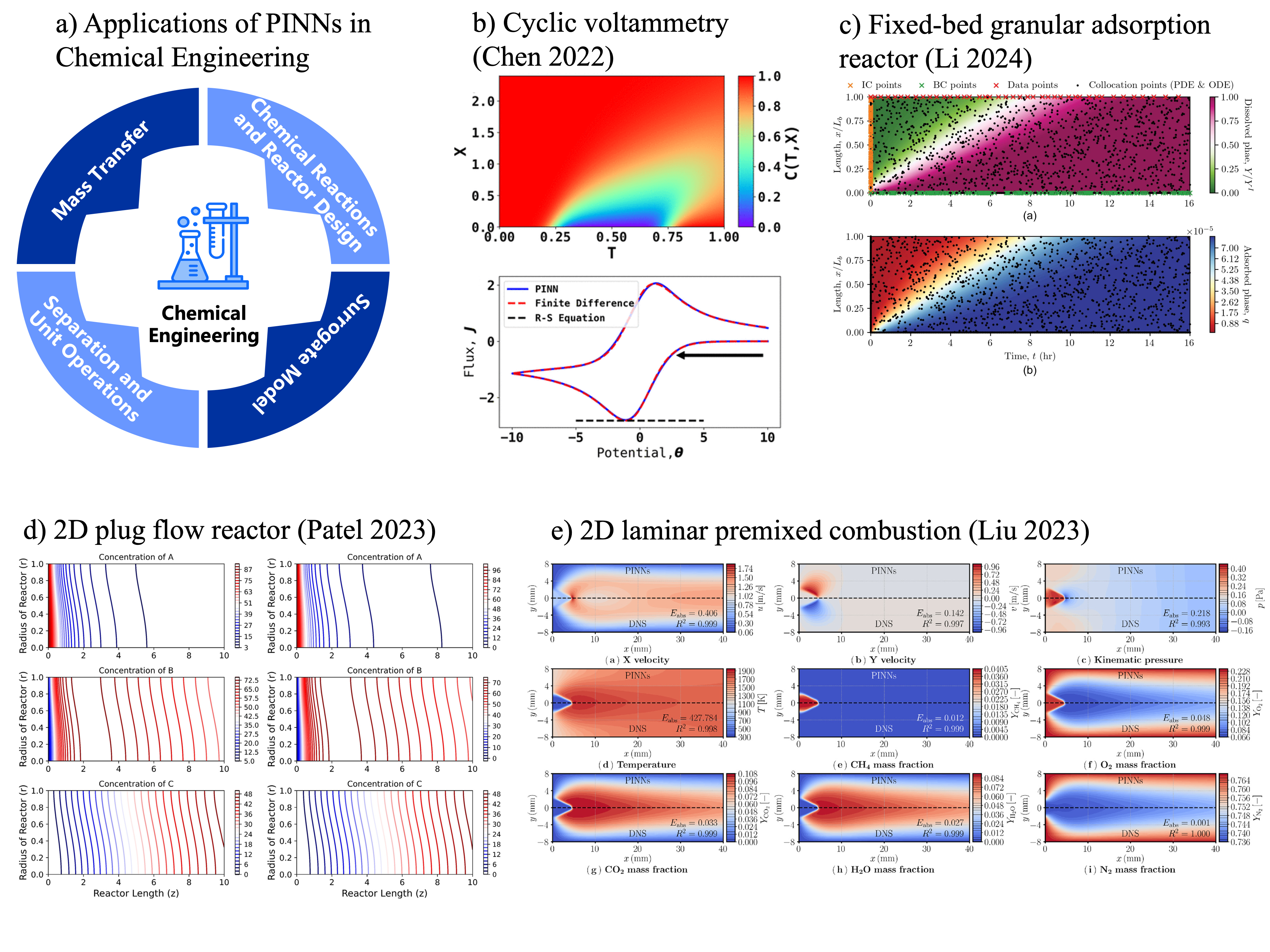}
    \caption{\textbf{Applications in chemical engineering:} (a) Applications of PINNs in Chemical Engineering. (b)   Prediction results of PINNs for cyclic voltammetry~\citep{chen2022predicting}. Temporal spatial concentration proﬁle (top) and the voltammogram predicted by PINNs (blue line) compared with the voltammogram generated (bottom).
    (c) Prediction of PINNs in fixed-bed granular adsorption reactor~\citep{li2024unit}. (d) Concentration profile of the species along the radius and length of the reactor as predicted by PINNs (right) and FDM (left) at temperature 370 K~\citep{patel2023optimal}. (e) PINNs prediction (upper half) and the DNS results (bottom half) in 2D laminar premixed combustion~\citep{liu2023surrogate}.}
    \label{fig:chem_engg}
\end{figure}

\subsection{Miscellaneous Applications}

In addition to the applications mentioned above, PINNs have also been utilized in diverse fields such as finance, topology optimization, porous media and environmental engineering, among others. The finance industry utilizes improved PINNs for more stable and accurate modeling of complex financial instruments like options under the Black-Scholes model, showcasing the potential of these networks to revolutionize financial analytics and risk management~\citep{bai2022application}.
In the domain of optimization and design, PINNs facilitate the design of electromagnetic metamaterials by solving Maxwell's equations and determining material properties like permeability and permittivity and are therefore crucial for the design of devices such as cylindrical cloaks and rotators~\citep{fang2019deep}. They also facilitate topology optimization in holography and fluid dynamics, where hard constraints are effectively managed through advanced methodologies like the penalty and augmented Lagrangian methods~\citep{lu2021physics}.

In molecular simulations, PINNs enable the prediction of material properties from molecular dynamics simulations and, therefore, provide insights into nanoscale phenomena~\citep{islam2021extraction}. The insights into material behavior gained through PINNs are crucial for sectors ranging from pharmaceuticals to nanotechnology. The power systems sector also benefits from PINNs, which facilitate the modeling of complex grid dynamics, allowing for faster and more accurate prediction of states such as rotor angles and system frequencies~\citep{misyris2020physics}.
In renewable energy, specifically wind power estimation, PINNs provide a method to optimize turbine layout and performance prediction, contributing significantly to the efficiency of wind farms~\citep{park2019physics}.
Environmental engineering applications, particularly in managing natural disasters such as wildfires, benefit from PINNs through enhanced data assimilation and modeling capabilities, which improve the accuracy of predictions and the effectiveness of response strategies~\citep{dabrowski2022bayesian}. PINNs also address challenges in physical domains with irregular geometries, solving complex PDEs in fields as varied as heat transfer, fluid dynamics, and electromagnetism, which are crucial for advancing engineering solutions~\citep{gao2021phygeonet}.

Furthermore, the ability of PINNs to model discontinuities and high-gradient phenomena also opens new possibilities in aerospace engineering, seismology, and meteorology, where accurate modeling of such phenomena is essential~\citep{liu2022discontinuity}.
Additionally, cusp-capturing PINNs~\citep{tseng2023cusp} introduce a cusp-enforced level set function to solve discontinuous-coefficient elliptic interface problems whose solution is continuous but has discontinuous first derivatives on the interface.

PINNs also contribute to the understanding and prediction of behaviors in systems governed by nonlinear diffusivity and Biot's equations, impacting fields such as geotechnical engineering, biomedical applications, and energy systems~\citep{kadeethum2020physics}.


\section{Uncertainty Quantification}
\label{uncertainty_quantification}

Uncertainty quantification (UQ) is essential for the reliable and trustworthy deployment of PINNs in solving differential equations for forward and inverse problems. Conventionally, in machine learning and data science, uncertainty is categorized into \textit{aleatoric} uncertainty and \textit{epistemic} uncertainty~\citep{abdar2021review, psaros2023uncertainty}. Aleatoric uncertainty (or data uncertainty)  arises from inherent variability in the data itself. This type of uncertainty is due to random noise or variability in the data distribution and is considered irreducible because it is not related to any lack of knowledge or flaws in the model but instead is an intrinsic property of the data. Epistemic uncertainty (or knowledge/model uncertainty) is due to a lack of knowledge about the underlying process or model. 
\begin{figure}[H]
    \centering
    \subfigure[Sources of uncertainty~\citep{psaros2023uncertainty}.]{
        \includegraphics[width=0.4\textwidth]{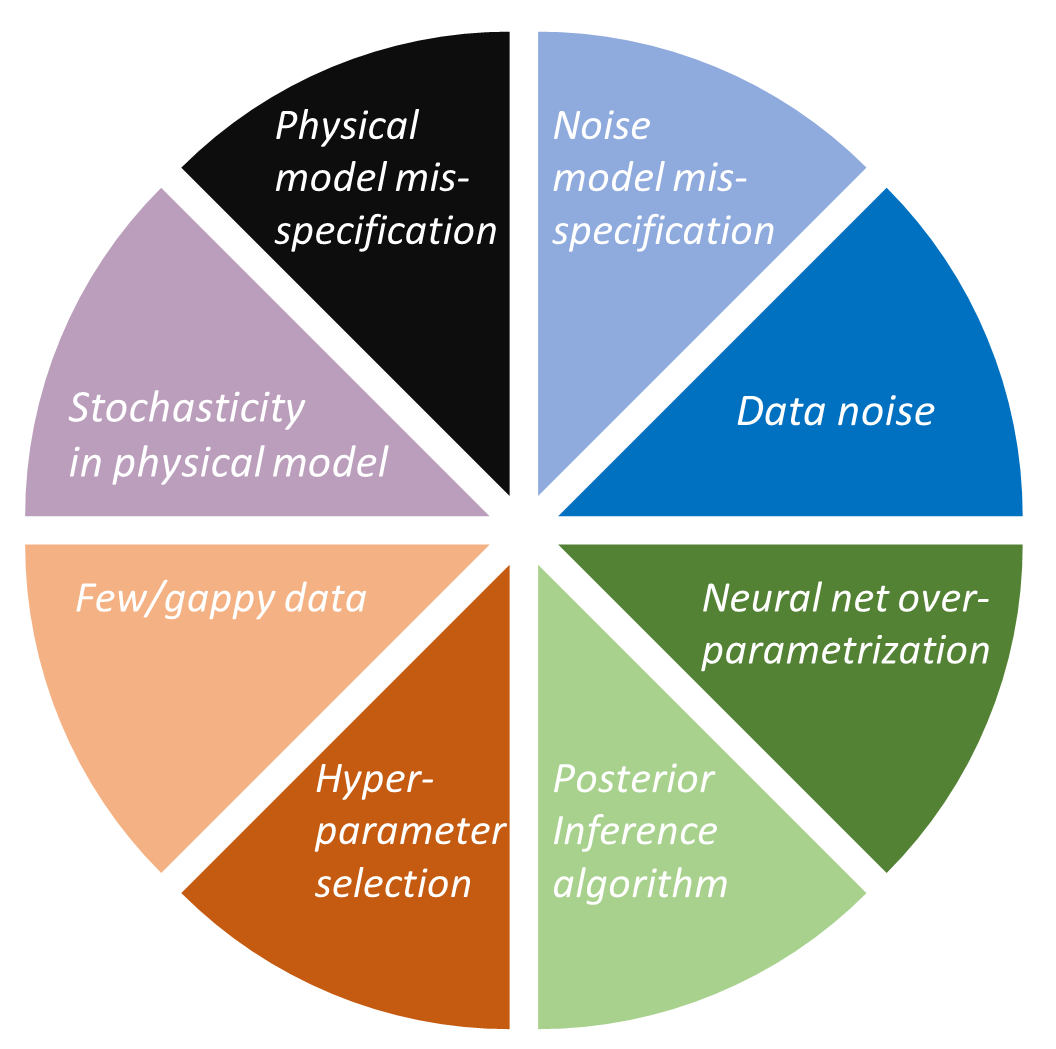}
    }
    \subfigure[Regressing noisy and gappy data~\citep{yang2021b}.]{
        \includegraphics[width=0.3\textwidth]{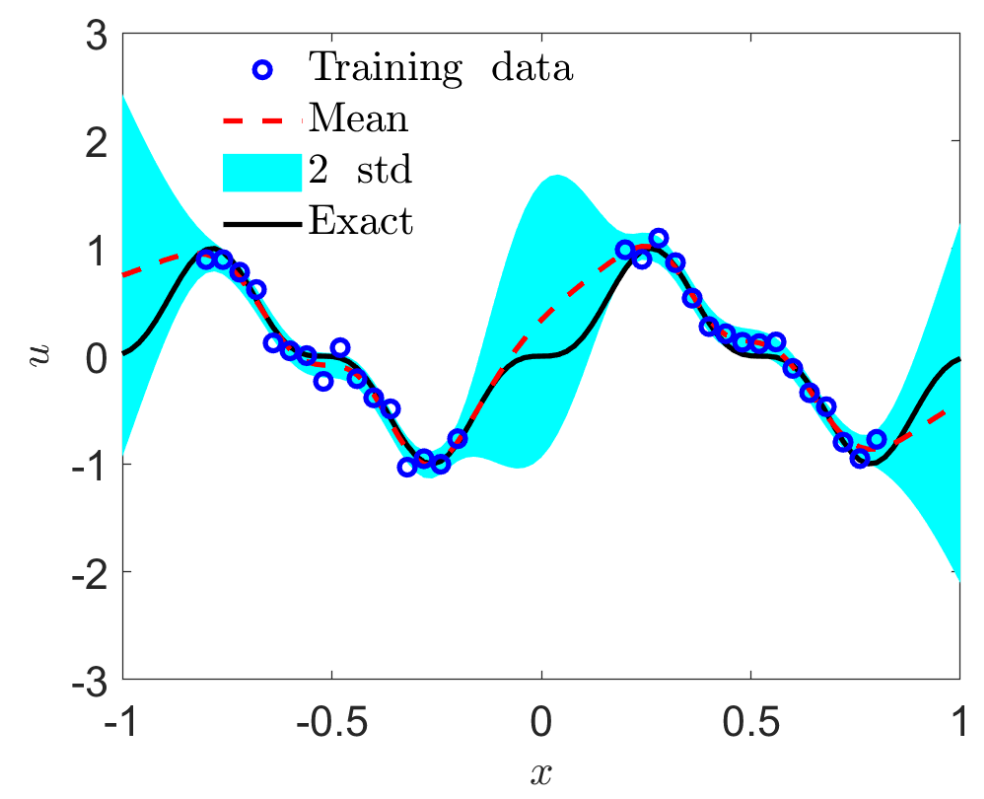}
    }
    \subfigure[Decomposition of total uncertainty~\citep{zou2024neuraluq}.]{
        \includegraphics[width=0.3\textwidth]{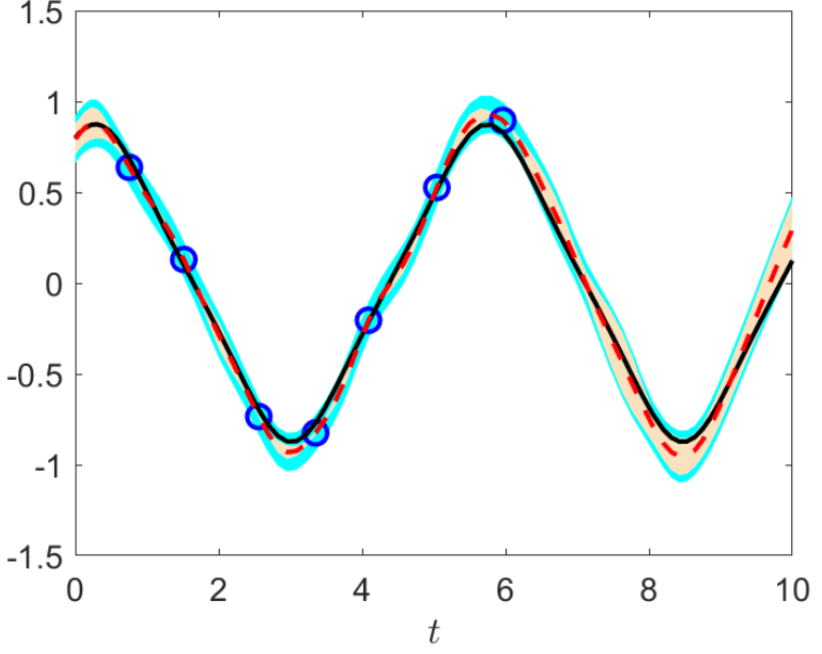}
    }
    \subfigure[Solving PDEs with few data based on a functional prior model~\citep{meng2022learning}.]{
        \includegraphics[width=0.3\textwidth]{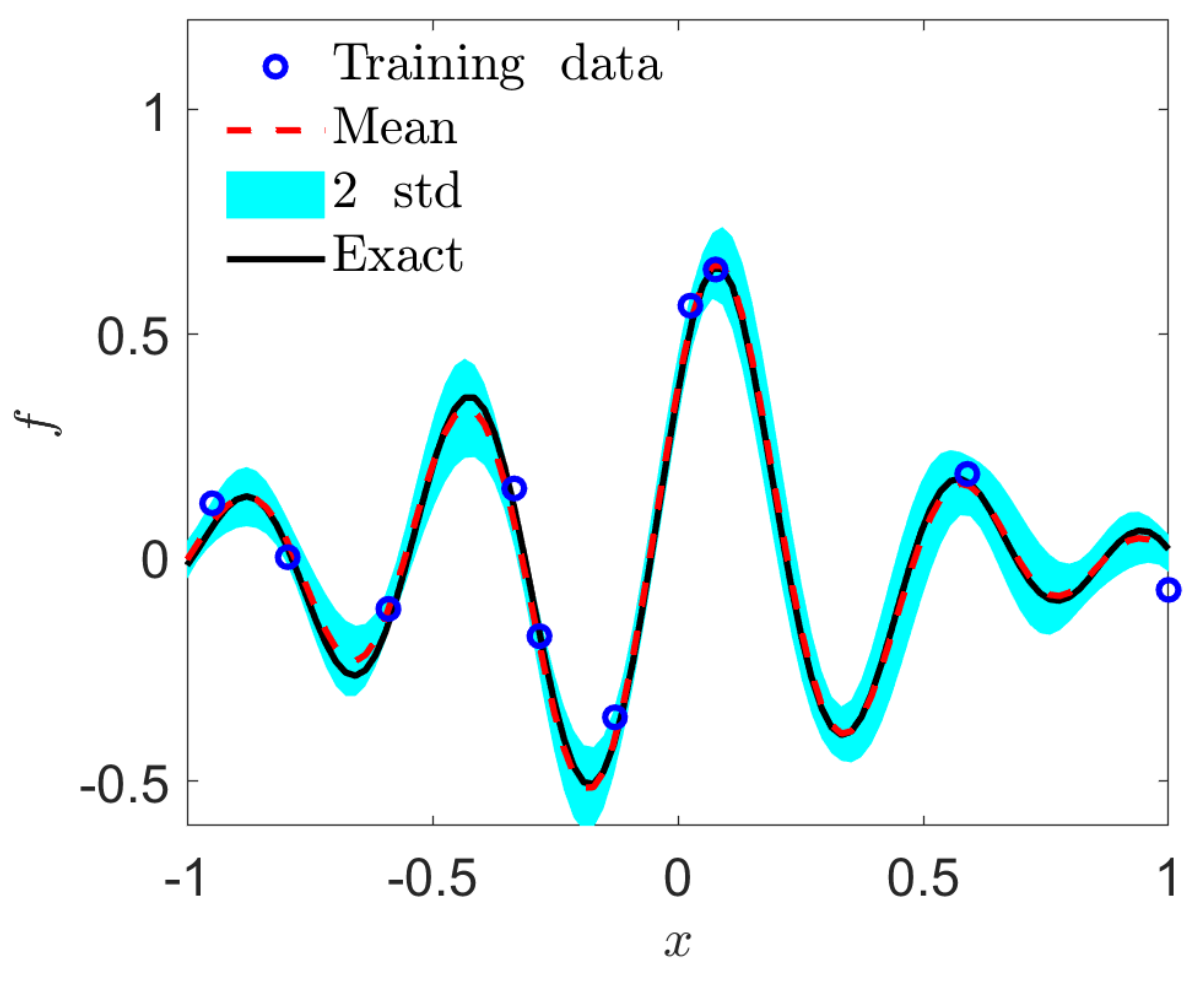}
    }
    \subfigure[When PINNs encounter model misspecifications~\citep{zou2024correcting}.]{
        \includegraphics[width=0.7\textwidth]{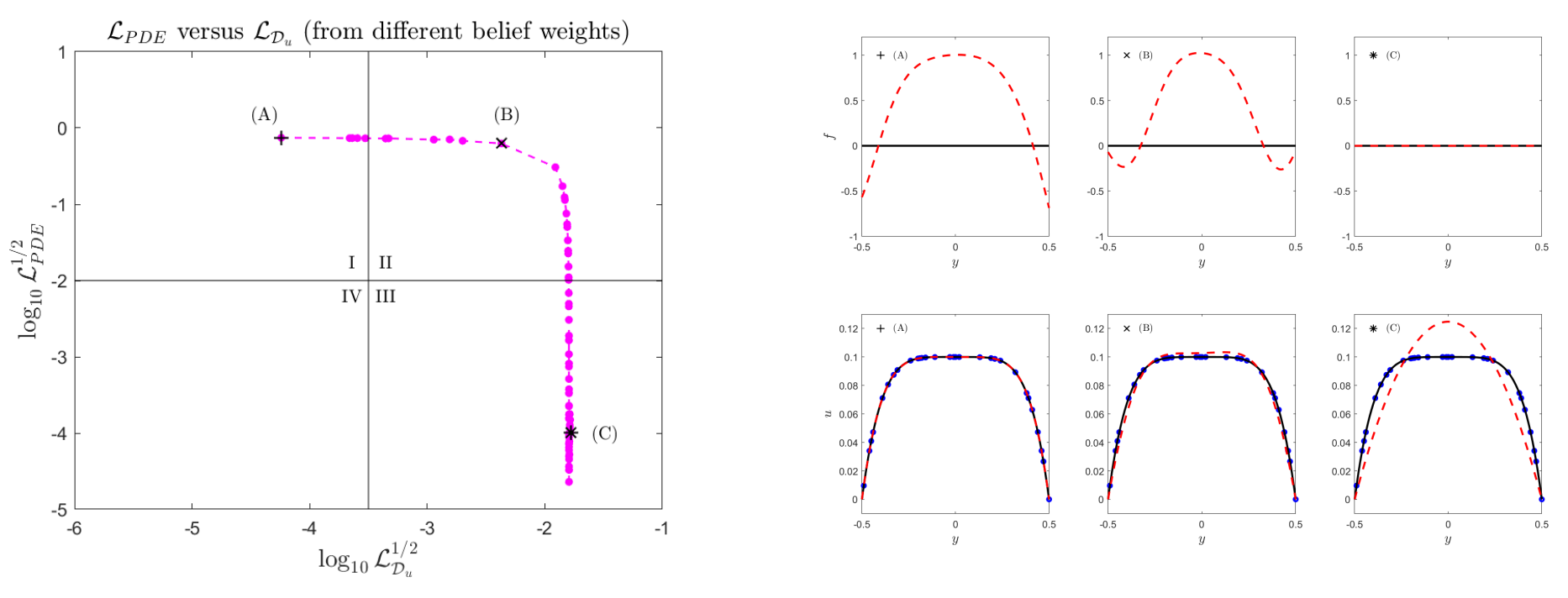}
    }
    \caption{(a) A breakdown of common sources of uncertainty in solving differential equations with PINNs (adapted from~\citep{psaros2023uncertainty}). (b) An example of noisy and gappy data in function approximation (adapted from~\citep{yang2021b}). (c) An illustration of decomposing total uncertainty into aleatoric and epistemic uncertainties in solving PDEs (adapted from~\citep{zou2024neuraluq}). (d) Utilizing informative functional prior in solving PDEs with noisy and gappy data (adapted from~\citep{meng2022learning}). (e) An example of physical model uncertainty in PINNs is assuming a Newtonian flow while the data comes from a non-Newtonian flow (adapted from~\citep{zou2024correcting}).}
    \label{fig:uq}
\end{figure}

This uncertainty occurs when the model lacks sufficient data or knowledge to make accurate predictions.
Aleatoric uncertainty can only be reduced by enhancing data quality, whereas epistemic uncertainty can be reduced by gathering more data, refining the model, or improving data quality~\citep{abdar2021review, psaros2023uncertainty}.

\subsection{Noisy and/or Gappy Data}
One common yet crucial source of uncertainty in solving differential equations is noisy data, which refers to uncertainty over the exact value of the sought solution, the model parameter, the source term, and/or the boundary term of the PDE. For example, in solving a forward problem, the source term $f$ in Eq. \eqref{eq:problem} can only be resolved by noisy measurements/observations at some discrete points rather than being precisely known.
Another important source of uncertainty is gappy data, which refers to incomplete information of aforementioned quantities, e.g., data of the source term are not available for a subdomain when solving a forward problem or data of the solution are not sufficient enough to make confident prediction/inference over the PDE parameter in an inverse problem. Fig.~\ref{fig:uq}(b) displays an illustrative example regarding the noisy and gappy data.

Tackling noisy and/or gappy data and quantifying associated uncertainty is often challenging in conventional numerical methods. Many PINN-based methods have been proposed to address such problems.
One notable method is the Bayesian PINNs (B-PINNs), introduced in~\citep{yang2021b}, which addresses forward and inverse PDE problems based on noisy and gappy data while quantifying the associated uncertainty through a Bayesian framework, estimating the corresponding posterior distributions. We note that the uncertainty quantified in~\citep{yang2021b} is epistemic, which can be interpreted as the uncertainty arising from gappy data and the noise model in observing the data. 
Besides B-PINNs, the Dropout method, which has been shown to effectively capture the epistemic uncertainty of neural networks~\citep{gal2016dropout}, was incorporated into PINNs for solving differential equations~\citep{zhang2019quantifying}. In~\citep{yang2019adversarial, gao2022wasserstein, daw2021pid}, generative adversarial networks were employed to quantify uncertainty stemming from uncertain data in solving PDE problems. In~\citep{psaros2023uncertainty, zou2024neuraluq, jiang2023practical, soibam2024inverse}, the deep ensemble PINNs method was developed as a practical UQ method to handle noisy and/or gappy data in both forward and inverse problems. In contrast, in~\citep{yang2022multi, de2024quantification}, a data perturbation technique was used to improve the performance of deep-ensemble-based UQ methods further. A multi-variance replica exchange method was introduced in~\citep{lin2022multi} to enhance the performance of B-PINNs in tackling the challenges posed by multimodal posterior distributions. In~\citep{lutjens2021pce}, the polynomial chaos expansion was integrated into the PINNs framework for propagation of parameter uncertainty. A continual learning perspective of solving PDEs with UQ under the PINN framework was discussed in~\citep{zou2024leveraging}.
In~\citep{psaros2023uncertainty}, a range of modern UQ methods for neural networks, such as snapshot ensemble and stochastic weight average Gaussian, were also integrated into the PINNs framework for UQ. 

Another important source of uncertainty when considering noisy data in PINNs is data (aleatoric) uncertainty. We note that in the noisy data regime, the data uncertainty essentially differs from the epistemic uncertainty stemming from imposing a noisy model. The former is caused by the randomness of the data and is irreducible even if more data are available, while the latter can be decreased given more available data. It was shown in~\citep{psaros2023uncertainty, zou2024neuraluq, de2024quantification} that the total uncertainty could be decomposed into separable aleatoric uncertainty and epistemic uncertainties when the noise model is additive and known by adopting the Bayesian framework. Fig.~\ref{fig:uq}(c) shows an illustrative example regarding total uncertainty.

Although the PINNs method can solve complex systems with sufficient data, it often struggles with sparse data. In~\citep{meng2022learning}, the authors showed that this issue could be resolved by leveraging an informative, functional prior model trained beforehand from historical or simulated data using generative adversarial networks. When a new task is considered, this functional prior model is employed to solve the PDE problem with UQ, given very little noisy data. An example is presented in Fig.~\ref{fig:uq}(d) for illustration. Similarly, in~\citep{zou2023hydra}, the authors employed the multi-head architecture and the normalizing flow~\citep{rezende2015variational} to learn a functional prior model to tackle the challenges of physics-informed learning with PINNs when data are sparse.
The functional prior model has proven to be very effective, particularly when data are sparse, and extrapolation is needed~\citep{psaros2023uncertainty, zou2024neuraluq, meng2023variational, yin2023generative}. 

While numerous UQ methods have been proposed to address noisy data in solving PDEs in PINNs, most of them assume that the input of data, i.e., the spatial-temporal coordinate $x$ in Eq. \eqref{eq:problem}, is certain and therefore fail to consider the uncertainty caused by the noisy inputs. Recently, in~\citep{zou2023uncertainty}, the authors investigated the impact of noisy inputs in solving forward and inverse PDE problems and employed a Bayesian approach to quantify the uncertainty arising from noisy inputs-outputs in PINNs. Specifically, they considered two independent noise models, one for the input and one for the output, and established likelihood distributions for both in the Bayesian framework.

\subsection{Physical Model Uncertainty}
Physical model uncertainty is one significant source of uncertainty, often overlooked in conventional numerical methods and refers to a lack of knowledge over the precise form of specific terms in PDE, e.g., the local production model of the reaction-diffusion system used to describe the spatiotemporal propagation of misfolded tau protein~\citep{zhang2024discovering} and the resistance model in the pulmonary compartment of the CVSim-6 cardiovascular ODE system~\citep{de2024quantification}. Ignoring this uncertainty results in physical model misspecifications, leading to significant discrepancies between data and physics, as well as inaccurate predictions in both forward and inverse problems~\citep{chen2021physics, chen2021generalized, zou2024correcting, ebers2024discrepancy, meng2024hj}. An example of when PINNs encounter model misspecification, where a Newtonian flow is assumed, but the underlying physics that generates the data is non-Newtonian, is illustrated in Fig.~\ref{fig:uq}(e).

In~\citep{zou2024correcting}, the issue of physical model misspecifications in PINNs was discussed. The authors proposed to use another neural network to model the discrepancy in the differential equation, which leads to two benefits: (1) the disagreement between physics and data, which is caused by the model misspecification, is corrected by adding this additional neural network, and (2) by successful training, the value of this neural network represents the discrepancy while its uncertainty represents the model uncertainty. Similarly to gappy data, this uncertainty could be categorized as epistemic uncertainty.

\subsection{Some Applications}

Recently developed UQ methods for PINNs have been utilized to tackle many real-world problems. For example,~\citep{lutjens2021pce} employed PINNs for uncertainty propagation of parameters in ocean modeling; the PINNs~\citep{kharazmi2021identifiability} and B-PINNs~\citep{linka2022bayesian} methods were used for UQ in epidemiology problems given real-world COVID-19 data;~\citep{oszkinat2022uncertainty} applied PINNs to estimate blood alcohol concentration with quantified uncertainty; traffic state estimation with UQ is accomplished by leveraging the PINN framework, generative adversarial networks and normalizing flows in~\citep{mo2022trafficflowgan}.

  

\section{Theoretical Advances in PIML}
\label{theoretical_advances}

\subsection{Error Estimates and Convergence in PIML}

\begin{figure}[h]
    \centering
    \includegraphics[width=0.75\linewidth]{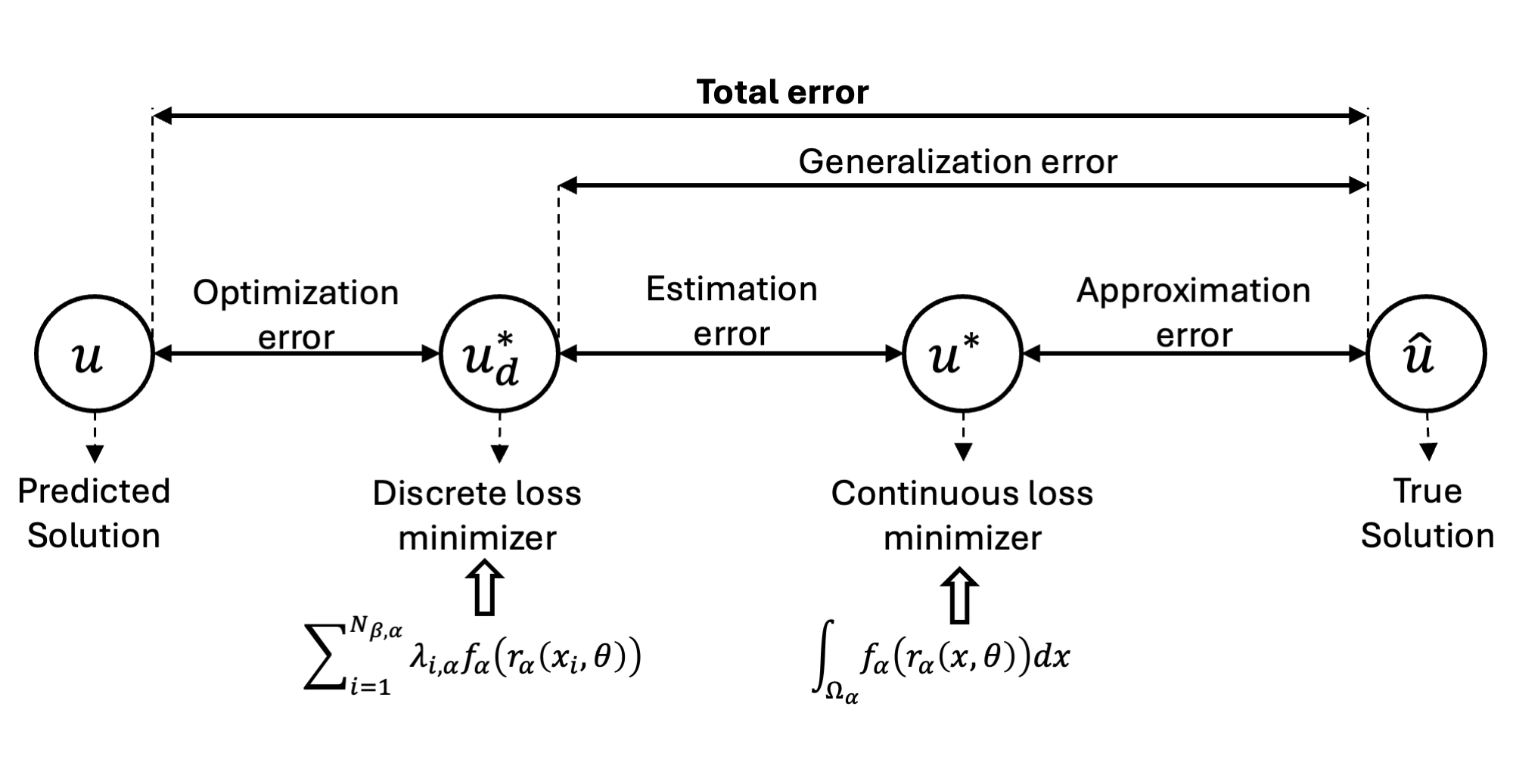}
\caption{\textbf{Decomposition of total error}. The total error in a PIML model can be decomposed into three parts~\citep{shin2020convergence}: (1) an optimization error resulting from the optimizer’s inability to fully minimize the loss function \( |u - u_d^*| \); (2) an estimation or quadrature error due to discretizing the loss function over a finite number of points \( |u^* - u_d^*| \); and (3) an approximation error, which reflects the expressivity or capacity of the representation model to capture the true solution \( |\hat{u} - u^*| \).Here $u_d^*$ is the minimizer of the loss with finite amount of data, and $u^*$ is the minimizer of the loss with infinitely many data.}

    \label{fig:Errors_PINNs}
\end{figure}

The goal of PIML is to find a representation model \( u^* \) that approximates the solution to a PDE/ODE, \( \hat{u} \), by minimizing a loss function (Eq.~\ref{loss_cont}) that accounts for the residuals from the governing equations, boundary conditions, and data in the continuous domain. As described in Section~\ref{optim_loss}, due to computational constraints, Eq.~\ref{loss_cont} is estimated in a discrete form (Eq.~\ref{loss_dic}), with a minimizer \( u^*_d \). However, as discussed in~\citep{shin2020convergence}, this function is minimized using a gradient-based optimization method which, due to the high non-convexity of \( \mathcal{L} \), does not fully recover \( u^*_d \), leading to a suboptimal solution \( u \). Under these assumptions, and as proposed by~\citep{shin2020convergence}, the total error in a PIML model (see Fig.~\ref{fig:Errors_PINNs}) can be decomposed into three parts: (1) an optimization error resulting from the optimizer’s inability to fully minimize the loss function \( |u - u_d^*| \); (2) an estimation or quadrature error due to the discretization of the loss function \( |u^* - u_d^*| \); and (3) an approximation error, which reflects the expressivity or capacity of the representation model to capture the true solution \( |\hat{u} - u^*| \).

Several studies have addressed convergence and error estimates in PIML~\citep{shin2020convergence, mishra2023estimates, de2022physics, wu2022convergence, qian2023error, zeinhofer2023unified, hu2023higher}, and a comprehensive review of these studies can be found in~\citep{cuomo2022scientific}. Shin et al.~\citep{shin2020convergence} pioneered this field by providing convergence estimates for linear second-order elliptic and parabolic PDEs. Subsequent studies~\citep{shin2023error} extended these results to all linear problems, including hyperbolic equations.~\citep{zeinhofer2023unified} provided sharp generalization error estimates for linear elliptic, parabolic, and hyperbolic PDEs, showing that the $L^2$ penalty approach for initial and boundary conditions weakens the error decay rate. On the other hand,~\citep{mishra2021physics} estimated generalization error in terms of the training error and number of training points, which was further extended to inverse problems in~\citep{mishra2022estimates}. Another line of work focuses on variational formulations of elliptic PDEs~\citep{muller2022error, yu2018deep}.

Other studies have focused on specific types of PDEs. For example,~\citep{de2024error} provided rigorous error bounds for the incompressible Navier-Stokes equations using X-PINNs~\citep{jagtap2020extended}, showing that the underlying PDE residual can be made arbitrarily small with tanh neural networks. Similarly,~\citep{biswas2022error} proposed an explicit error estimate and stability analysis for the incompressible Navier-Stokes equations. On the other hand,~\citep{de2022physics} focused on Kolmogorov-type equations, deriving bounds on the expectation of the $L^2$ error under the assumption that the weights of the neural network are bounded. Similarly,~\citep{mishra2021physics} analyzed the radiative transfer equation and proved that the generalization error is bounded by the training error, providing an estimate that depends only on the number of training points.

Finally,~\citep{doumeche2023convergence} provided a comprehensive theoretical analysis of the mathematical foundations driving PIML in both the hybrid modeling and PDE solver settings, showing that an additional level of regularization is sufficient to guarantee strong convergence.

\subsection{Training Dynamics} 

\subsubsection{Neural Tangent Kernel Perspective in PINNs} 
The Neural Tangent Kernel (NTK) offers a theoretical framework to understand the training dynamics of fully connected neural networks, particularly in the infinite-width limit. 
Jacot et al.~\citep{jacot2018neural} introduced the NTK theory, which provides insights into how network parameters evolve during training by linking them to kernel regression. 
For PINNs, the NTK describes the behavior of both the data-driven and physics-based components of the loss function. 
Wang et al.~\citep{wang2022when} derived the NTK of PINNs and demonstrated that, in the infinite-width limit, it converges to a deterministic kernel that remains constant during training. 
This kernel allows for an analysis of the training process in terms of eigenvalues, revealing that PINNs suffer from spectral bias, where higher frequency components of the solution are learned more slowly. 
Furthermore, a significant discrepancy in the convergence rates of the different loss components - 1) PDE residual loss and 2) data loss, can lead to stiff training dynamics and poor accuracy. 
To mitigate this,~\citep{wang2022when} proposed an adaptive training algorithm that uses the eigenvalues of the NTK to balance the contributions of these loss components. 

The NTK framework has been extended and used as a theoretical framework to develop and justify several algorithmic enhancements of PIML, such as Fourier feature embeddings~\citep{wang2021eigenvector}, weight factorization~\citep{wang2022random}, adaptive residual connections~\citep{wang2024piratenets}, global weights~\citep{wang2022when}, causality~\citep{wang2022respecting} and local weights~\citep{chen2024self}.

\subsubsection{Information Bottleneck Theory}

\begin{figure}[!h]
    \centering
    \includegraphics[width=\linewidth]{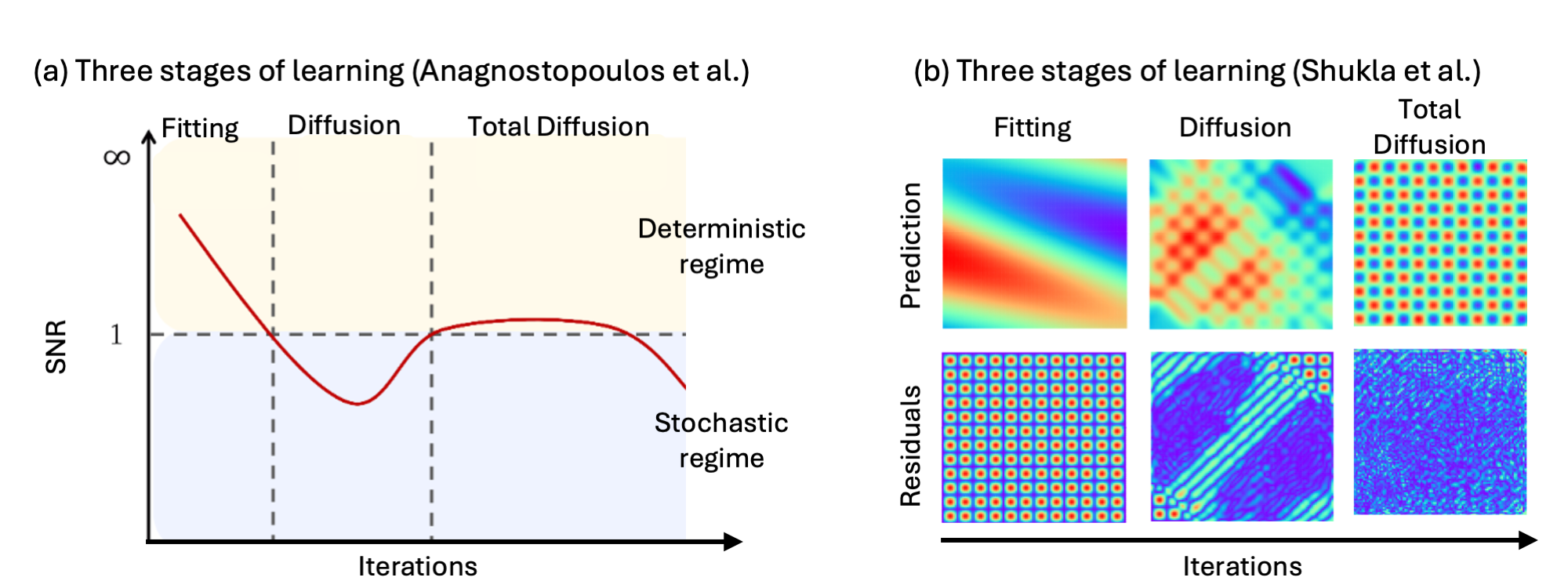}
    \caption{\textbf{Stages of Learning.}(a) The stages of learning can be identified using the signal-to-noise ratio (SNR) of the gradients. Highly deterministic regimes are characterized by a high SNR, while highly stochastic regimes are characterized by a low SNR~\citep{anagnostopoulos2024learning}. During \textbf{fitting}, PIML model's SNR go from high to low. This suggests an initial phase where the model closely fits the training data. The \textbf{diffusion} phase is considered an exploratory stage characterized by a fluctuating low SNR. In the last stage, \textbf{total diffusion}, the SNR suddenly increases and converges to a critical value, and the generalization (i.e., relative  $L^2$) error decreases faster, suggesting an optimal convergence~\citep{anagnostopoulos2024learning}. (b) Prediction and residual distributions at different stages of learning in for Helmholtz equation~\citep{shukla2024comprehensive}. The \textbf{fitting} phase is highly deterministic, so the residuals display an ordered pattern. As the SNR decreases and the model transitions to a stochastic stage, \textbf{diffusion}, the residuals gradually become disordered. Finally, in \textbf{total diffusion}, the model reaches an equilibrium state, simplifies internal representations, and reduces their complexity, making it much more efficient and generalizable. Notice that during this stage, the predictions closely match the analytical solution. This phase is characterized by highly stochastic (i.e., noisy) residuals~\citep{anagnostopoulos2024learning}.}
    \label{fig:Stages_learning}
\end{figure}

The Information Bottleneck (IB) theory provides an information-theoretic perspective on the training and performance of neural networks. It presents a framework for forming a condensed representation of layer activations, \( \mathcal{T} \), with respect to an input variable \( x \in \mathcal{X} \), retaining as much information as possible about an output variable \( y \in \mathcal{Y} \)~\citep{tishby2000information, tishby2015deep}. A key concept in this theory is the mutual information \( I(x, y) \), which suggests that optimal model representations preserve all relevant information about the output while discarding irrelevant input information, thus creating an ``information bottleneck". IB theory identifies two distinct phases of learning: fitting and diffusion, separated by a phase transition driven by the signal-to-noise ratio (SNR) of the gradients~\citep{shwartz2017opening, goldfeld2020information, shwartz2022information}. Anagnostopoulos et al.~\citep{anagnostopoulos2024learning} extended this framework to study learning dynamics in PINNs, proposing the existence of a third phase called total diffusion (see Fig.~\ref{fig:Stages_learning}(a)). Subsequent studies~\citep{shukla2024comprehensive} demonstrated that these three phases are also observable in other representation models, such as KANs.

\paragraph{Signal-to-Noise Ratio (SNR)}

As described in~\citep{anagnostopoulos2024learning, shukla2024comprehensive, goldfeld2020information, shwartz2022information}, the batch-wise signal-to-noise ratio (SNR) is a key metric for understanding the training dynamics of neural networks. It is defined as:

\[
\text{SNR} = \frac{\lVert \mu \rVert_{2}}{\lVert \sigma \rVert_{2}} = \frac{\lVert \mathbb{E}[\nabla_{\theta}{\mathcal{L}_{\mathcal{B}}}] \rVert_{2}}{\lVert \text{std}[\nabla_{\theta}{\mathcal{L}_{\mathcal{B}}}] \rVert_{2}},
\]
\noindent where \( \theta \) represents the network parameters, and \( \lVert \mu \rVert_{2} \) and \( \lVert \sigma \rVert_{2} \) are the \( L^2 \)-norms of the batch-wise mean and standard deviation of the gradients of the total loss \( \nabla_{\theta}{\mathcal{L}_{\mathcal{B}}} \). According to~\citep{shukla2024comprehensive}, the ``signal" refers to an idealized gradient that drives the optimizer to reduce errors across all subdomains, while ``noise" represents perturbations from this ideal gradient due to the finite number of observations.

\paragraph{Stages of Learning}

The stages of learning can be viewed as a process where the model initially fits the data (capturing relevant information) and then compresses it (discarding irrelevant information), thereby enhancing generalization capabilities~\citep{anagnostopoulos2024learning}. Each phase is defined by the dominant term in the SNR. Deterministic regimes exhibit high SNR, characterized by a dominant signal, while stochastic stages are marked by low SNR, where noise dominates (see Fig.~\ref{fig:Stages_learning}(a)).

\subparagraph{Fitting}
At the start of training, both the loss and its gradients are high across all subdomains. This agreement results in an initially high SNR, where a clear signal (i.e., direction) helps reduce the training error across the subdomains. However, as the loss and gradients decrease, disagreement between subdomains (i.e., noise) increases, leading to a decline in the SNR. Therefore, the fitting stage is a deterministic phase, transitioning from high to low SNR as the residuals display an ordered pattern~\citep{anagnostopoulos2024learning} (see Fig.~\ref{fig:Stages_learning}(b)).

\subparagraph{Diffusion}
After the model has adequately fitted the data, it enters an exploration phase, searching for a signal (i.e., direction) that reduces the training error across all subdomains. During this phase, the network weights diffuse, enhancing the model's generalization capabilities by disrupting the initial order of parameters. The diffusion stage is characterized by a fluctuating, low SNR~\citep{anagnostopoulos2024learning}. In this stochastic stage, the residuals become disordered (see Fig.~\ref{fig:Stages_learning}(b)), and the SNR begins to oscillate.

\subparagraph{Total Diffusion}
In the final phase, after the model identifies an optimal signal, the SNR increases rapidly, reaching equilibrium as the model exploits a consistent direction to minimize generalization error across subdomains. During this phase, the model simplifies its internal representation by retaining essential features and discarding irrelevant ones, effectively reducing its complexity and breaking the order of the residuals. As shown in Fig.~\ref{fig:Stages_learning}(b), this phase is marked by highly disordered residuals. However, as observed in~\citep{anagnostopoulos2024learning, shukla2024comprehensive}, once the optimal direction (i.e., signal) is found and total diffusion begins, the generalization error (i.e., relative \( L^2 \)) decreases rapidly, indicating optimal convergence. Unsurprisingly, the best-performing models transition to total diffusion first~\citep{anagnostopoulos2024learning, shukla2024comprehensive}.

One of the main advantages of this approach is that it provides a measure of convergence that can be used to evaluate a model's performance. Previous studies~\citep{anagnostopoulos2024learning, shukla2024comprehensive} have shown that the best-performing models typically reach the total diffusion phase earlier than others. Conversely, models that fail to converge often remain trapped in the diffusion stage, unable to progress further. Additionally, this theoretical framework reflects the underlying training dynamics, which could potentially be leveraged to develop new algorithms or architectures that optimize learning efficiency and enhance model generalization.
\section{Computational Frameworks and Software}
\label{computational_frameworks}

As the PINNs method emerges as a revolutionary approach to solving PDE problems, many software packages have been developed for fast and reliable implementations of PINNs based on different computational frameworks. Most of these were developed in Python to leverage modern machine learning computing tools, while some are in other programming languages, e.g., the NeuralPDE.jl~\citep{zubov2021neuralpde}, and the ADCME libraries were developed in Julia. We note that in this section, we provide a brief review of these libraries only for PINNs, while some have been developed further to address broader applications in scientific machine learning, such as neural operators, which are beyond the scope of this paper.

A notable library is DeepXDE~\citep{lu2021deepxde}, which was initially developed based on the TensorFlow platform and now supports multiple backends, including PyTorch, JAX, and PaddlePaddle. The DeepXDE library also supports complex domain geometries, various boundary conditions, different automatic differentiation methods, and many more in solving differential equations using PINNs. On the other hand, NVIDIA's Modulus~\citep{modulus} supports only the PyTorch backend but specializes in scalable training with the power of NVIDIA GPUs maximized for training PINNs. The NeuroDiffEq~\citep{chen2020neurodiffeq} and TorchPhysics~\citep{TorchPhysics} libraries are PyTorch-based: the former provides additional toolsets for transfer learning and conveniently obtaining solution bundle for inverse problems, while the latter features complex domain generations. SciANN~\citep{haghighat2021sciann} is a high-level wrapper based on TensorFlow and Keras. 
PyDENs~\citep{koryagin2019pydens} (PyTorch-based) focuses more on solving parametric families of PDEs, such as heat and wave equations. NeuralPDE.jl~\citep{zubov2021neuralpde} and ADCME~\citep{xu2020adcme} are two libraries in Julia: the former is more comprehensive for PINNs while the latter specializes more in tackling inverse problems and was developed based on the Julia version of TensorFlow.
The TensorDiffEq library~\citep{mcclenny2021tensordiffeq} is based on the TensorFlow platform and supports multi-GPU distributed training for large systems. The NeuralUQ library~\citep{zou2024neuraluq} (Tensorflow-based) specializes in uncertainty quantification. Some other packages for solving differential equations with PINNs can be found in~\citep{araz2021elvet, peng2021idrlnet, pedro2019solving}.

\section{Discussion and Outlook}
\label{discussion}
Physics-informed machine learning and PINNs, in particular, have 
disrupted scientific computing in a fundamental way, enabling a seamless integration of data and physics, unlike conventional numerical methods that require expensive and often inaccurate data assimilation schemes. This new capability has been applied in diverse applications, well beyond the intention in the original paper. Despite the numerous limitations of the vanilla PINNs, the computational community has embraced this new capability and attempted to further tailor PINNs and enhance both their efficiency and accuracy in various settings, from mechanics to geophysics and even to quantitative pharmacology. 

In the Appendix, we provide a table of the evolution of the algorithmic variants and enhancements so far, which addressed the early limitations of PINNs. The first key development was the introduction of self-adaptive weights based on adversarial (min-max) training, which is particularly important for multiscale problems~\citep{mcclenny2020self}. Similarly, introducing feature expansion layers to deal with the spectral bias of neural networks was an essential development in tackling problems with high frequencies and wavenumbers~\citep{cai2019multi}. The development of the jax framework has been instrumental in computing high-order derivatives fast and accurately, in addition to overall speeding up PINN computations~\citep{frostig2018compiling}.
To this end, domain decomposition for conservation laws but also for other PDEs has been introduced to scale up PINNs to large spatio-temporal domains and, moreover, to take advantage of multi-GPU computing~\citep{jagtap2020extended}. In this setting,  a separate PINN with different hyperparameters can be assigned in each subdomain, which also helps with accuracy enhancement for multiphysics problems.
In addition to adaptive weights, the introduction of adaptive activation functions affects both the accuracy but also convergence as, under some conditions, there are theoretical guarantees of avoiding bad minima~\citep{jagtap2020locally}.

A significant advance was the introduction of separable PINNs and tensor networks that demonstrated a speed-up of almost two orders of magnitude~\citep{cho2023separablephysicsinformedneuralnetworks}.
However, even with such a computational speed-up, there is currently a debate if PINNs can compete with FEM for forward problems since for inverse problems PINNs is a clear winner. This question is important and will continue to drive research but it is important to put it in the right context and also define the proper benchmarks. For example, for solving PDEs in 1D and 2D,  FEM performs better both in terms of accuracy and speed but 
in higher dimensions PINNs will outperform not only FEM but any method. For example, recent work on high-dimensional PDEs, such as the 
Hamilton-Jacobi-Bellman and Fokker-Planck equations demonstrated that proper formulations of PINNs can tackle even 100,000 dimensions at the cost of only an hour on a single GPU, a task that no other method can handle~\citep{hu2024tackling}. 

A recognized difficulty with PINNs is the inference of dynamical systems that exhibit chaotic and turbulent responses, especially for long-time integration. This is primarily due to the limited accuracy of the vanilla PINNs in forward problems with no data available due to the optimization error that dominates the overall error. However, recent proposals for multi-staged training and stacked architectures have demonstrated the ability of these modified versions of PINNs to reach accuracy down to machine precision, which is important for resolving chaotic trajectories in chaotic dynamical systems~\citep{howard2023stacked,wang2023multistageneuralnetworksfunction}.

The question of the high computational cost currently under debate for general neural networks also pertains to PINNs. The introduction of deep neural operators in 2019, such as DeepOnet, by our group has paved the way for significant progress in alleviating this cost~\citep{lu2019deeponet}. Moreover, new representation models like KANs~\citep{liu2024kan} and PIKANs~\citep{shukla2024comprehensive} that we reviewed here
lead to smaller models and may also contribute to lowering the computational cost while maintaining good accuracy. Another direction is to further advance spiking neural networks~\citep{zhang2023artificial} that will be working on specially designed neuromorphic chips, like Intel's Loihi 2, but at the moment, the demonstrated accuracy is below the one obtained with NNs and KANs. Yet another possible research direction is to develop hybrid methods that blend PINNs and conventional numerical methods via domain decomposition techniques. For example, PINNs or PIKANs can be used only where data is available and interfaced with FEM or other methods via proper boundary conditions. In addition to algorithmic developments, new software is required to handle the typically heterogeneous programming environments of classical methods with deep learning methods.

In summary, PINNs and PIKANs have significantly advanced in a very short time. They can outperform conventional methods for inverse problems, e.g., parameter estimation and discovering missing physics in gray-box type scenarios. Future work should focus on speeding up the optimization process, increasing the overall accuracy, and automating hyperparameter tuning so that these methods become robust and competitive with FEM and other classical methods, even for forward-type simulation problems.

\section*{Acknowledgements}
We acknowledge the support of the MURI/AFOSR FA9550-20-1-0358 project, the DOE-MMICS SEA-CROGS DE-SC0023191 award, and the ONR Vannevar Bush Faculty Fellowship (N00014-22-1-2795).

 \bibliographystyle{elsarticle-num} 
 \bibliography{cas-refs}

\begin{thebibliography}{100}
\expandafter\ifx\csname url\endcsname\relax
  \def\url#1{\texttt{#1}}\fi
\expandafter\ifx\csname urlprefix\endcsname\relax\def\urlprefix{URL }\fi
\expandafter\ifx\csname href\endcsname\relax
  \def\href#1#2{#2} \def\path#1{#1}\fi

\bibitem{raissi2019physics}
M.~Raissi, P.~Perdikaris, G.~E. Karniadakis, {Physics-informed neural networks: A deep learning framework for solving forward and inverse problems involving nonlinear partial differential equations}, Journal of Computational Physics 378 (2019) 686--707.

\bibitem{raissi2017physicsI}
M.~Raissi, P.~Perdikaris, G.~E. Karniadakis, {Physics Informed Deep Learning (Part I): Data-driven Solutions of Nonlinear Partial Differential Equations}, arXiv preprint arXiv:1711.10561 (2017).

\bibitem{raissi2017physicsII}
M.~Raissi, P.~Perdikaris, G.~E. Karniadakis, {Physics Informed Deep Learning (Part II): Data-driven Discovery of Nonlinear Partial Differential Equations}, arXiv preprint arXiv:1711.10566 (2017).

\bibitem{liu2024kan}
Z.~Liu, Y.~Wang, S.~Vaidya, F.~Ruehle, J.~Halverson, M.~Solja{\v{c}}i{\'c}, T.~Y. Hou, M.~Tegmark, {KAN: Kolmogorov-Arnold Networks}, arXiv preprint arXiv:2404.19756 (2024).

\bibitem{cuomo2022scientific}
S.~Cuomo, V.~S. Di~Cola, F.~Giampaolo, G.~Rozza, M.~Raissi, F.~Piccialli, {Scientific machine learning through physics--informed neural networks: Where we are and what’s next}, Journal of Scientific Computing 92~(3) (2022) 88.

\bibitem{farea2024understanding}
A.~Farea, O.~Yli-Harja, F.~Emmert-Streib, {{Understanding Physics-Informed Neural Networks: Techniques, Applications, Trends, and Challenges}}, AI 5~(3) (2024) 1534--1557.

\bibitem{ganga2024exploringphysicsinformedneuralnetworks}
S.~Ganga, Z.~Uddin, {Exploring Physics-Informed Neural Networks: From Fundamentals to Applications in Complex Systems}, arXiv preprint arXiv:2410.00422 (2024).

\bibitem{raissi2024physics}
M.~Raissi, P.~Perdikaris, N.~Ahmadi, G.~E. Karniadakis, {Physics-informed neural networks and extensions}, arXiv preprint arXiv:2408.16806 (2024).

\bibitem{chi2024comprehensive}
C.~Zhao, F.~Zhang, W.~Lou, X.~Wang, J.~Yang, {A comprehensive review of advances in physics-informed neural networks and their applications in complex fluid dynamics}, Physics of Fluids 36~(10) (2024) 101301.

\bibitem{cai2021physics}
S.~Cai, Z.~Mao, Z.~Wang, M.~Yin, G.~E. Karniadakis, {Physics-informed neural networks ({PINNs}}) for fluid mechanics: A review, Acta Mechanica Sinica 37~(12) (2021) 1727--1738.

\bibitem{huang2022applications}
B.~Huang, J.~Wang, {Applications of physics-informed neural networks in power systems-a review}, IEEE Transactions on Power Systems 38~(1) (2022) 572--588.

\bibitem{lawal2022physics}
Z.~K. Lawal, H.~Yassin, D.~T.~C. Lai, A.~Che~Idris, {Physics-informed neural network ({PINN}}) evolution and beyond: A systematic literature review and bibliometric analysis, Big Data and Cognitive Computing 6~(4) (2022) 140.

\bibitem{raissi2017machine}
M.~Raissi, P.~Perdikaris, G.~E. Karniadakis, {Machine learning of linear differential equations using Gaussian processes}, Journal of Computational Physics 348 (2017) 683--693.

\bibitem{raissi2021physics}
M.~Raissi, P.~Perdikaris, G.~E. Karniadakis, {Physics informed learning machine}, {US Patent 10,963,540} (Mar.~30 2021).

\bibitem{dissanayake1994neural}
M.~G. Dissanayake, N.~Phan-Thien, {Neural-network-based approximations for solving partial differential equations}, Communications in Numerical Methods in Engineering 10~(3) (1994) 195--201.

\bibitem{lagaris1998artificial}
I.~E. Lagaris, A.~Likas, D.~I. Fotiadis, {Artificial neural networks for solving ordinary and partial differential equations}, IEEE Transactions on Neural Networks 9~(5) (1998) 987--1000.

\bibitem{shukla2024comprehensive}
K.~Shukla, J.~D. Toscano, Z.~Wang, Z.~Zou, G.~E. Karniadakis, {A comprehensive and {FAIR}} comparison between {MLP} and {KAN} representations for differential equations and operator networks, Computer Methods in Applied Mechanics and Engineering 431 (2024) 117290.

\bibitem{cai2021flow}
S.~Cai, Z.~Wang, F.~Fuest, Y.~J. Jeon, C.~Gray, G.~E. Karniadakis, {Flow over an espresso cup: inferring 3-D velocity and pressure fields from tomographic background oriented Schlieren via physics-informed neural networks}, Journal of Fluid Mechanics 915 (2021).

\bibitem{raissi2020hidden}
M.~Raissi, A.~Yazdani, G.~E. Karniadakis, {Hidden fluid mechanics: Learning velocity and pressure fields from flow visualizations}, Science 367~(6481) (2020) 1026--1030.

\bibitem{hu2024tackling}
Z.~Hu, K.~Shukla, G.~E. Karniadakis, K.~Kawaguchi, {Tackling the curse of dimensionality with physics-informed neural networks}, Neural Networks 176 (2024) 106369.

\bibitem{jin2021nsfnets}
X.~Jin, S.~Cai, H.~Li, G.~E. Karniadakis, {{NSF}}nets ({Navier-Stokes flow nets}): Physics-informed neural networks for the incompressible {Navier-Stokes} equations, Journal of Computational Physics 426 (2021) 109951.

\bibitem{shukla2024neurosem}
K.~Shukla, Z.~Zou, C.~H. Chan, A.~Pandey, Z.~Wang, G.~E. Karniadakis, {{NeuroSEM}}: A hybrid framework for simulating multiphysics problems by coupling {PINNs} and spectral elements, arXiv preprint arXiv:2407.21217 (2024).

\bibitem{yang2019adversarial}
Y.~Yang, P.~Perdikaris, {Adversarial uncertainty quantification in physics-informed neural networks}, Journal of Computational Physics 394 (2019) 136--152.

\bibitem{pang2019fpinns}
G.~Pang, L.~Lu, G.~E. Karniadakis, {{fPINNs}}: Fractional physics-informed neural networks, SIAM Journal on Scientific Computing 41~(4) (2019) A2603--A2626.

\bibitem{karniadakis2021physics}
G.~E. Karniadakis, I.~G. Kevrekidis, L.~Lu, P.~Perdikaris, S.~Wang, L.~Yang, {Physics-informed machine learning}, Nature Reviews Physics 3~(6) (2021) 422--440.

\bibitem{baydin2018automatic}
A.~G. Baydin, B.~A. Pearlmutter, A.~A. Radul, J.~M. Siskind, {Automatic differentiation in machine learning: a survey}, Journal of Machine Learning Research 18~(153) (2018) 1--43.

\bibitem{meng2020ppinn}
X.~Meng, Z.~Li, D.~Zhang, G.~E. Karniadakis, {{PPINN}}: Parareal physics-informed neural network for time-dependent {PDEs}, Computer Methods in Applied Mechanics and Engineering 370 (2020) 113250.

\bibitem{hornik1989multilayer}
K.~Hornik, M.~Stinchcombe, H.~White, {Multilayer feedforward networks are universal approximators}, Neural Networks 2~(5) (1989) 359--366.

\bibitem{dong2021method}
S.~Dong, N.~Ni, {{A method for representing periodic functions and enforcing exactly periodic boundary conditions with deep neural networks}}, Journal of Computational Physics 435 (2021) 110242.

\bibitem{guan2023dimension}
W.~Guan, K.~Yang, Y.~Chen, S.~Liao, Z.~Guan, {A dimension-augmented physics-informed neural network ({DaPINN}}) with high level accuracy and efficiency, Journal of Computational Physics 491 (2023) 112360.

\bibitem{wang2021eigenvector}
S.~Wang, H.~Wang, P.~Perdikaris, {{On the eigenvector bias of Fourier feature networks: From regression to solving multi-scale PDEs with physics-informed neural networks}}, Computer Methods in Applied Mechanics and Engineering 384 (2021) 113938.

\bibitem{wang2024piratenets}
S.~Wang, B.~Li, Y.~Chen, P.~Perdikaris, {PirateNets: Physics-informed Deep Learning with Residual Adaptive Networks}, arXiv preprint arXiv:2402.00326 (2024).

\bibitem{salimans2016weight}
T.~Salimans, D.~P. Kingma, {Weight normalization: A simple reparameterization to accelerate training of deep neural networks}, Advances in Neural Information Processing Systems 29 (2016).

\bibitem{wang2022random}
S.~Wang, H.~Wang, J.~H. Seidman, P.~Perdikaris, {Random weight factorization improves the training of continuous neural representations}, arXiv preprint arXiv:2210.01274 (2022).

\bibitem{jagtap2022deep}
A.~D. Jagtap, D.~Mitsotakis, G.~E. Karniadakis, {Deep learning of inverse water waves problems using multi-fidelity data: Application to Serre--Green--Naghdi equations}, Ocean Engineering 248 (2022) 110775.

\bibitem{sukumar2022exact}
N.~Sukumar, A.~Srivastava, {{Exact imposition of boundary conditions with distance functions in physics-informed deep neural networks}}, Computer Methods in Applied Mechanics and Engineering 389 (2022) 114333.

\bibitem{wang2021understanding}
S.~Wang, Y.~Teng, P.~Perdikaris, {Understanding and mitigating gradient flow pathologies in physics-informed neural networks}, SIAM Journal on Scientific Computing 43~(5) (2021) A3055--A3081.

\bibitem{cai2021artificial}
S.~Cai, H.~Li, F.~Zheng, F.~Kong, M.~Dao, G.~E. Karniadakis, S.~Suresh, {Artificial intelligence velocimetry and microaneurysm-on-a-chip for three-dimensional analysis of blood flow in physiology and disease}, Proceedings of the National Academy of Sciences 118~(13) (2021).

\bibitem{anagnostopoulos2024residual}
S.~J. Anagnostopoulos, J.~D. Toscano, N.~Stergiopulos, G.~E. Karniadakis, {Residual-based attention in physics-informed neural networks}, Computer Methods in Applied Mechanics and Engineering 421 (2024) 116805.

\bibitem{mao2020physics}
Z.~Mao, A.~D. Jagtap, G.~E. Karniadakis, {Physics-informed neural networks for high-speed flows}, Computer Methods in Applied Mechanics and Engineering 360 (2020) 112789.

\bibitem{zapf2022investigating}
B.~Zapf, J.~Haubner, M.~Kuchta, G.~Ringstad, P.~K. Eide, K.-A. Mardal, {{Investigating molecular transport in the human brain from {MRI}}} with physics-informed neural networks, Scientific Reports 12~(1) (2022) 15475.

\bibitem{rahaman2019spectral}
N.~Rahaman, A.~Baratin, D.~Arpit, F.~Draxler, M.~Lin, F.~Hamprecht, Y.~Bengio, A.~Courville, {On the spectral bias of neural networks}, in: {International conference on machine learning}, PMLR, 2019, pp. 5301--5310.

\bibitem{cao2019towards}
Y.~Cao, Z.~Fang, Y.~Wu, D.-X. Zhou, Q.~Gu, {Towards understanding the spectral bias of deep learning}, arXiv preprint arXiv:1912.01198 (2019).

\bibitem{cai2019multi}
W.~Cai, Z.-Q.~J. Xu, {Multi-scale deep neural networks for solving high dimensional PDEs}, arXiv preprint arXiv:1910.11710 (2019).

\bibitem{liu2020multi}
Z.~Liu, W.~Cai, Z.-Q.~J. Xu, {Multi-scale deep neural network (MscaleDNN) for solving Poisson-Boltzmann equation in complex domains}, arXiv preprint arXiv:2007.11207 (2020).

\bibitem{wang2020multi}
B.~Wang, W.~Zhang, W.~Cai, {Multi-scale deep neural network (MscaleDNN) methods for oscillatory stokes flows in complex domains}, arXiv preprint arXiv:2009.12729 (2020).

\bibitem{liu2022linearized}
L.~Liu, W.~Cai, et~al., {Linearized learning with multiscale deep neural networks for stationary Navier-Stokes equations with oscillatory solutions}, East Asian Journal on Applied Mathematics 13~(3) (2022).

\bibitem{ahmadi2024ai}
N.~Ahmadi~Daryakenari, M.~De~Florio, K.~Shukla, G.~E. Karniadakis, {AI-Aristotle: A physics-informed framework for systems biology gray-box identification}, PLOS Computational Biology 20~(3) (2024) e1011916.

\bibitem{zhang4957859pkan}
Z.~Zhang, T.~Shen, Y.~Zhang, W.~Zhang, Q.~Wang, {AL-PKAN: A Hybrid GRU-KAN Network with Augmented Lagrangian Function for Solving PDEs}, Available at SSRN 4957859 (2024).

\bibitem{toscano2024inferring}
J.~D. Toscano, T.~K{\"a}ufer, M.~Maxey, C.~Cierpka, G.~E. Karniadakis, {Inferring turbulent velocity and temperature fields and their statistics from Lagrangian velocity measurements using physics-informed Kolmogorov-Arnold Networks}, arXiv preprint arXiv:2407.15727 (2024).

\bibitem{chen2020comparison}
J.~Chen, R.~Du, K.~Wu, {A comparison study of deep Galerkin method and deep Ritz method for elliptic problems with different boundary conditions}, arXiv preprint arXiv:2005.04554 (2020).

\bibitem{zeinhofer2023unified}
M.~Zeinhofer, R.~Masri, K.-A. Mardal, {A unified framework for the error analysis of physics-informed neural networks}, arXiv preprint arXiv:2311.00529 (2023).

\bibitem{berrone2023enforcing}
S.~Berrone, C.~Canuto, M.~Pintore, N.~Sukumar, {Enforcing Dirichlet boundary conditions in physics-informed neural networks and variational physics-informed neural networks}, Heliyon 9~(8) (2023).

\bibitem{leake2020deep}
C.~Leake, D.~Mortari, {{Deep theory of functional connections: A new method for estimating the solutions of partial differential equations}}, Machine Learning and Knowledge Extraction 2~(1) (2020) 37--55.

\bibitem{lu2021physics}
L.~Lu, R.~Pestourie, W.~Yao, Z.~Wang, F.~Verdugo, S.~G. Johnson, {Physics-informed neural networks with hard constraints for inverse design}, SIAM Journal on Scientific Computing 43~(6) (2021) B1105--B1132.

\bibitem{wang2022respecting}
S.~Wang, S.~Sankaran, P.~Perdikaris, {Respecting causality is all you need for training physics-informed neural networks}, arXiv preprint arXiv:2203.07404 (2022).

\bibitem{barschkis2023exact}
S.~Barschkis, {Exact and soft boundary conditions in Physics-Informed Neural Networks for the Variable Coefficient Poisson equation}, arXiv preprint arXiv:2310.02548 (2023).

\bibitem{anagnostopoulos2024learning}
S.~J. Anagnostopoulos, J.~D. Toscano, N.~Stergiopulos, G.~E. Karniadakis, {Learning in {PINNs}}: Phase transition, total diffusion, and generalization, arXiv preprint arXiv:2403.18494 (2024).

\bibitem{jin2020sympnets}
P.~Jin, Z.~Zhang, A.~Zhu, Y.~Tang, G.~E. Karniadakis, {{SympNets: Intrinsic structure-preserving symplectic networks for identifying Hamiltonian systems}}, Neural Networks 132 (2020) 166--179.

\bibitem{zhou2023flow}
K.~Zhou, S.~J. Grauer, {Flow reconstruction and particle characterization from inertial Lagrangian tracks}, arXiv preprint arXiv:2311.09076 (2023).

\bibitem{jagtap2020adaptive}
A.~D. Jagtap, K.~Kawaguchi, G.~E. Karniadakis, {{Adaptive activation functions accelerate convergence in deep and physics-informed neural networks}}, Journal of Computational Physics 404 (2020) 109136.

\bibitem{jagtap2020locally}
A.~D. Jagtap, K.~Kawaguchi, G.~Em~Karniadakis, {Locally adaptive activation functions with slope recovery for deep and physics-informed neural networks}, Proceedings of the Royal Society A 476~(2239) (2020) 20200334.

\bibitem{karniadakis2005spectral}
G.~Karniadakis, S.~J. Sherwin, {Spectral/hp element methods for computational fluid dynamics}, Oxford University Press, USA, 2005.

\bibitem{howard2024finite}
A.~A. Howard, B.~Jacob, S.~H. Murphy, A.~Heinlein, P.~Stinis, {Finite basis Kolmogorov-Arnold networks: domain decomposition for data-driven and physics-informed problems}, arXiv preprint arXiv:2406.19662 (2024).

\bibitem{rigas2024adaptive}
S.~Rigas, M.~Papachristou, T.~Papadopoulos, F.~Anagnostopoulos, G.~Alexandridis, {Adaptive training of grid-dependent physics-informed kolmogorov-arnold networks}, arXiv preprint arXiv:2407.17611 (2024).

\bibitem{shuai2024physics}
H.~Shuai, F.~Li, {Physics-Informed Kolmogorov-Arnold Networks for Power System Dynamics}, arXiv preprint arXiv:2408.06650 (2024).

\bibitem{wang2024kolmogorovarnoldinformedneural}
Y.~Wang, J.~Sun, J.~Bai, C.~Anitescu, M.~S. Eshaghi, X.~Zhuang, T.~Rabczuk, Y.~Liu, {Kolmogorov Arnold Informed neural network: A physics-informed deep learning framework for solving forward and inverse problems based on Kolmogorov Arnold Networks}, arXiv preprint arXiv:2406.11045 (2024).

\bibitem{guilhoto2024deeplearningalternativeskolmogorov}
L.~F. Guilhoto, P.~Perdikaris, {Deep Learning Alternatives of the {Kolmogorov}} superposition theorem, arXiv preprint arXiv:2410.01990 (2024).

\bibitem{koenig2024kan}
B.~C. Koenig, S.~Kim, S.~Deng, Kan-odes: Kolmogorov--arnold network ordinary differential equations for learning dynamical systems and hidden physics, Computer Methods in Applied Mechanics and Engineering 432 (2024) 117397.

\bibitem{patra2024physics}
S.~Patra, S.~Panda, B.~K. Parida, M.~Arya, K.~Jacobs, D.~I. Bondar, A.~Sen, Physics informed kolmogorov-arnold neural networks for dynamical analysis via efficent-kan and wav-kan, arXiv preprint arXiv:2407.18373 (2024).

\bibitem{wang2024expressiveness}
Y.~Wang, J.~W. Siegel, Z.~Liu, T.~Y. Hou, On the expressiveness and spectral bias of kans, arXiv preprint arXiv:2410.01803 (2024).

\bibitem{liu2024kan2}
Z.~Liu, P.~Ma, Y.~Wang, W.~Matusik, M.~Tegmark, Kan 2.0: Kolmogorov-arnold networks meet science, arXiv preprint arXiv:2408.10205 (2024).

\bibitem{gao2021super}
H.~Gao, L.~Sun, J.-X. Wang, {Super-resolution and denoising of fluid flow using physics-informed convolutional neural networks without high-resolution labels}, Physics of Fluids 33~(7) (2021) 073603.

\bibitem{wandel2022spline}
N.~Wandel, M.~Weinmann, M.~Neidlin, R.~Klein, {Spline-PINN: Approaching PDEs without data using fast, physics-informed hermite-spline CNNs}, in: {Proceedings of the AAAI conference on artificial intelligence}, Vol.~36, 2022, pp. 8529--8538.

\bibitem{yang2020physics}
L.~Yang, D.~Zhang, G.~E. Karniadakis, {Physics-informed generative adversarial networks for stochastic differential equations}, SIAM Journal on Scientific Computing 42~(1) (2020) A292--A317.

\bibitem{bullwinkel2022deqgan}
B.~Bullwinkel, D.~Randle, P.~Protopapas, D.~Sondak, {{DEQGAN: learning the loss function for {PINN}}}s with generative adversarial networks, arXiv preprint arXiv:2209.07081 (2022).

\bibitem{zhang2023artificial}
Q.~Zhang, C.~Wu, A.~Kahana, Y.~Kim, Y.~Li, G.~E. Karniadakis, P.~Panda, {Artificial to Spiking Neural Networks Conversion for Scientific Machine Learning}, arXiv preprint arXiv:2308.16372 (2023).

\bibitem{zhao2023pinnsformer}
Z.~Zhao, X.~Ding, B.~A. Prakash, {Pinnsformer: A transformer-based framework for physics-informed neural networks}, arXiv preprint arXiv:2307.11833 (2023).

\bibitem{cho2022lstm}
G.~Cho, D.~Zhu, J.~J. Campbell, M.~Wang, {{An {LSTM-PINN}}} hybrid method to estimate lithium-ion battery pack temperature, IEEE Access 10 (2022) 100594--100604.

\bibitem{nathasarma2023physics}
R.~Nathasarma, B.~K. Roy, {Physics-informed long-short-term memory neural network for parameters estimation of nonlinear systems}, IEEE Transactions on Industry Applications 59~(5) (2023) 5376--5384.

\bibitem{guo2024ib}
L.~Guo, H.~Wu, Y.~Wang, W.~Zhou, T.~Zhou, {{IB-UQ}}: Information bottleneck based uncertainty quantification for neural function regression and neural operator learning, Journal of Computational Physics (2024) 113089.

\bibitem{banerjee2023survey}
C.~Banerjee, K.~Nguyen, C.~Fookes, M.~Raissi, {A survey on physics informed reinforcement learning: Review and open problems}, arXiv preprint arXiv:2309.01909 (2023).

\bibitem{ramesh2023physics}
A.~Ramesh, B.~Ravindran, {Physics-informed model-based reinforcement learning}, in: {Learning for Dynamics and Control Conference}, PMLR, 2023, pp. 26--37.

\bibitem{radaideh2021physics}
M.~I. Radaideh, I.~Wolverton, J.~Joseph, J.~J. Tusar, U.~Otgonbaatar, N.~Roy, B.~Forget, K.~Shirvan, {Physics-informed reinforcement learning optimization of nuclear assembly design}, Nuclear Engineering and Design 372 (2021) 110966.

\bibitem{jiang2024densely}
F.~Jiang, X.~Hou, M.~Xia, {Densely Multiplied Physics Informed Neural Network}, arXiv preprint arXiv:2402.04390 (2024).

\bibitem{cho2024separable}
J.~Cho, S.~Nam, H.~Yang, S.-B. Yun, Y.~Hong, E.~Park, {Separable physics-informed neural networks}, Advances in Neural Information Processing Systems 36 (2024).

\bibitem{wang2022tensor_integral}
Y.~Wang, P.~Jin, H.~Xie, {Tensor neural network and its numerical integration}, arXiv preprint arXiv:2207.02754 (2022).

\bibitem{wang2024tensor}
T.~Wang, Z.~Hu, K.~Kawaguchi, Z.~Zhang, G.~E. Karniadakis, {Tensor neural networks for high-dimensional Fokker-Planck equations}, arXiv preprint arXiv:2404.05615 (2024).

\bibitem{vemuri2024functional}
S.~K. Vemuri, T.~B{\"u}chner, J.~Niebling, J.~Denzler, {Functional Tensor Decompositions for Physics-Informed Neural Networks}, arXiv preprint arXiv:2408.13101 (2024).

\bibitem{toscano2024invivo}
J.~D. Toscano, C.~Wu, A.~Ladron-de Guevara, T.~Du, M.~Nedergaard, D.~H. Kelley, G.~E. Karniadakis, K.~Boster, {Inferring in vivo murine cerebrospinal fluid flow using artificial intelligence velocimetry with moving boundaries and uncertainty quantification}, bioRxiv (2024) 2024--08.

\bibitem{lakshminarayanan2017simple}
B.~Lakshminarayanan, A.~Pritzel, C.~Blundell, {Simple and scalable predictive uncertainty estimation using deep ensembles}, Advances in Neural Information Processing Systems 30 (2017).

\bibitem{quinonero2006machine}
J.~Qui{\~n}onero-Candela, I.~Dagan, B.~Magnini, F.~D'Alch{\'e}-Buc, {Machine Learning Challenges: Evaluating Predictive Uncertainty, Visual Object Classification, and Recognizing Textual Entailment, First Pascal Machine Learning Challenges Workshop, MLCW 2005, Southampton, UK, April 11-13, 2005, Revised Selected Papers}, Vol. 3944, Springer, 2006.

\bibitem{cawley2005estimating}
G.~C. Cawley, N.~L. Talbot, O.~Chapelle, {Estimating predictive variances with kernel ridge regression}, in: {Machine Learning Challenges Workshop}, Springer, 2005, pp. 56--77.

\bibitem{lim2022physics}
K.~L. Lim, R.~Dutta, M.~Rotaru, {Physics informed neural network using finite difference method}, in: {2022 IEEE International Conference on Systems, Man, and Cybernetics (SMC)}, IEEE, 2022, pp. 1828--1833.

\bibitem{gladstone2022fo}
R.~J. Gladstone, M.~A. Nabian, H.~Meidani, {{FO-PINN}}s: A first-order formulation for physics informed neural networks, arXiv preprint arXiv:2210.14320 (2022).

\bibitem{huang2022hompinns}
Y.~Huang, W.~Hao, G.~Lin, {{HomPINNs: Homotopy physics-informed neural networks for learning multiple solutions of nonlinear elliptic differential equations}}, Computers \& Mathematics with Applications 121 (2022) 62--73.

\bibitem{wang2022and}
S.~Wang, X.~Yu, P.~Perdikaris, {When and why {PINNs}} fail to train: A neural tangent kernel perspective, Journal of Computational Physics 449 (2022) 110768.

\bibitem{raissi2019deep}
M.~Raissi, H.~Babaee, P.~Givi, {Deep learning of turbulent scalar mixing}, Physical Review Fluids 4~(12) (2019) 124501.

\bibitem{kharazmi2021hp}
E.~Kharazmi, Z.~Zhang, G.~E. Karniadakis, {{hp-VPINNs}}: Variational physics-informed neural networks with domain decomposition, Computer Methods in Applied Mechanics and Engineering 374 (2021) 113547.

\bibitem{raissi2024forward}
M.~Raissi, {Forward--backward stochastic neural networks: deep learning of high-dimensional partial differential equations}, in: {Peter Carr Gedenkschrift: Research Advances in Mathematical Finance}, World Scientific, 2024, pp. 637--655.

\bibitem{buzaev2024hybrid}
F.~Buzaev, J.~Gao, I.~Chuprov, E.~Kazakov, {Hybrid acceleration techniques for the physics-informed neural networks: a comparative analysis}, Machine Learning 113~(6) (2024) 3675--3692.

\bibitem{mehta2019discovering}
P.~P. Mehta, G.~Pang, F.~Song, G.~E. Karniadakis, {Discovering a universal variable-order fractional model for turbulent Couette flow using a physics-informed neural network}, Fractional Calculus and Applied Analysis 22~(6) (2019) 1675--1688.

\bibitem{ren2023class}
H.~Ren, X.~Meng, R.~Liu, J.~Hou, Y.~Yu, {A class of improved fractional physics informed neural networks}, Neurocomputing 562 (2023) 126890.

\bibitem{wang2024gmc}
S.~Wang, G.~E. Karniadakis, {{GMC-PINNs}}: A new general monte carlo {PINNs} method for solving fractional partial differential equations on irregular domains, arXiv preprint arXiv:2405.00217 (2024).

\bibitem{sivalingam2024physics}
S.~Sivalingam, V.~Govindaraj, {Physics informed neural network based scheme and its error analysis for $\psi$-Caputo type fractional differential equations}, Physica Scripta 99~(9) (2024) 096002.

\bibitem{hu2024tackling_fractional}
Z.~Hu, K.~Kawaguchi, Z.~Zhang, G.~E. Karniadakis, {Tackling the Curse of Dimensionality in Fractional and Tempered Fractional PDEs with Physics-Informed Neural Networks}, arXiv preprint arXiv:2406.11708 (2024).

\bibitem{wang20222}
C.~Wang, S.~Li, D.~He, L.~Wang, {Is L2 Physics Informed Loss Always Suitable for Training Physics Informed Neural Network?}, Advances in Neural Information Processing Systems 35 (2022) 8278--8290.

\bibitem{he2023learning}
D.~He, S.~Li, W.~Shi, X.~Gao, J.~Zhang, J.~Bian, L.~Wang, T.-Y. Liu, {Learning physics-informed neural networks without stacked back-propagation}, in: {International Conference on Artificial Intelligence and Statistics}, PMLR, 2023, pp. 3034--3047.

\bibitem{wang2023expert}
S.~Wang, S.~Sankaran, H.~Wang, P.~Perdikaris, {An expert's guide to training physics-informed neural networks}, arXiv preprint arXiv:2308.08468 (2023).

\bibitem{wang2023solution}
Z.~Wang, X.~Meng, X.~Jiang, H.~Xiang, G.~E. Karniadakis, {Solution multiplicity and effects of data and eddy viscosity on {Navier-Stokes}} solutions inferred by physics-informed neural networks, arXiv preprint arXiv:2309.06010 (2023).

\bibitem{basir2022investigating}
S.~Basir, {{Investigating and Mitigating Failure Modes in Physics-informed Neural Networks ({PINNs}}}), arXiv preprint arXiv:2209.09988 (2022).

\bibitem{kharazmi2019variational}
E.~Kharazmi, Z.~Zhang, G.~E. Karniadakis, {Variational physics-informed neural networks for solving partial differential equations}, arXiv preprint arXiv:1912.00873 (2019).

\bibitem{yu2018deep}
B.~Yu, et~al., {The deep Ritz method: a deep learning-based numerical algorithm for solving variational problems}, Communications in Mathematics and Statistics 6~(1) (2018) 1--12.

\bibitem{khodayi2020varnet}
R.~Khodayi-Mehr, M.~Zavlanos, {VarNet: Variational neural networks for the solution of partial differential equations}, in: {Learning for dynamics and control}, PMLR, 2020, pp. 298--307.

\bibitem{berrone2024meshfree}
S.~Berrone, M.~Pintore, {Meshfree Variational Physics Informed Neural Networks ({MF-VPINN}}): an adaptive training strategy, arXiv preprint arXiv:2406.19831 (2024).

\bibitem{miao2023vc}
Z.~Miao, Y.~Chen, {{VC-PINN}}: Variable coefficient physics-informed neural network for forward and inverse problems of pdes with variable coefficient, Physica D: Nonlinear Phenomena 456 (2023) 133945.

\bibitem{song2024vw}
J.~Song, W.~Cao, F.~Liao, W.~Zhang, {{VW-PINNs}: A volume weighting method for PDE residuals in physics-informed neural networks}, arXiv preprint arXiv:2401.06196 (2024).

\bibitem{ghose2024fastvpinns}
D.~Ghose, T.~Anandh, S.~Ganesan, {{FastVPINNs: A fast, versatile and robust Variational PINNs framework for forward and inverse problems in science}}, in: {ICLR 2024 Workshop on AI4DifferentialEquations In Science}.

\bibitem{anandh2024efficient}
T.~Anandh, D.~Ghose, A.~Tyagi, A.~Gupta, S.~Sarkar, S.~Ganesan, {An efficient hp-Variational {PINNs} framework for incompressible Navier-Stokes equations}, arXiv preprint arXiv:2409.04143 (2024).

\bibitem{guo2022monte}
L.~Guo, H.~Wu, X.~Yu, T.~Zhou, {Monte Carlo {fPINNs}}: Deep learning method for forward and inverse problems involving high dimensional fractional partial differential equations, Computer Methods in Applied Mechanics and Engineering 400 (2022) 115523.

\bibitem{zhang2020learning}
D.~Zhang, L.~Guo, G.~E. Karniadakis, {Learning in modal space: Solving time-dependent stochastic PDEs using physics-informed neural networks}, SIAM Journal on Scientific Computing 42~(2) (2020) A639--A665.

\bibitem{hu2024score}
Z.~Hu, Z.~Zhang, G.~E. Karniadakis, K.~Kawaguchi, {{Score-fPINN: Fractional Score-Based Physics-Informed Neural Networks for High-Dimensional Fokker-Planck-Levy Equations}}, arXiv preprint arXiv:2406.11676 (2024).

\bibitem{howard2023stacked}
A.~A. Howard, S.~H. Murphy, S.~E. Ahmed, P.~Stinis, {Stacked networks improve physics-informed training: applications to neural networks and deep operator networks}, arXiv preprint arXiv:2311.06483 (2023).

\bibitem{shukla2021parallel}
K.~Shukla, A.~D. Jagtap, G.~E. Karniadakis, {Parallel physics-informed neural networks via domain decomposition}, Journal of Computational Physics 447 (2021) 110683.

\bibitem{wu2023comprehensive}
C.~Wu, M.~Zhu, Q.~Tan, Y.~Kartha, L.~Lu, {A comprehensive study of non-adaptive and residual-based adaptive sampling for physics-informed neural networks}, Computer Methods in Applied Mechanics and Engineering 403 (2023) 115671.

\bibitem{urban2024unveiling}
J.~F. Urb{\'a}n, P.~Stefanou, J.~A. Pons, {Unveiling the optimization process of Physics Informed Neural Networks: How accurate and competitive can {PINNs}} be?, arXiv preprint arXiv:2405.04230 (2024).

\bibitem{liu2024config}
Q.~Liu, M.~Chu, N.~Thuerey, {ConFIG: Towards Conflict-free Training of Physics Informed Neural Networks}, arXiv preprint arXiv:2408.11104 (2024).

\bibitem{akhter2024common}
J.~Akhter, P.~D. F{\"a}hrmann, K.~Sonntag, S.~Peitz, {Common pitfalls to avoid while using multiobjective optimization in machine learning}, arXiv preprint arXiv:2405.01480 (2024).

\bibitem{davi2022pso}
C.~Davi, U.~Braga-Neto, {PSO-PINN: Physics-informed neural networks trained with particle swarm optimization}, arXiv preprint arXiv:2202.01943 (2022).

\bibitem{jagtap2020extended}
A.~D. Jagtap, G.~E. Karniadakis, {Extended physics-informed neural networks ({XPINNs}}): A generalized space-time domain decomposition based deep learning framework for nonlinear partial differential equations, Communications in Computational Physics 28~(5) (2020).

\bibitem{hu2021extended}
Z.~Hu, A.~D. Jagtap, G.~E. Karniadakis, K.~Kawaguchi, {When do extended physics-informed neural networks ({XPINNs}}) improve generalization?, arXiv preprint arXiv:2109.09444 (2021).

\bibitem{de2024error}
T.~De~Ryck, A.~D. Jagtap, S.~Mishra, {Error estimates for physics-informed neural networks approximating the Navier--Stokes equations}, IMA Journal of Numerical Analysis 44~(1) (2024) 83--119.

\bibitem{jagtap2020conservative}
A.~D. Jagtap, E.~Kharazmi, G.~E. Karniadakis, {Conservative physics-informed neural networks on discrete domains for conservation laws: Applications to forward and inverse problems}, Computer Methods in Applied Mechanics and Engineering 365 (2020) 113028.

\bibitem{hu2023augmented}
Z.~Hu, A.~D. Jagtap, G.~E. Karniadakis, K.~Kawaguchi, {Augmented Physics-Informed Neural Networks ({APINNs}}): A gating network-based soft domain decomposition methodology, Engineering Applications of Artificial Intelligence 126 (2023) 107183.

\bibitem{moseley2023finite}
B.~Moseley, A.~Markham, T.~Nissen-Meyer, {Finite Basis Physics-Informed Neural Networks ({FBPINNs}}): a scalable domain decomposition approach for solving differential equations, Advances in Computational Mathematics 49~(4) (2023) 62.

\bibitem{dolean2024multilevel}
V.~Dolean, A.~Heinlein, S.~Mishra, B.~Moseley, {Multilevel domain decomposition-based architectures for physics-informed neural networks}, Computer Methods in Applied Mechanics and Engineering 429 (2024) 117116.

\bibitem{liu2023cv}
C.~Liu, H.~Wu, {cv-{PINN}}: Efficient learning of variational physics-informed neural network with domain decomposition, Extreme Mechanics Letters 63 (2023) 102051.

\bibitem{nguyen2022efficient}
L.~Nguyen, M.~Raissi, P.~Seshaiyer, {Efficient physics informed neural networks coupled with domain decomposition methods for solving coupled multi-physics problems}, in: {Advances in Computational Modeling and Simulation}, Springer, 2022, pp. 41--53.

\bibitem{kopanivcakova2024enhancing}
A.~Kopani{\v{c}}{\'a}kov{\'a}, H.~Kothari, G.~E. Karniadakis, R.~Krause, {Enhancing training of physics-informed neural networks using domain decomposition--based preconditioning strategies}, SIAM Journal on Scientific Computing (2024) S46--S67.

\bibitem{CMINNs}
N.~A. Daryakenari, S.~Wang, G.~Karniadakis, Cminns: Compartment model informed neural networks--unlocking drug dynamics, arXiv preprint arXiv:2409.12998 (2024).

\bibitem{wight2020solving}
C.~L. Wight, J.~Zhao, {Solving Allen-Cahn and Cahn-Hilliard equations using the adaptive physics informed neural networks}, arXiv preprint arXiv:2007.04542 (2020).

\bibitem{krishnapriyan2021characterizing}
A.~Krishnapriyan, A.~Gholami, S.~Zhe, R.~Kirby, M.~W. Mahoney, {{Characterizing possible failure modes in physics-informed neural networks}}, Advances in Neural Information Processing Systems 34 (2021) 26548--26560.

\bibitem{mattey2022novel}
R.~Mattey, S.~Ghosh, {A novel sequential method to train physics informed neural networks for Allen Cahn and Cahn Hilliard equations}, Computer Methods in Applied Mechanics and Engineering 390 (2022) 114474.

\bibitem{haitsiukevich2023improved}
K.~Haitsiukevich, A.~Ilin, {Improved training of physics-informed neural networks with model ensembles}, in: {2023 International Joint Conference on Neural Networks (IJCNN)}, IEEE, 2023, pp. 1--8.

\bibitem{chen2024leveraging2}
P.~Chen, T.~Meng, Z.~Zou, J.~Darbon, G.~E. Karniadakis, {Leveraging {Hamilton-Jacobi PDEs}} with time-dependent {Hamiltonians} for continual scientific machine learning, in: {6th Annual Learning for Dynamics \& Control Conference}, PMLR, 2024, pp. 1--12.

\bibitem{penwarden2023unified}
M.~Penwarden, A.~D. Jagtap, S.~Zhe, G.~E. Karniadakis, R.~M. Kirby, {A unified scalable framework for causal sweeping strategies for physics-informed neural networks ({PINNs}}) and their temporal decompositions, Journal of Computational Physics 493 (2023) 112464.

\bibitem{desai2021one}
S.~Desai, M.~Mattheakis, H.~Joy, P.~Protopapas, S.~Roberts, {One-shot transfer learning of physics-informed neural networks}, arXiv preprint arXiv:2110.11286 (2021).

\bibitem{liu2023adaptive}
Y.~Liu, W.~Liu, X.~Yan, S.~Guo, C.-a. Zhang, {Adaptive transfer learning for {PINN}}, Journal of Computational Physics 490 (2023) 112291.

\bibitem{xu2023transfer}
C.~Xu, B.~T. Cao, Y.~Yuan, G.~Meschke, {Transfer learning based physics-informed neural networks for solving inverse problems in engineering structures under different loading scenarios}, Computer Methods in Applied Mechanics and Engineering 405 (2023) 115852.

\bibitem{chen2024enhancing}
Y.~Chen, H.~Xiao, X.~Teng, W.~Liu, L.~Lan, {Enhancing accuracy of physically informed neural networks for nonlinear Schr{\"o}}dinger equations through multi-view transfer learning, Information Fusion 102 (2024) 102041.

\bibitem{chen2024leveraging}
P.~Chen, T.~Meng, Z.~Zou, J.~Darbon, G.~E. Karniadakis, {Leveraging multitime {Hamilton--Jacobi PDEs}} for certain scientific machine learning problems, SIAM Journal on Scientific Computing 46~(2) (2024) C216--C248.

\bibitem{zhang2024meshless}
H.~Zhang, R.~H. Chan, X.-C. Tai, {A Meshless Solver for Blood Flow Simulations in Elastic Vessels Using a Physics-Informed Neural Network}, SIAM Journal on Scientific Computing 46~(4) (2024) C479--C507.

\bibitem{meng2020composite}
X.~Meng, G.~E. Karniadakis, {A composite neural network that learns from multi-fidelity data: Application to function approximation and inverse PDE problems}, Journal of Computational Physics 401 (2020) 109020.

\bibitem{meng2021multi}
X.~Meng, H.~Babaee, G.~E. Karniadakis, {Multi-fidelity Bayesian neural networks: Algorithms and applications}, Journal of Computational Physics 438 (2021) 110361.

\bibitem{regazzoni2021physics}
F.~Regazzoni, S.~Pagani, A.~Cosenza, A.~Lombardi, A.~Quarteroni, {A physics-informed multi-fidelity approach for the estimation of differential equations parameters in low-data or large-noise regimes}, Rendiconti Lincei 32~(3) (2021) 437--470.

\bibitem{wang2024multi}
Y.~Wang, C.-Y. Lai, {Multi-stage neural networks: Function approximator of machine precision}, Journal of Computational Physics 504 (2024) 112865.

\bibitem{heinlein2024multifidelity}
A.~Heinlein, A.~A. Howard, D.~Beecroft, P.~Stinis, {Multifidelity domain decomposition-based physics-informed neural networks for time-dependent problems}, arXiv preprint arXiv:2401.07888 (2024).

\bibitem{penwarden2022multifidelity}
M.~Penwarden, S.~Zhe, A.~Narayan, R.~M. Kirby, {Multifidelity modeling for physics-informed neural networks (PINNs)}, Journal of Computational Physics 451 (2022) 110844.

\bibitem{liu2001monte}
J.~S. Liu, J.~S. Liu, {Monte Carlo strategies in scientific computing}, Vol.~10, Springer, 2001.

\bibitem{boster2023artificial}
K.~A. Boster, S.~Cai, A.~Ladr{\'o}n-de Guevara, J.~Sun, X.~Zheng, T.~Du, J.~H. Thomas, M.~Nedergaard, G.~E. Karniadakis, D.~H. Kelley, {Artificial intelligence velocimetry reveals in vivo flow rates, pressure gradients, and shear stresses in murine perivascular flows}, Proceedings of the National Academy of Sciences 120~(14) (2023) e2217744120.

\bibitem{xiang2022self}
Z.~Xiang, W.~Peng, X.~Liu, W.~Yao, {Self-adaptive loss balanced Physics-informed neural networks}, Neurocomputing 496 (2022) 11--34.

\bibitem{liu2021dual}
D.~Liu, Y.~Wang, {{A Dual-Dimer method for training physics-constrained neural networks with minimax architecture}}, Neural Networks 136 (2021) 112--125.

\bibitem{mcclenny2023self}
L.~D. McClenny, U.~M. Braga-Neto, {Self-adaptive physics-informed neural networks}, Journal of Computational Physics 474 (2023) 111722.

\bibitem{zhang2023dasa}
G.~Zhang, H.~Yang, F.~Zhu, Y.~Chen, et~al., {{DASA-PINNs: Differentiable Adversarial Self-Adaptive Pointwise Weighting Scheme for Physics-Informed Neural Networks}}, SSRN (2023).

\bibitem{basir2022physics}
S.~Basir, I.~Senocak, {Physics and equality constrained artificial neural networks: Application to forward and inverse problems with multi-fidelity data fusion}, Journal of Computational Physics 463 (2022) 111301.

\bibitem{basir2023adaptive}
S.~Basir, I.~Senocak, {An adaptive augmented Lagrangian method for training physics and equality constrained artificial neural networks}, arXiv preprint arXiv:2306.04904 (2023).

\bibitem{son2023enhanced}
H.~Son, S.~W. Cho, H.~J. Hwang, {Enhanced Physics-Informed Neural Networks with Augmented Lagrangian Relaxation Method ({AL-PINNs}}), Neurocomputing (2023) 126424.

\bibitem{anagnostopoulos2023residual}
S.~J. Anagnostopoulos, J.~D. Toscano, N.~Stergiopulos, G.~E. Karniadakis, {Residual-based attention and connection to information bottleneck theory in {PINNs}}, arXiv preprint arXiv:2307.00379 (2023).

\bibitem{song2024loss}
Y.~Song, H.~Wang, H.~Yang, M.~L. Taccari, X.~Chen, {Loss-attentional physics-informed neural networks}, Journal of Computational Physics 501 (2024) 112781.

\bibitem{ramirez2024residual}
I.~Ramirez, J.~Pino, D.~Pardo, M.~Sanz, L.~del Rio, A.~Ortiz, K.~Morozovska, J.~I. Aizpurua, {Residual-based Attention Physics-informed Neural Networks for Efficient Spatio-Temporal Lifetime Assessment of Transformers Operated in Renewable Power Plants}, arXiv preprint arXiv:2405.06443 (2024).

\bibitem{chen2024self}
W.~Chen, A.~A. Howard, P.~Stinis, {Self-adaptive weights based on balanced residual decay rate for physics-informed neural networks and deep operator networks}, arXiv preprint arXiv:2407.01613 (2024).

\bibitem{lu2021deepxde}
L.~Lu, X.~Meng, Z.~Mao, G.~E. Karniadakis, {DeepXDE: A deep learning library for solving differential equations}, SIAM Review 63~(1) (2021) 208--228.

\bibitem{daw2022rethinking}
A.~Daw, J.~Bu, S.~Wang, P.~Perdikaris, A.~Karpatne, {Rethinking the importance of sampling in physics-informed neural networks}, arXiv preprint arXiv:2207.02338 (2022).

\bibitem{gao2023active}
W.~Gao, C.~Wang, {Active learning based sampling for high-dimensional nonlinear partial differential equations}, Journal of Computational Physics 475 (2023) 111848.

\bibitem{tang2021deep}
K.~Tang, X.~Wan, C.~Yang, {DAS: A deep adaptive sampling method for solving partial differential equations}, arXiv preprint arXiv:2112.14038 (2021).

\bibitem{peng2022rang}
W.~Peng, W.~Zhou, X.~Zhang, W.~Yao, Z.~Liu, {Rang: A residual-based adaptive node generation method for physics-informed neural networks}, arXiv preprint arXiv:2205.01051 (2022).

\bibitem{zeng2022adaptive}
S.~Zeng, Z.~Zhang, Q.~Zou, {Adaptive deep neural networks methods for high-dimensional partial differential equations}, Journal of Computational Physics 463 (2022) 111232.

\bibitem{hanna2022residual}
J.~M. Hanna, J.~V. Aguado, S.~Comas-Cardona, R.~Askri, D.~Borzacchiello, {Residual-based adaptivity for two-phase flow simulation in porous media using physics-informed neural networks}, Computer Methods in Applied Mechanics and Engineering 396 (2022) 115100.

\bibitem{subramanian2023adaptive}
S.~Subramanian, R.~M. Kirby, M.~W. Mahoney, A.~Gholami, {Adaptive self-supervision algorithms for physics-informed neural networks}, in: {ECAI 2023}, IOS Press, 2023, pp. 2234--2241.

\bibitem{nabian2021efficient}
M.~A. Nabian, R.~J. Gladstone, H.~Meidani, {Efficient training of physics-informed neural networks via importance sampling}, Computer-Aided Civil and Infrastructure Engineering 36~(8) (2021) 962--977.

\bibitem{daw2022mitigating}
A.~Daw, J.~Bu, S.~Wang, P.~Perdikaris, A.~Karpatne, {Mitigating propagation failures in physics-informed neural networks using retain-resample-release (r3) sampling}, arXiv preprint arXiv:2207.02338 (2022).

\bibitem{gao2023failure}
Z.~Gao, L.~Yan, T.~Zhou, {Failure-informed adaptive sampling for {PINNs}}, SIAM Journal on Scientific Computing 45~(4) (2023) A1971--A1994.

\bibitem{gao2023failure2}
Z.~Gao, T.~Tang, L.~Yan, T.~Zhou, {Failure-informed adaptive sampling for {PINNs}}, part ii: combining with re-sampling and subset simulation, Communications on Applied Mathematics and Computation (2023) 1--22.

\bibitem{zhou2023stochastic}
K.~Zhou, J.~Li, J.~Hong, S.~J. Grauer, {Stochastic particle advection velocimetry (SPAV): theory, simulations, and proof-of-concept experiments}, Measurement Science and Technology 34~(6) (2023) 065302.

\bibitem{nocedal1999numerical}
J.~Nocedal, S.~J. Wright, {Numerical optimization}, Springer, 1999.

\bibitem{kingma2014adam}
D.~P. Kingma, J.~Ba, {Adam: A method for stochastic optimization}, arXiv preprint arXiv:1412.6980 (2014).

\bibitem{liu1989limited}
D.~C. Liu, J.~Nocedal, {On the limited memory {BFGS}} method for large scale optimization, Mathematical Programming 45~(1) (1989) 503--528.

\bibitem{lu2023nsga}
B.~Lu, C.~Moya, G.~Lin, {{NSGA-PINN}}: a multi-objective optimization method for physics-informed neural network training, Algorithms 16~(4) (2023) 194.

\bibitem{zhou2023generic}
T.~Zhou, X.~Zhang, E.~L. Droguett, A.~Mosleh, {A generic physics-informed neural network-based framework for reliability assessment of multi-state systems}, Reliability Engineering \& System Safety 229 (2023) 108835.

\bibitem{yao2023multiadam}
J.~Yao, C.~Su, Z.~Hao, S.~Liu, H.~Su, J.~Zhu, {Multiadam: Parameter-wise scale-invariant optimizer for multiscale training of physics-informed neural networks}, in: {International Conference on Machine Learning}, PMLR, 2023, pp. 39702--39721.

\bibitem{fang2023ensemble}
Z.~Fang, S.~Wang, P.~Perdikaris, {Ensemble learning for physics informed neural networks: A gradient boosting approach}, arXiv preprint arXiv:2302.13143 (2023).

\bibitem{cyr2020robust}
E.~C. Cyr, M.~A. Gulian, R.~G. Patel, M.~Perego, N.~A. Trask, {Robust training and initialization of deep neural networks: An adaptive basis viewpoint}, in: {Mathematical and Scientific Machine Learning}, PMLR, 2020, pp. 512--536.

\bibitem{ainsworth2021plateau}
M.~Ainsworth, Y.~Shin, Plateau phenomenon in gradient descent training of relu networks: Explanation, quantification, and avoidance, SIAM Journal on Scientific Computing 43~(5) (2021) A3438--A3468.

\bibitem{ainsworth2022active}
M.~Ainsworth, Y.~Shin, Active neuron least squares: A training method for multivariate rectified neural networks, SIAM Journal on Scientific Computing 44~(4) (2022) A2253--A2275.

\bibitem{jnini2024gauss}
A.~Jnini, F.~Vella, M.~Zeinhofer, {Gauss-Newton Natural Gradient Descent for Physics-Informed Computational Fluid Dynamics}, arXiv preprint arXiv:2402.10680 (2024).

\bibitem{rathore2024challenges}
P.~Rathore, W.~Lei, Z.~Frangella, L.~Lu, M.~Udell, {Challenges in training {PINNs}}: A loss landscape perspective, arXiv preprint arXiv:2402.01868 (2024).

\bibitem{muller2023achieving}
J.~M{\"u}ller, M.~Zeinhofer, Achieving high accuracy with pinns via energy natural gradient descent, in: International Conference on Machine Learning, PMLR, 2023, pp. 25471--25485.

\bibitem{dangel2024kronecker}
F.~Dangel, J.~M{\"u}ller, M.~Zeinhofer, Kronecker-factored approximate curvature for physics-informed neural networks, arXiv preprint arXiv:2405.15603 (2024).

\bibitem{lee2024two}
Y.~Lee, A.~Kopani{\v{c}}{\'a}kov{\'a}, G.~E. Karniadakis, {Two-level overlapping additive Schwarz preconditioner for training scientific machine learning applications}, arXiv preprint arXiv:2406.10997 (2024).

\bibitem{OpenAI2024DALLE}
OpenAI, {DALL·E 3}, \url{https://openai.com/dall-e} (2024).

\bibitem{yazdani2020systems}
A.~Yazdani, L.~Lu, M.~Raissi, G.~E. Karniadakis, {Systems biology informed deep learning for inferring parameters and hidden dynamics}, PLOS Computational Biology 16~(11) (2020).

\bibitem{daneker2022systems}
M.~Daneker, Z.~Zhang, G.~E. Karniadakis, L.~Lu, {Systems Biology: Identifiability analysis and parameter identification via systems-biology informed neural networks}, arXiv preprint arXiv:2202.01723 (2022).

\bibitem{jo2024density}
H.~Jo, H.~Hong, H.~J. Hwang, W.~Chang, J.~K. Kim, {Density physics-informed neural networks reveal sources of cell heterogeneity in signal transduction}, Patterns 5~(2) (2024).

\bibitem{sahli2020physics}
F.~Sahli~Costabal, Y.~Yang, P.~Perdikaris, D.~E. Hurtado, E.~Kuhl, {Physics-informed neural networks for cardiac activation mapping}, Frontiers in Physics 8 (2020) 42.

\bibitem{ruiz2022physics}
C.~Ruiz~Herrera, T.~Grandits, G.~Plank, P.~Perdikaris, F.~Sahli~Costabal, S.~Pezzuto, {Physics-informed neural networks to learn cardiac fiber orientation from multiple electroanatomical maps}, Engineering with Computers 38~(5) (2022) 3957--3973.

\bibitem{QIAN2024106732}
Y.~Qian, G.~Zhu, Z.~Zhang, S.~Modepalli, Y.~Zheng, X.~Zheng, G.~Frydman, H.~Li, {Coagulo-Net: Enhancing the mathematical modeling of blood coagulation using physics-informed neural networks}, Neural Networks 180 (2024) 106732.

\bibitem{chen2023tgm}
Q.~Chen, Q.~Ye, W.~Zhang, H.~Li, X.~Zheng, {TGM-Nets: A deep learning framework for enhanced forecasting of tumor growth by integrating imaging and modeling}, Engineering Applications of Artificial Intelligence 126 (2023) 106867.

\bibitem{herrero2022ep}
C.~Herrero~Martin, A.~Oved, R.~A. Chowdhury, E.~Ullmann, N.~S. Peters, A.~A. Bharath, M.~Varela, {{EP-PINNs}}: Cardiac electrophysiology characterisation using physics-informed neural networks, Frontiers in Cardiovascular Medicine 8 (2022) 768419.

\bibitem{goswami2021study}
K.~Goswami, A.~Sharma, M.~Pruthi, R.~Gupta, {Study of Drug Assimilation in Human System using Physics Informed Neural Networks}, arXiv preprint arXiv:2110.05531 (2021).

\bibitem{podina2024learning}
L.~Podina, A.~Ghodsi, M.~Kohandel, {Learning Chemotherapy Drug Action via Universal Physics-Informed Neural Networks}, arXiv preprint arXiv:2404.08019 (2024).

\bibitem{math12081195}
J.~A. Rodrigues, {Using Physics-Informed Neural Networks ({PINNs}}) for tumor cell growth modeling, Mathematics 12~(8) (2024).

\bibitem{soukarieh2024hypersbinn}
I.~Soukarieh, G.~Hessler, H.~Minoux, M.~Mohr, F.~Schmidt, J.~Wenzel, P.~Barbillon, H.~Gangloff, P.~Gloaguen, {HyperSBINN: A Hypernetwork-Enhanced Systems Biology-Informed Neural Network for Efficient Drug Cardiosafety Assessment}, arXiv preprint arXiv:2408.14266 (2024).

\bibitem{chiu2024characterisation}
C.-E. Chiu, A.~L. Pinto, R.~A. Chowdhury, K.~Christensen, M.~Varela, {Characterisation of Anti-Arrhythmic Drug Effects on Cardiac Electrophysiology using Physics-Informed Neural Networks}, arXiv preprint arXiv:2403.08439 (2024).

\bibitem{kissas2020machine}
G.~Kissas, Y.~Yang, E.~Hwuang, W.~R. Witschey, J.~A. Detre, P.~Perdikaris, {Machine learning in cardiovascular flows modeling: Predicting arterial blood pressure from non-invasive 4D flow MRI data using physics-informed neural networks}, Computer Methods in Applied Mechanics and Engineering 358 (2020) 112623.

\bibitem{sun2020surrogate}
L.~Sun, H.~Gao, S.~Pan, J.-X. Wang, {Surrogate modeling for fluid flows based on physics-constrained deep learning without simulation data}, Computer Methods in Applied Mechanics and Engineering 361 (2020) 112732.

\bibitem{arzani2021uncovering}
A.~Arzani, J.-X. Wang, R.~M. D'Souza, {Uncovering near-wall blood flow from sparse data with physics-informed neural networks}, Physics of Fluids 33~(7) (2021) 071905.

\bibitem{daneker2024transfer}
M.~Daneker, S.~Cai, Y.~Qian, E.~Myzelev, A.~Kumbhat, H.~Li, L.~Lu, Transfer learning on physics-informed neural networks for tracking the hemodynamics in the evolving false lumen of dissected aorta, Nexus 1~(2) (2024).

\bibitem{yin2021non}
M.~Yin, X.~Zheng, J.~D. Humphrey, G.~E. Karniadakis, {Non-invasive inference of thrombus material properties with physics-informed neural networks}, Computer Methods in Applied Mechanics and Engineering 375 (2021) 113603.

\bibitem{fathi2020super}
M.~F. Fathi, I.~Perez-Raya, A.~Baghaie, P.~Berg, G.~Janiga, A.~Arzani, R.~M. D’Souza, {Super-resolution and denoising of 4D-flow MRI using physics-informed deep neural nets}, Computer Methods and Programs in Biomedicine 197 (2020) 105729.

\bibitem{liu2020generic}
M.~Liu, L.~Liang, W.~Sun, {A generic physics-informed neural network-based constitutive model for soft biological tissues}, Computer Methods in Applied Mechanics and Engineering 372 (2020) 113402.

\bibitem{lagergren2020biologically}
J.~H. Lagergren, J.~T. Nardini, R.~E. Baker, M.~J. Simpson, K.~B. Flores, {Biologically-informed neural networks guide mechanistic modeling from sparse experimental data}, PLOS Computational Biology 16~(12) (2020) e1008462.

\bibitem{SAINZDEMENA2024104092}
D.~Sainz-DeMena, M.~Pérez, J.~García-Aznar, {Exploring the potential of Physics-Informed Neural Networks to extract vascularization data from DCE-MRI in the presence of diffusion}, Medical Engineering \& Physics 123 (2024) 104092.

\bibitem{Awojoyogbe2024}
B.~O. Awojoyogbe, M.~O. Dada, {Simulation of Temperature Distribution in Biological Tissues Using Physics-Informed Neural Networks}, in: {Digital Molecular Magnetic Resonance Imaging}, Springer, 2024, pp. 217--228.

\bibitem{caforio2024physics}
F.~Caforio, F.~Regazzoni, S.~Pagani, E.~Karabelas, C.~Augustin, G.~Haase, G.~Plank, A.~Quarteroni, {Physics-informed neural network estimation of material properties in soft tissue nonlinear biomechanical models}, Computational Mechanics (2024) 1--27.

\bibitem{wu2024identifying}
W.~Wu, M.~Daneker, K.~T. Turner, M.~A. Jolley, L.~Lu, {Identifying heterogeneous micromechanical properties of biological tissues via physics-informed neural networks}, arXiv preprint arXiv:2402.10741 (2024).

\bibitem{Ragoza2023}
M.~Ragoza, K.~Batmanghelich, {Physics-Informed Neural Networks for Tissue Elasticity Reconstruction in Magnetic Resonance Elastography}, Medical Image Computing and Computer-Assisted Intervention (MICCAI) 14229 (2023) 333--343.

\bibitem{movahhedi2023predicting}
M.~Movahhedi, X.-Y. Liu, B.~Geng, C.~Elemans, Q.~Xue, J.-X. Wang, X.~Zheng, {Predicting 3D soft tissue dynamics from 2D imaging using physics informed neural networks}, Communications Biology 6~(1) (2023) 541.

\bibitem{ling2024physics}
H.~J. Ling, S.~Bru, J.~Puig, F.~Vix{\`e}ge, S.~Mendez, F.~Nicoud, P.-Y. Courand, O.~Bernard, D.~Garcia, {Physics-Guided Neural Networks for Intraventricular Vector Flow Mapping}, IEEE Transactions on Ultrasonics, Ferroelectrics, and Frequency Control (2024).

\bibitem{sel2023physics}
K.~Sel, A.~Mohammadi, R.~I. Pettigrew, R.~Jafari, {Physics-informed neural networks for modeling physiological time series for cuffless blood pressure estimation}, npj Digital Medicine 6~(1) (2023) 110.

\bibitem{du2023evaluation}
J.~F. Du~Toit, R.~Laubscher, {Evaluation of Physics-Informed Neural Network Solution Accuracy and Efficiency for Modeling Aortic Transvalvular Blood Flow}, Mathematical and Computational Applications 28~(2) (2023) 62.

\bibitem{jagtap2023coolpinns}
N.~V. Jagtap, M.~K. Mudunuru, K.~B. Nakshatrala, {Cool{PINNs}}: A physics-informed neural network modeling of active cooling in vascular systems, Applied Mathematical Modelling 122 (2023) 265--287.

\bibitem{heldmann2022pinn}
F.~Heldmann, S.~Berkhahn, M.~Ehrhardt, K.~Klamroth, Pinn training using biobjective optimization: The trade-off between data loss and residual loss, Journal of Computational Physics 488 (2023) 112211.

\bibitem{treibert2022physics}
S.~Treibert, M.~Ehrhardt, {A Physics-Informed Neural Network to Model COVID-19 Infection and Hospitalization Scenarios} (2022).

\bibitem{epiii}
C.~Millevoi, D.~Pasetto, M.~Ferronato, {A Physics-Informed Neural Network approach for compartmental epidemiological models}, PLOS Computational Biology 20 (2024) 1--29.

\bibitem{shaier2022data}
S.~Shaier, M.~Raissi, P.~Seshaiyer, {Data-driven approaches for predicting spread of infectious diseases through DINNs: Disease Informed Neural Networks}, Letters in Biomathematics 9~(1) (2022) 71--105.

\bibitem{nguyen2022modeling}
L.~Nguyen, M.~Raissi, P.~Seshaiyer, {Modeling, Analysis and Physics Informed Neural Network approaches for studying the dynamics of COVID-19 involving human-human and human-pathogen interaction}, Computational and Mathematical Biophysics 10~(1) (2022) 1--17.

\bibitem{schiassi2021physics}
E.~Schiassi, M.~De~Florio, A.~D’ambrosio, D.~Mortari, R.~Furfaro, {Physics-informed neural networks and functional interpolation for data-driven parameters discovery of epidemiological compartmental models}, Mathematics 9~(17) (2021) 2069.

\bibitem{kharazmi2021identifiability}
E.~Kharazmi, M.~Cai, X.~Zheng, Z.~Zhang, G.~Lin, G.~E. Karniadakis, {Identifiability and predictability of integer-and fractional-order epidemiological models using physics-informed neural networks}, Nature Computational Science 1~(11) (2021) 744--753.

\bibitem{cai2024physics}
S.~Cai, C.~Gray, G.~E. Karniadakis, {Physics-Informed Neural Networks Enhanced Particle Tracking Velocimetry: An Example for Turbulent Jet Flow}, IEEE Transactions on Instrumentation and Measurement (2024).

\bibitem{calicchia2022reconstructing}
M.~Calicchia, R.~Ni, R.~Mittal, J.-H. Seo, {Reconstructing the pressure field around an undulating body using a physics-informed neural network}, Bulletin of the American Physical Society (2022).

\bibitem{jagtap2022physics}
A.~D. Jagtap, Z.~Mao, N.~Adams, G.~E. Karniadakis, {Physics-informed neural networks for inverse problems in supersonic flows}, arXiv preprint arXiv:2202.11821 (2022).

\bibitem{kag2022physics}
V.~Kag, K.~Seshasayanan, V.~Gopinath, {Physics-informed data based neural networks for two-dimensional turbulence}, Physics of Fluids 34~(5) (2022) 055130.

\bibitem{reyes2021learning}
B.~Reyes, A.~A. Howard, P.~Perdikaris, A.~M. Tartakovsky, {Learning unknown physics of non-Newtonian fluids}, Physical Review Fluids 6~(7) (2021) 073301.

\bibitem{raissi2019vortex}
M.~Raissi, Z.~Wang, M.~S. Triantafyllou, G.~E. Karniadakis, {Deep learning of vortex-induced vibrations}, Journal of Fluid Mechanics 861 (2019) 119--137.

\bibitem{di2023reconstructing}
P.~C. Di~Leoni, L.~Agasthya, M.~Buzzicotti, L.~Biferale, {Reconstructing Rayleigh-Benard flows out of temperature-only measurements using Physics-Informed Neural Networks}, arXiv preprint arXiv:2301.07769 (2023).

\bibitem{de2021physics}
M.~De~Florio, E.~Schiassi, B.~D. Ganapol, R.~Furfaro, {Physics-informed neural networks for rarefied-gas dynamics: Thermal creep flow in the Bhatnagar--Gross--Krook approximation}, Physics of Fluids 33~(4) (2021) 047110.

\bibitem{thakur2022viscoelasticnet}
S.~Thakur, M.~Raissi, A.~M. Ardekani, {ViscoelasticNet: A physics informed neural network framework for stress discovery and model selection}, arXiv preprint arXiv:2209.06972 (2022).

\bibitem{yang2019predictive}
X.~Yang, S.~Zafar, J.-X. Wang, H.~Xiao, {Predictive large-eddy-simulation wall modeling via physics-informed neural networks}, Physical Review Fluids 4~(3) (2019) 034602.

\bibitem{song2021solving}
C.~Song, T.~Alkhalifah, U.~B. Waheed, {Solving the frequency-domain acoustic VTI wave equation using physics-informed neural networks}, Geophysical Journal International 225~(2) (2021) 846--859.

\bibitem{lou2021physics}
Q.~Lou, X.~Meng, G.~E. Karniadakis, {Physics-informed neural networks for solving forward and inverse flow problems via the Boltzmann-BGK formulation}, Journal of Computational Physics 447 (2021) 110676.

\bibitem{erichson2019physics}
N.~B. Erichson, M.~Muehlebach, M.~W. Mahoney, {Physics-informed autoencoders for Lyapunov-stable fluid flow prediction}, arXiv preprint arXiv:1905.10866 (2019).

\bibitem{calicchia2023reconstructing}
M.~A. Calicchia, R.~Mittal, J.-H. Seo, R.~Ni, {Reconstructing the pressure field around swimming fish using a physics-informed neural network}, Journal of Experimental Biology 226~(8) (2023).

\bibitem{shukla2020physics}
K.~Shukla, P.~C. Di~Leoni, J.~Blackshire, D.~Sparkman, G.~E. Karniadakis, {Physics-informed neural network for ultrasound nondestructive quantification of surface breaking cracks}, Journal of Nondestructive Evaluation 39 (2020) 1--20.

\bibitem{shukla2021physics}
K.~Shukla, A.~D. Jagtap, J.~L. Blackshire, D.~Sparkman, G.~E. Karniadakis, {A physics-informed neural network for quantifying the microstructural properties of polycrystalline nickel using ultrasound data: A promising approach for solving inverse problems}, IEEE Signal Processing Magazine 39~(1) (2021) 68--77.

\bibitem{mahmoudabadbozchelou2022nn}
M.~Mahmoudabadbozchelou, G.~E. Karniadakis, S.~Jamali, {nn-{PINN}}s: Non-newtonian physics-informed neural networks for complex fluid modeling, Soft Matter 18~(1) (2022) 172--185.

\bibitem{zhang2022analyses}
E.~Zhang, M.~Dao, G.~E. Karniadakis, S.~Suresh, {Analyses of internal structures and defects in materials using physics-informed neural networks}, Science Advances 8~(7) (2022).

\bibitem{li2021physics}
W.~Li, M.~Z. Bazant, J.~Zhu, {A physics-guided neural network framework for elastic plates: Comparison of governing equations-based and energy-based approaches}, Computer Methods in Applied Mechanics and Engineering 383 (2021) 113933.

\bibitem{bastek2022physics}
J.-H. Bastek, D.~M. Kochmann, {Physics-Informed Neural Networks for Shell Structures}, arXiv preprint arXiv:2207.14291 (2022).

\bibitem{goswami2020transfer}
S.~Goswami, C.~Anitescu, S.~Chakraborty, T.~Rabczuk, {Transfer learning enhanced physics informed neural network for phase-field modeling of fracture}, Theoretical and Applied Fracture Mechanics 106 (2020) 102447.

\bibitem{zhang2021physics}
Z.~Zhang, G.~X. Gu, {Physics-informed deep learning for digital materials}, Theoretical and Applied Mechanics Letters 11~(1) (2021) 100220.

\bibitem{pantidis2022integrated}
P.~Pantidis, M.~E. Mobasher, {Integrated Finite Element Neural Network (I-FENN) for non-local continuum damage mechanics}, arXiv preprint arXiv:2207.09908 (2022).

\bibitem{zhang2020physics}
E.~Zhang, M.~Yin, G.~E. Karniadakis, {Physics-informed neural networks for nonhomogeneous material identification in elasticity imaging}, arXiv preprint arXiv:2009.04525 (2020).

\bibitem{zhang2020physicsLSTM}
R.~Zhang, Y.~Liu, H.~Sun, {Physics-informed multi-LSTM networks for metamodeling of nonlinear structures}, Computer Methods in Applied Mechanics and Engineering 369 (2020) 113226.

\bibitem{bin2021pinneik}
U.~bin Waheed, E.~Haghighat, T.~Alkhalifah, C.~Song, Q.~Hao, {{PINNeik}}: Eikonal solution using physics-informed neural networks, Computers \& Geosciences 155 (2021) 104833.

\bibitem{ross2021hyposvi}
Z.~Ross, J.~Smith, K.~Azizzadenesheli, J.~Muir, {HypoSVI: Hypocenter inversion with stein variational inference and physics informed neural networks}, in: {AGU Fall Meeting Abstracts}, Vol. 2021, 2021, pp. S33B--08.

\bibitem{ihunde2022application}
T.~A. Ihunde, O.~Olorode, {Application of physics informed neural networks to compositional modeling}, Journal of Petroleum Science and Engineering 211 (2022) 110175.

\bibitem{nazari2022physics}
L.~F. Nazari, E.~Camponogara, L.~O. Seman, {Physics-Informed Neural Networks for Modeling Water Flows in a River Channel}, IEEE Transactions on Artificial Intelligence (2022).

\bibitem{rasht2022physics}
M.~Rasht-Behesht, C.~Huber, K.~Shukla, G.~E. Karniadakis, {Physics-Informed Neural Networks ({PINNs}}) for wave propagation and full waveform inversions, Journal of Geophysical Research: Solid Earth 127~(5) (2022).

\bibitem{zheng2020physics}
Q.~Zheng, L.~Zeng, G.~E. Karniadakis, {Physics-informed semantic inpainting: Application to geostatistical modeling}, Journal of Computational Physics 419 (2020) 109676.

\bibitem{abreu2021study}
E.~Abreu, J.~B. Florindo, {A study on a feedforward neural network to solve partial differential equations in hyperbolic-transport problems}, in: {International Conference on Computational Science}, Springer, 2021, pp. 398--411.

\bibitem{almajid2022prediction}
M.~M. Almajid, M.~O. Abu-Al-Saud, {Prediction of porous media fluid flow using physics informed neural networks}, Journal of Petroleum Science and Engineering 208 (2022) 109205.

\bibitem{he2020physics}
Q.~He, D.~Barajas-Solano, G.~Tartakovsky, A.~M. Tartakovsky, {Physics-informed neural networks for multiphysics data assimilation with application to subsurface transport}, Advances in Water Resources 141 (2020) 103610.

\bibitem{tartakovsky2020physics}
A.~M. Tartakovsky, C.~O. Marrero, P.~Perdikaris, G.~D. Tartakovsky, D.~Barajas-Solano, {Physics-informed deep neural networks for learning parameters and constitutive relationships in subsurface flow problems}, Water Resources Research 56~(5) (2020) e2019WR026731.

\bibitem{haghighat2022physics}
E.~Haghighat, D.~Amini, R.~Juanes, {Physics-informed neural network simulation of multiphase poroelasticity using stress-split sequential training}, Computer Methods in Applied Mechanics and Engineering 397 (2022) 115141.

\bibitem{amini2022inverse}
D.~Amini, E.~Haghighat, R.~Juanes, {Inverse modeling of nonisothermal multiphase poromechanics using physics-informed neural networks}, arXiv preprint arXiv:2209.03276 (2022).

\bibitem{amini2022physics}
D.~Amini, E.~Haghighat, R.~Juanes, {Physics-informed neural network solution of thermo-hydro-mechanical (THM) processes in porous media}, arXiv preprint arXiv:2203.01514 (2022).

\bibitem{giampaolo2022physics}
F.~Giampaolo, M.~De~Rosa, P.~Qi, S.~Izzo, S.~Cuomo, {Physics-informed neural networks approach for 1D and 2D Gray-Scott systems}, Advanced Modeling and Simulation in Engineering Sciences 9~(1) (2022) 1--17.

\bibitem{nicodemus2022physics}
J.~Nicodemus, J.~Kneifl, J.~Fehr, B.~Unger, {Physics-informed neural networks-based model predictive control for multi-link manipulators}, IFAC-PapersOnLine 55~(20) (2022) 331--336.

\bibitem{xu2022physics}
P.-F. Xu, C.-B. Han, H.-X. Cheng, C.~Cheng, T.~Ge, {A physics-informed neural network for the prediction of unmanned surface vehicle dynamics}, Journal of Marine Science and Engineering 10~(2) (2022) 148.

\bibitem{antonelo2021physics}
E.~A. Antonelo, E.~Camponogara, L.~O. Seman, E.~R. de~Souza, J.~P. Jordanou, J.~F. Hubner, {Physics-informed neural nets for control of dynamical systems}, arXiv preprint arXiv:2104.02556 (2021).

\bibitem{sanyal2022ramp}
S.~Sanyal, K.~Roy, {RAMP-Net: A Robust Adaptive MPC for Quadrotors via Physics-informed Neural Network}, arXiv preprint arXiv:2209.09025 (2022).

\bibitem{gu2024physics}
W.~Gu, S.~Primatesta, A.~Rizzo, {Physics-informed Neural Network for Quadrotor Dynamical Modeling}, Robotics and Autonomous Systems 171 (2024) 104569.

\bibitem{liu2024physics}
J.~Liu, P.~Borja, C.~Della~Santina, {Physics-Informed Neural Networks to Model and Control Robots: A Theoretical and Experimental Investigation}, Advanced Intelligent Systems 6~(5) (2024) 2300385.

\bibitem{wang2024pinn}
X.~Wang, J.~J. Dabrowski, J.~Pinskier, L.~Liow, V.~Viswanathan, R.~Scalzo, D.~Howard, {PINN-Ray: A Physics-Informed Neural Network to Model Soft Robotic Fin Ray Fingers}, arXiv preprint arXiv:2407.08222 (2024).

\bibitem{ni2023progressive}
{Progressive Learning for Physics-informed Neural Motion Planning}, author={Ni, Ruiqi and Qureshi, Ahmed H}, arXiv preprint arXiv:2306.00616 (2023).

\bibitem{mo2021physics}
Z.~Mo, R.~Shi, X.~Di, {A physics-informed deep learning paradigm for car-following models}, Transportation Research part C: Emerging Technologies 130 (2021) 103240.

\bibitem{kim2022physics}
T.~Kim, H.~Lee, W.~Lee, {Physics Embedded Neural Network Vehicle Model and Applications in Risk-Aware Autonomous Driving Using Latent Features}, in: {2022 IEEE/RSJ International Conference on Intelligent Robots and Systems (IROS)}, IEEE, 2022, pp. 4182--4189.

\bibitem{hennigh2021nvidia}
O.~Hennigh, S.~Narasimhan, M.~A. Nabian, A.~Subramaniam, K.~Tangsali, Z.~Fang, M.~Rietmann, W.~Byeon, S.~Choudhry, {NVIDIA SimNet™: An AI-accelerated multi-physics simulation framework}, in: {International Conference on Computational Science}, Springer, 2021, pp. 447--461.

\bibitem{zhu2021machine}
Q.~Zhu, Z.~Liu, J.~Yan, {Machine learning for metal additive manufacturing: predicting temperature and melt pool fluid dynamics using physics-informed neural networks}, Computational Mechanics 67~(2) (2021) 619--635.

\bibitem{zobeiry2021physics}
N.~Zobeiry, K.~D. Humfeld, {A physics-informed machine learning approach for solving heat transfer equation in advanced manufacturing and engineering applications}, Engineering Applications of Artificial Intelligence 101 (2021) 104232.

\bibitem{bora2021neural}
A.~Bora, W.~Dai, J.~P. Wilson, J.~C. Boyt, {Neural network method for solving parabolic two-temperature microscale heat conduction in double-layered thin films exposed to ultrashort-pulsed lasers}, International Journal of Heat and Mass Transfer 178 (2021) 121616.

\bibitem{bora2022neural}
A.~Bora, W.~Dai, J.~P. Wilson, J.~C. Boyt, S.~L. Sobolev, {Neural network method for solving nonlocal two-temperature nanoscale heat conduction in gold films exposed to ultrashort-pulsed lasers}, International Journal of Heat and Mass Transfer 190 (2022) 122791.

\bibitem{niaki2021physics}
S.~A. Niaki, E.~Haghighat, T.~Campbell, A.~Poursartip, R.~Vaziri, {Physics-informed neural network for modelling the thermochemical curing process of composite-tool systems during manufacture}, Computer Methods in Applied Mechanics and Engineering 384 (2021) 113959.

\bibitem{patel2022thermodynamically}
R.~G. Patel, I.~Manickam, N.~A. Trask, M.~A. Wood, M.~Lee, I.~Tomas, E.~C. Cyr, {Thermodynamically consistent physics-informed neural networks for hyperbolic systems}, Journal of Computational Physics 449 (2022) 110754.

\bibitem{mathews2021uncovering}
A.~Mathews, M.~Francisquez, J.~W. Hughes, D.~R. Hatch, B.~Zhu, B.~N. Rogers, {Uncovering turbulent plasma dynamics via deep learning from partial observations}, Physical Review E 104~(2) (2021) 025205.

\bibitem{stielow2021reconstruction}
T.~Stielow, S.~Scheel, {Reconstruction of nanoscale particles from single-shot wide-angle free-electron-laser diffraction patterns with physics-informed neural networks}, Physical Review E 103~(5) (2021) 053312.

\bibitem{kovacs2022magnetostatics}
A.~Kovacs, L.~Exl, A.~Kornell, J.~Fischbacher, M.~Hovorka, M.~Gusenbauer, L.~Breth, H.~Oezelt, D.~Praetorius, D.~Suess, et~al., {Magnetostatics and micromagnetics with physics informed neural networks}, Journal of Magnetism and Magnetic Materials 548 (2022) 168951.

\bibitem{chen2022predicting}
H.~Chen, E.~Katelhon, R.~G. Compton, {Predicting voltammetry using physics-informed neural networks}, The Journal of Physical Chemistry Letters 13~(2) (2022) 536--543.

\bibitem{shah2022physics}
K.~Shah, P.~Stiller, N.~Hoffmann, A.~Cangi, {Physics-Informed Neural Networks as Solvers for the Time-Dependent Schr{\"o}dinger Equation}, arXiv preprint arXiv:2210.12522 (2022).

\bibitem{wang2021data}
L.~Wang, Z.~Yan, {Data-driven rogue waves and parameter discovery in the defocusing nonlinear Schr{\"o}}dinger equation with a potential using the {PINN} deep learning, Physics Letters A 404 (2021) 127408.

\bibitem{pu2021solving}
J.~Pu, J.~Li, Y.~Chen, {Solving localized wave solutions of the derivative nonlinear Schr{\"o}}dinger equation using an improved {PINN} method, Nonlinear Dynamics 105~(2) (2021) 1723--1739.

\bibitem{li2021physicsphonon}
R.~Li, E.~Lee, T.~Luo, {Physics-informed neural networks for solving multiscale mode-resolved phonon Boltzmann transport equation}, Materials Today Physics 19 (2021) 100429.

\bibitem{singh2024tracking}
G.~Singh, V.~Kumar, A.~B. Buduru, S.~K. Biswas, {Tracking an untracked space debris after an inelastic collision using physics informed neural network}, Scientific Reports 14~(1) (2024) 3350.

\bibitem{chen2020physics}
Y.~Chen, L.~Lu, G.~E. Karniadakis, L.~Dal~Negro, {Physics-informed neural networks for inverse problems in nano-optics and metamaterials}, Optics Express 28~(8) (2020) 11618--11633.

\bibitem{jiang2022physics}
X.~Jiang, D.~Wang, Q.~Fan, M.~Zhang, C.~Lu, A.~P.~T. Lau, {Physics-Informed Neural Network for Nonlinear Dynamics in Fiber Optics}, Laser \& Photonics Reviews 16~(9) (2022) 2100483.

\bibitem{wu2021predicting}
G.-Z. Wu, Y.~Fang, Y.-Y. Wang, G.-C. Wu, C.-Q. Dai, {Predicting the dynamic process and model parameters of the vector optical solitons in birefringent fibers via the modified {PINN}}, Chaos, Solitons \& Fractals 152 (2021) 111393.

\bibitem{hager2020physics}
C.~H{\"a}ger, H.~D. Pfister, {Physics-based deep learning for fiber-optic communication systems}, IEEE Journal on Selected Areas in Communications 39~(1) (2020) 280--294.

\bibitem{camporeale2022data}
E.~Camporeale, G.~J. Wilkie, A.~Y. Drozdov, J.~Bortnik, {Data-driven discovery of Fokker-Planck equation for the Earth's radiation belts electrons using Physics-Informed Neural Networks} (2022).

\bibitem{schiassi2022orbit}
E.~Schiassi, A.~D’Ambrosio, K.~Drozd, F.~Curti, R.~Furfaro, {Physics-informed neural networks for optimal planar orbit transfers}, Journal of Spacecraft and Rockets 59~(3) (2022) 834--849.

\bibitem{furfaro2022physics}
R.~Furfaro, A.~D'Ambrosio, E.~Schiassi, A.~Scorsoglio, {Physics-Informed Neural Networks for Closed-Loop Guidance and Control in Aerospace Systems}, in: {AIAA SCITECH 2022 Forum}, 2022, p. 0361.

\bibitem{d2021physics}
A.~D’Ambrosio, E.~Schiassi, F.~Curti, R.~Furfaro, {Physics-Informed Neural Networks Applied to a Series of Constrained Space Guidance Problems}, in: {31st AAS/AIAA Space Flight Mechanics Meeting}, 2021.

\bibitem{martin2022reinforcement}
J.~Martin, H.~Schaub, {Reinforcement learning and orbit-discovery enhanced by small-body physics-informed neural network gravity models}, in: {AIAA SCITECH 2022 Forum}, 2022, p. 2272.

\bibitem{martin2022physics}
J.~Martin, H.~Schaub, {Physics-informed neural networks for gravity field modeling of the Earth and Moon}, Celestial Mechanics and Dynamical Astronomy 134~(2) (2022) 1--28.

\bibitem{martin2022periodic}
J.~R. Martin, H.~Schaub, {Periodic Orbit Discovery Enhanced by Physics-Informed Neural Networks}, in: {2022 Astrodynamics Specialist Conference, Charlotte, North Carolina}, 2022, pp. 7--11.

\bibitem{mishra2021physics}
S.~Mishra, R.~Molinaro, {Physics informed neural networks for simulating radiative transfer}, Journal of Quantitative Spectroscopy and Radiative Transfer 270 (2021) 107705.

\bibitem{wu2023application}
Z.~Wu, H.~Wang, C.~He, B.~Zhang, T.~Xu, Q.~Chen, {The application of physics-informed machine learning in multiphysics modeling in chemical engineering}, Industrial \& Engineering Chemistry Research 62~(44) (2023) 18178--18204.

\bibitem{zhu2022review}
L.-T. Zhu, X.-Z. Chen, B.~Ouyang, W.-C. Yan, H.~Lei, Z.~Chen, Z.-H. Luo, {Review of machine learning for hydrodynamics, transport, and reactions in multiphase flows and reactors}, Industrial \& Engineering Chemistry Research 61~(28) (2022) 9901--9949.

\bibitem{batuwatta2023novel}
C.~P. Batuwatta-Gamage, C.~Rathnayaka, H.~C. Karunasena, H.~Jeong, A.~Karim, Y.~T. Gu, {A novel physics-informed neural networks approach ({PINN-MT}}) to solve mass transfer in plant cells during drying, Biosystems Engineering 230 (2023) 219--241.

\bibitem{xuan2023physics}
W.~Xuan, H.~Lou, S.~Fu, Z.~Zhang, N.~Ding, {Physics-informed deep learning method for the refrigerant filling mass flow metering}, Flow Measurement and Instrumentation 93 (2023) 102418.

\bibitem{ji2021stiff}
W.~Ji, W.~Qiu, Z.~Shi, S.~Pan, S.~Deng, {Stiff-pinn: Physics-informed neural network for stiff chemical kinetics}, The Journal of Physical Chemistry A 125~(36) (2021) 8098--8106.

\bibitem{de2022physics}
M.~De~Florio, E.~Schiassi, R.~Furfaro, {Physics-informed neural networks and functional interpolation for stiff chemical kinetics}, Chaos: An Interdisciplinary Journal of Nonlinear Science 32~(6) (2022) 063107.

\bibitem{weng2022multiscale}
Y.~Weng, D.~Zhou, {Multiscale physics-informed neural networks for stiff chemical kinetics}, The Journal of Physical Chemistry A 126~(45) (2022) 8534--8543.

\bibitem{ngo2021solution}
S.~I. Ngo, Y.-I. Lim, {Solution and parameter identification of a fixed-bed reactor model for catalytic CO2 methanation using physics-informed neural networks}, Catalysts 11~(11) (2021) 1304.

\bibitem{cohen2024data}
B.~Cohen, B.~Beykal, G.~M. Bollas, {Data-driven Discovery of Reaction Kinetic Models in Dynamic Plug Flow Reactors using Symbolic Regression}, in: {Computer Aided Chemical Engineering}, Vol.~53, Elsevier, 2024, pp. 2947--2952.

\bibitem{bibeau2024physics}
V.~Bibeau, D.~C. Boffito, B.~Blais, {Physics-informed Neural Network to predict kinetics of biodiesel production in microwave reactors}, Chemical Engineering and Processing-Process Intensification 196 (2024) 109652.

\bibitem{hou2023pinn}
Q.~Hou, H.~Du, Z.~Sun, J.~Wang, X.~Wang, J.~Wei, {{PINN-CDR}}: A neural network-based simulation tool for convection-diffusion-reaction systems, International Journal of Intelligent Systems 2023~(1) (2023) 2973249.

\bibitem{sun2023physics}
Z.~Sun, H.~Du, C.~Miao, Q.~Hou, {A physics-informed neural network based simulation tool for reacting flow with multicomponent reactants}, Advances in Engineering Software 185 (2023) 103525.

\bibitem{choi2022physics}
S.~Choi, I.~Jung, H.~Kim, J.~Na, J.~M. Lee, {Physics-informed deep learning for data-driven solutions of computational fluid dynamics}, Korean Journal of Chemical Engineering 39~(3) (2022) 515--528.

\bibitem{patel2023optimal}
R.~Patel, S.~Bhartiya, R.~Gudi, {Optimal temperature trajectory for tubular reactor using physics informed neural networks}, Journal of Process Control 128 (2023) 103003.

\bibitem{ngo2022forward}
S.~I. Ngo, Y.-I. Lim, {Forward Physics-Informed Neural Networks Suitable for Multiple Operating Conditions of Catalytic CO2 Methanation Isothermal Fixed-Bed}, IFAC-PapersOnLine 55~(7) (2022) 429--434.

\bibitem{elhareef2023physics}
M.~H. Elhareef, Z.~Wu, {Physics-informed neural network method and application to nuclear reactor calculations: A pilot study}, Nuclear Science and Engineering 197~(4) (2023) 601--622.

\bibitem{schiassi2022physics}
E.~Schiassi, M.~De~Florio, B.~D. Ganapol, P.~Picca, R.~Furfaro, {Physics-informed neural networks for the point kinetics equations for nuclear reactor dynamics}, Annals of Nuclear Energy 167 (2022) 108833.

\bibitem{liu2023multi}
Y.-T. Liu, C.-Y. Wu, T.~Chen, Y.~Yao, {Multi-fidelity surrogate modeling for chemical processes with physics-informed neural networks}, in: {Computer Aided Chemical Engineering}, Vol.~52, Elsevier, 2023, pp. 57--63.

\bibitem{antonello2023physics}
F.~Antonello, J.~Buongiorno, E.~Zio, {Physics informed neural networks for surrogate modeling of accidental scenarios in nuclear power plants}, Nuclear Engineering and Technology 55~(9) (2023) 3409--3416.

\bibitem{liu2023surrogate}
K.~Liu, K.~Luo, Y.~Cheng, A.~Liu, H.~Li, J.~Fan, S.~Balachandar, {Surrogate modeling of parameterized multi-dimensional premixed combustion with physics-informed neural networks for rapid exploration of design space}, Combustion and Flame 258 (2023) 113094.

\bibitem{zou2024parameter}
T.~Zou, T.~Yajima, Y.~Kawajiri, {A parameter estimation method for chromatographic separation process based on physics-informed neural network}, Journal of Chromatography A (2024) 465077.

\bibitem{soderstrom2022physics}
P.~S{\"o}derstr{\"o}m, {Physics-Informed Neural Networks for Liquid Chromatography} (2022).

\bibitem{tang2023physics}
S.-Y. Tang, Y.-H. Yuan, Y.-C. Chen, S.-J. Yao, Y.~Wang, D.-Q. Lin, {Physics-informed neural networks to solve lumped kinetic model for chromatography process}, Journal of Chromatography A 1708 (2023) 464346.

\bibitem{li2024unit}
H.~Li, D.~Spelman, J.~Sansalone, {Unit Operation and Process Modeling with Physics-Informed Machine Learning}, Journal of Environmental Engineering 150~(4) (2024) 04024002.

\bibitem{bai2022application}
Y.~Bai, T.~Chaolu, S.~Bilige, {The application of improved physics-informed neural network ({IPINN}}) method in finance, Nonlinear Dynamics 107~(4) (2022) 3655--3667.

\bibitem{fang2019deep}
Z.~Fang, J.~Zhan, {Deep physical informed neural networks for metamaterial design}, IEEE Access 8 (2019) 24506--24513.

\bibitem{islam2021extraction}
M.~Islam, M.~S.~H. Thakur, S.~Mojumder, M.~N. Hasan, {Extraction of material properties through multi-fidelity deep learning from molecular dynamics simulation}, Computational Materials Science 188 (2021) 110187.

\bibitem{misyris2020physics}
G.~S. Misyris, A.~Venzke, S.~Chatzivasileiadis, {Physics-informed neural networks for power systems}, in: {2020 IEEE Power \& Energy Society General Meeting (PESGM)}, IEEE, 2020, pp. 1--5.

\bibitem{park2019physics}
J.~Park, J.~Park, {Physics-induced graph neural network: An application to wind-farm power estimation}, Energy 187 (2019) 115883.

\bibitem{dabrowski2022bayesian}
J.~J. Dabrowski, D.~E. Pagendam, J.~Hilton, C.~Sanderson, D.~MacKinlay, C.~Huston, A.~Bolt, P.~Kuhnert, {Bayesian Physics Informed Neural Networks for Data Assimilation and Spatio-Temporal Modelling of Wildfires}, arXiv preprint arXiv:2212.00970 (2022).

\bibitem{gao2021phygeonet}
H.~Gao, L.~Sun, J.-X. Wang, {PhyGeoNet: Physics-informed geometry-adaptive convolutional neural networks for solving parameterized steady-state PDEs on irregular domain}, Journal of Computational Physics 428 (2021) 110079.

\bibitem{liu2022discontinuity}
L.~Liu, S.~Liu, H.~Yong, F.~Xiong, T.~Yu, {Discontinuity Computing with Physics-Informed Neural Network}, arXiv preprint arXiv:2206.03864 (2022).

\bibitem{tseng2023cusp}
Y.-H. Tseng, T.-S. Lin, W.-F. Hu, M.-C. Lai, A cusp-capturing pinn for elliptic interface problems, Journal of Computational Physics 491 (2023) 112359.

\bibitem{kadeethum2020physics}
T.~Kadeethum, T.~M. J{\o}rgensen, H.~M. Nick, {Physics-informed neural networks for solving nonlinear diffusivity and Biot’s equations}, PLOS One 15~(5) (2020) e0232683.

\bibitem{abdar2021review}
M.~Abdar, F.~Pourpanah, S.~Hussain, D.~Rezazadegan, L.~Liu, M.~Ghavamzadeh, P.~Fieguth, X.~Cao, A.~Khosravi, U.~R. Acharya, et~al., {A review of uncertainty quantification in deep learning: Techniques, applications and challenges}, Information Fusion 76 (2021) 243--297.

\bibitem{psaros2023uncertainty}
A.~F. Psaros, X.~Meng, Z.~Zou, L.~Guo, G.~E. Karniadakis, {Uncertainty quantification in scientific machine learning: Methods, metrics, and comparisons}, Journal of Computational Physics 477 (2023) 111902.

\bibitem{yang2021b}
L.~Yang, X.~Meng, G.~E. Karniadakis, {{B-PINNs: Bayesian physics-informed neural networks for forward and inverse PDE problems with noisy data}}, Journal of Computational Physics 425 (2021) 109913.

\bibitem{zou2024neuraluq}
Z.~Zou, X.~Meng, A.~F. Psaros, G.~E. Karniadakis, {{NeuralUQ}}: A comprehensive library for uncertainty quantification in neural differential equations and operators, SIAM Review 66~(1) (2024) 161--190.

\bibitem{meng2022learning}
X.~Meng, L.~Yang, Z.~Mao, J.~del {\'A}guila~Ferrandis, G.~E. Karniadakis, {Learning functional priors and posteriors from data and physics}, Journal of Computational Physics 457 (2022) 111073.

\bibitem{zou2024correcting}
Z.~Zou, X.~Meng, G.~E. Karniadakis, {Correcting model misspecification in physics-informed neural networks ({PINNs}}), Journal of Computational Physics 505 (2024) 112918.

\bibitem{gal2016dropout}
Y.~Gal, Z.~Ghahramani, {Dropout as a Bayesian approximation: Representing model uncertainty in deep learning}, in: {International Conference on Machine Learning}, PMLR, 2016, pp. 1050--1059.

\bibitem{zhang2019quantifying}
D.~Zhang, L.~Lu, L.~Guo, G.~E. Karniadakis, {Quantifying total uncertainty in physics-informed neural networks for solving forward and inverse stochastic problems}, Journal of Computational Physics 397 (2019) 108850.

\bibitem{gao2022wasserstein}
Y.~Gao, M.~K. Ng, {Wasserstein generative adversarial uncertainty quantification in physics-informed neural networks}, Journal of Computational Physics 463 (2022) 111270.

\bibitem{daw2021pid}
A.~Daw, M.~Maruf, A.~Karpatne, {{PID-GAN: A GAN Framework based on a Physics-informed Discriminator for Uncertainty Quantification with Physics}}, in: {Proceedings of the 27th ACM SIGKDD Conference on Knowledge Discovery \& Data Mining}, 2021, pp. 237--247.

\bibitem{jiang2023practical}
X.~Jiang, X.~Wang, Z.~Wen, E.~Li, H.~Wang, {Practical uncertainty quantification for space-dependent inverse heat conduction problem via ensemble physics-informed neural networks}, International Communications in Heat and Mass Transfer 147 (2023) 106940.

\bibitem{soibam2024inverse}
J.~Soibam, I.~Aslanidou, K.~Kyprianidis, R.~B. Fdhila, {Inverse flow prediction using ensemble {PINNs}} and uncertainty quantification, International Journal of Heat and Mass Transfer 226 (2024) 125480.

\bibitem{yang2022multi}
M.~Yang, J.~T. Foster, {Multi-output physics-informed neural networks for forward and inverse PDE problems with uncertainties}, Computer Methods in Applied Mechanics and Engineering 402 (2022) 115041.

\bibitem{de2024quantification}
M.~De~Florio, Z.~Zou, D.~E. Schiavazzi, G.~E. Karniadakis, {Quantification of total uncertainty in the physics-informed reconstruction of {CVSim}}-6 physiology, arXiv preprint arXiv:2408.07201 (2024).

\bibitem{lin2022multi}
G.~Lin, Y.~Wang, Z.~Zhang, {{Multi-variance replica exchange SGMCMC for inverse and forward problems via Bayesian PINN}}, Journal of Computational Physics 460 (2022) 111173.

\bibitem{lutjens2021pce}
B.~L{\"u}tjens, C.~H. Crawford, M.~Veillette, D.~Newman, {{PCE-PINNs}}: Physics-informed neural networks for uncertainty propagation in ocean modeling, arXiv preprint arXiv:2105.02939 (2021).

\bibitem{zou2024leveraging}
Z.~Zou, T.~Meng, P.~Chen, J.~Darbon, G.~E. Karniadakis, {{Leveraging Viscous {Hamilton–Jacobi}} {PDEs} for Uncertainty Quantification in Scientific Machine Learning}, SIAM/ASA Journal on Uncertainty Quantification 12~(4) (2024) 1165--1191.

\bibitem{zou2023hydra}
Z.~Zou, G.~E. Karniadakis, {{L-HYDRA:}} multi-head physics-informed neural networks, arXiv preprint arXiv:2301.02152 (2023).

\bibitem{rezende2015variational}
D.~Rezende, S.~Mohamed, {Variational inference with normalizing flows}, in: {International Conference on Machine Learning}, PMLR, 2015, pp. 1530--1538.

\bibitem{meng2023variational}
X.~Meng, {Variational inference in neural functional prior using normalizing flows: application to differential equation and operator learning problems}, Applied Mathematics and Mechanics 44~(7) (2023) 1111--1124.

\bibitem{yin2023generative}
M.~Yin, Z.~Zou, E.~Zhang, C.~Cavinato, J.~D. Humphrey, G.~E. Karniadakis, {A generative modeling framework for inferring families of biomechanical constitutive laws in data-sparse regimes}, Journal of the Mechanics and Physics of Solids 181 (2023) 105424.

\bibitem{zou2023uncertainty}
Z.~Zou, X.~Meng, G.~E. Karniadakis, {Uncertainty quantification for noisy inputs-outputs in physics-informed neural networks and neural operators}, arXiv preprint arXiv:2311.11262 (2023).

\bibitem{zhang2024discovering}
Z.~Zhang, Z.~Zou, E.~Kuhl, G.~E. Karniadakis, {{Discovering a reaction--diffusion model for Alzheimer’s disease by combining {PINNs}} with symbolic regression}, Computer Methods in Applied Mechanics and Engineering 419 (2024) 116647.

\bibitem{chen2021physics}
Z.~Chen, Y.~Liu, H.~Sun, {Physics-informed learning of governing equations from scarce data}, Nature Communications 12~(1) (2021) 6136.

\bibitem{chen2021generalized}
Z.~Chen, D.~Xiu, {On generalized residual network for deep learning of unknown dynamical systems}, Journal of Computational Physics 438 (2021) 110362.

\bibitem{ebers2024discrepancy}
M.~R. Ebers, K.~M. Steele, J.~N. Kutz, {Discrepancy Modeling Framework: Learning Missing Physics, Modeling Systematic Residuals, and Disambiguating between Deterministic and Random Effects}, SIAM Journal on Applied Dynamical Systems 23~(1) (2024) 440--469.

\bibitem{meng2024hj}
T.~Meng, Z.~Zou, J.~Darbon, G.~E. Karniadakis, {{HJ-sampler: A Bayesian sampler for inverse problems of a stochastic process by leveraging Hamilton-Jacobi PDEs and score-based generative models}}, arXiv preprint arXiv:2409.09614 (2024).

\bibitem{linka2022bayesian}
K.~Linka, A.~Sch{\"a}fer, X.~Meng, Z.~Zou, G.~E. Karniadakis, E.~Kuhl, {Bayesian Physics Informed Neural Networks for real-world nonlinear dynamical systems}, Computer Methods in Applied Mechanics and Engineering 402 (2022) 115346.

\bibitem{oszkinat2022uncertainty}
C.~Oszkinat, S.~E. Luczak, I.~Rosen, {Uncertainty quantification in estimating blood alcohol concentration from transdermal alcohol level with physics-informed neural networks}, IEEE Transactions on Neural Networks and Learning Systems 34~(10) (2022) 8094--8101.

\bibitem{mo2022trafficflowgan}
Z.~Mo, Y.~Fu, D.~Xu, X.~Di, {{TrafficFlowGAN: Physics-informed flow based generative adversarial network for uncertainty quantification}}, in: {Joint European Conference on Machine Learning and Knowledge Discovery in Databases}, Springer, 2022, pp. 323--339.

\bibitem{shin2020convergence}
Y.~Shin, J.~Darbon, G.~E. Karniadakis, {On the convergence of physics informed neural networks for linear second-order elliptic and parabolic type PDEs}, arXiv preprint arXiv:2004.01806 (2020).

\bibitem{mishra2023estimates}
S.~Mishra, R.~Molinaro, {Estimates on the generalization error of physics-informed neural networks for approximating PDEs}, IMA Journal of Numerical Analysis 43~(1) (2023) 1--43.

\bibitem{wu2022convergence}
S.~Wu, A.~Zhu, Y.~Tang, B.~Lu, {Convergence of physics-informed neural networks applied to linear second-order elliptic interface problems}, arXiv preprint arXiv:2203.03407 (2022).

\bibitem{qian2023error}
Y.~Qian, Y.~Zhang, Y.~Huang, S.~Dong, {Error analysis of physics-informed neural networks for approximating dynamic PDEs of second order in time}, arXiv preprint arXiv:2303.12245 (2023).

\bibitem{hu2023higher}
R.~Hu, Q.~Lin, A.~Raydan, S.~Tang, {Higher-order error estimates for physics-informed neural networks approximating the primitive equations}, Partial Differential Equations and Applications 4~(4) (2023) 34.

\bibitem{shin2023error}
Y.~Shin, Z.~Zhang, G.~E. Karniadakis, {Error estimates of residual minimization using neural networks for linear PDEs}, Journal of Machine Learning for Modeling and Computing 4~(4) (2023).

\bibitem{mishra2022estimates}
S.~Mishra, R.~Molinaro, {Estimates on the generalization error of physics-informed neural networks for approximating a class of inverse problems for PDEs}, IMA Journal of Numerical Analysis 42~(2) (2022) 981--1022.

\bibitem{muller2022error}
J.~M{\"u}ller, M.~Zeinhofer, {Error estimates for the deep Ritz method with boundary penalty}, in: {Mathematical and Scientific Machine Learning}, PMLR, 2022, pp. 215--230.

\bibitem{biswas2022error}
A.~Biswas, J.~Tian, S.~Ulusoy, {Error estimates for deep learning methods in fluid dynamics}, Numerische Mathematik 151~(3) (2022) 753--777.

\bibitem{doumeche2023convergence}
N.~Doum{\`e}che, G.~Biau, C.~Boyer, {Convergence and error analysis of {PINNs}}, arXiv preprint arXiv:2305.01240 (2023).

\bibitem{jacot2018neural}
A.~Jacot, F.~Gabriel, C.~Hongler, {Neural Tangent Kernel: Convergence and generalization in neural networks}, Advances in Neural Information Processing Systems 31 (2018).

\bibitem{wang2022when}
S.~Wang, X.~Yu, P.~Perdikaris, {When and why {PINNs}} fail to train: A neural tangent kernel perspective, Journal of Computational Physics 449 (2022) 110768.

\bibitem{tishby2000information}
N.~Tishby, F.~C. Pereira, W.~Bialek, {The information bottleneck method}, arXiv preprint physics/0004057 (2000).

\bibitem{tishby2015deep}
N.~Tishby, N.~Zaslavsky, {Deep learning and the information bottleneck principle}, in: {2015 IEEE Information Theory Workshop (ITW)}, IEEE, 2015, pp. 1--5.

\bibitem{shwartz2017opening}
R.~Shwartz-Ziv, N.~Tishby, {{Opening the black box of deep neural networks via information}}, arXiv preprint arXiv:1703.00810 (2017).

\bibitem{goldfeld2020information}
Z.~Goldfeld, Y.~Polyanskiy, {The information bottleneck problem and its applications in machine learning}, IEEE Journal on Selected Areas in Information Theory 1~(1) (2020) 19--38.

\bibitem{shwartz2022information}
R.~Shwartz-Ziv, {Information flow in deep neural networks}, arXiv preprint arXiv:2202.06749 (2022).

\bibitem{zubov2021neuralpde}
K.~Zubov, Z.~McCarthy, Y.~Ma, F.~Calisto, V.~Pagliarino, S.~Azeglio, L.~Bottero, E.~Luj{\'a}n, V.~Sulzer, A.~Bharambe, et~al., {{NeuralPDE: Automating physics-informed neural networks ({PINNs}}) with error approximations}, arXiv preprint arXiv:2107.09443 (2021).

\bibitem{modulus}
{Modulus Contributors}, \href{https://github.com/NVIDIA/modulus}{{{NVIDIA Modulus: An open-source framework for physics-based deep learning in science and engineering}}} (2023).
\newline\urlprefix\url{https://github.com/NVIDIA/modulus}

\bibitem{chen2020neurodiffeq}
F.~Chen, D.~Sondak, P.~Protopapas, M.~Mattheakis, S.~Liu, D.~Agarwal, M.~Di~Giovanni, {NeuroDiffEq: A Python package for solving differential equations with neural networks}, Journal of Open Source Software 5~(46) (2020) 1931.

\bibitem{TorchPhysics}
D.~N. Tanyu, J.~Ning, T.~Freudenberg, N.~Heilenkötter, A.~Rademacher, U.~Iben, P.~Maass, {Deep learning methods for partial differential equations and related parameter identification problems}, Inverse Problems 39~(10) (2023) 103001.

\bibitem{haghighat2021sciann}
E.~Haghighat, R.~Juanes, {SciANN: A Keras/Tensorflow wrapper for scientific computations and physics-informed deep learning using artificial neural networks}, Computer Methods in Applied Mechanics and Engineering 373 (2021) 113552.

\bibitem{koryagin2019pydens}
A.~Koryagin, R.~Khudorozkov, S.~Tsimfer, {{PyDEns}: A Python framework for solving differential equations with neural networks}, arXiv preprint arXiv:1909.11544 (2019).

\bibitem{xu2020adcme}
K.~Xu, E.~Darve, {{ADCME: Learning spatially-varying physical fields using deep neural networks}}, arXiv preprint arXiv:2011.11955 (2020).

\bibitem{mcclenny2021tensordiffeq}
L.~D. McClenny, M.~A. Haile, U.~M. Braga-Neto, {{TensorDiffEq: Scalable Multi-GPU Forward and Inverse Solvers for Physics Informed Neural Networks}}, arXiv preprint arXiv:2103.16034 (2021).

\bibitem{araz2021elvet}
J.~Y. Araz, J.~C. Criado, M.~Spannowsky, {Elvet--a neural network-based differential equation and variational problem solver}, arXiv preprint arXiv:2103.14575 (2021).

\bibitem{peng2021idrlnet}
W.~Peng, J.~Zhang, W.~Zhou, X.~Zhao, W.~Yao, X.~Chen, {{IDRLnet: A physics-informed neural network library}}, arXiv preprint arXiv:2107.04320 (2021).

\bibitem{pedro2019solving}
J.~B. Pedro, J.~Maro{\~n}as, R.~Paredes, {Solving partial differential equations with neural networks}, arXiv preprint arXiv:1912.04737 (2019).

\bibitem{mcclenny2020self}
L.~McClenny, U.~Braga-Neto, {Self-adaptive physics-informed neural networks using a soft attention mechanism}, arXiv preprint arXiv:2009.04544 (2020).

\bibitem{frostig2018compiling}
R.~Frostig, M.~J. Johnson, C.~Leary, {Compiling machine learning programs via high-level tracing}, Systems for Machine Learning 4~(9) (2018).

\bibitem{cho2023separablephysicsinformedneuralnetworks}
J.~Cho, S.~Nam, H.~Yang, S.-B. Yun, Y.~Hong, E.~Park, {Separable Physics-Informed Neural Networks}, arXiv preprint arXiv:2306.15969 (2023).

\bibitem{wang2023multistageneuralnetworksfunction}
Y.~Wang, C.-Y. Lai, {Multi-stage Neural Networks: Function Approximator of Machine Precision}, arXiv preprint arXiv:2307.08934 (2023).

\bibitem{lu2019deeponet}
L.~Lu, P.~Jin, G.~E. Karniadakis, {DeepOnet: Learning nonlinear operators for identifying differential equations based on the universal approximation theorem of operators}, arXiv preprint arXiv:1910.03193 (2019).

\bibitem{raissi2018hiddenfluidmechanicsnavierstokes}
M.~Raissi, A.~Yazdani, G.~E. Karniadakis, {Hidden Fluid Mechanics: A Navier-Stokes Informed Deep Learning Framework for Assimilating Flow Visualization Data}, arXiv preprint arXiv:1808.04327 (2018).

\bibitem{kharazmi2019variationalphysicsinformedneuralnetworks}
E.~Kharazmi, Z.~Zhang, G.~E. Karniadakis, {Variational Physics-Informed Neural Networks For Solving Partial Differential Equations}, arXiv preprint arXiv:1912.00873 (2019).

\bibitem{wang2020pinnsfailtrainneural}
S.~Wang, X.~Yu, P.~Perdikaris, {When and why {PINNs} fail to train: A neural tangent kernel perspective}, arXiv preprint arXiv:2007.14527 (2020).

\bibitem{Wang2020_Fourier_nets}
S.~Wang, H.~Wang, P.~Perdikaris, {On the eigenvector bias of Fourier feature networks: From regression to solving multi-scale {PDEs} with physics-informed neural networks}, arXiv preprint arXiv:2012.10047 (2020).

\end{thebibliography}

\newpage
\appendix
\section{Chronological overview of key advancements in PIML}
\begin{table}[H]
    \centering
    \scriptsize
    \begin{tabular}{|p{8cm}|p{4.5cm}|p{0.5cm}|}
    \hline
        Publication & Contribution  & Year \\\hline
         Physics Informed Deep Learning (Part I): Data-driven solutions of nonlinear Partial Differential Equations~\citep{raissi2017physicsI}& PINN framework for solving PDEs & 2017 \\\hline
         Physics Informed Deep Learning (Part II): Data-driven discovery of nonlinear Partial Differential Equations~\citep{raissi2017physicsII}& PINN framework for PDE discovery & 2017 \\\hline
         Hidden Fluid Mechanics: A Navier-Stokes informed deep learning framework for assimilating flow visualization Data~\citep{raissi2018hiddenfluidmechanicsnavierstokes}& Uncovers hidden fields, uses weight normalization & 2018 \\\hline
        fPINNs: Fractional Physics-Informed Neural Networks~ \citep{pang2019fpinns}& Extends PINNs to fractional PDEs & 2018 \\\hline
        Quantifying total uncertainty in physics-informed neural networks for solving forward and inverse stochastic problems \citep{zhang2019quantifying}& Introduces UQ for PINNs & 2018 \\\hline
         Physics-informed neural networks: A deep learning framework for solving forward and inverse problems involving nonlinear partial differential equations~\citep{raissi2019physics}& Journal paper formalizing PINNs & 2019 \\\hline
        Multi-scale Deep Neural Networks for Solving High-Dimensional PDEs~\citep{cai2019multi}& Introduces multi-scale feature expansions & 2019 \\\hline
         Variational Physics-Informed Neural Networks For Solving Partial Differential Equations~\citep{kharazmi2019variationalphysicsinformedneuralnetworks}& Extends PINNs using variational approaches & 2019 \\\hline
         DeepOnet: Learning nonlinear operators for identifying differential equations based on the universal approximation theorem of operators \citep{lu2019deeponet}& Introduces DeepONets & 2019 \\\hline
         DeepXDE: A deep learning library for solving differential equations \citep{lu2021deepxde}& First PIML library and introduction to resampling & 2019 \\\hline
         On the convergence of physics-informed neural networks for linear second-order elliptic and parabolic type PDEs~\citep{shin2020convergence}& Convergence and error bounds for PINNs & 2020 \\\hline
         When and why PINNs fail to train: A neural tangent kernel perspective~\citep{wang2020pinnsfailtrainneural}& NTK analysis for PINN  & 2020 \\\hline
        Extended physics-informed neural networks (XPINNs): A generalized space-time domain decomposition-based deep learning framework for nonlinear partial differential equations~\citep{jagtap2020extended}& Domain decomposition for PINNs & 2020 \\\hline
        Locally adaptive activation functions with slope recovery for deep and physics-informed neural networks \citep{jagtap2020locally}& Introduces adaptive activation functions & 2020 \\\hline
         B-PINNs: Bayesian Physics-Informed Neural Networks for forward and inverse PDE problems with noisy Data \citep{yang2021b}& Bayesian PINNs for noisy data & 2020 \\\hline
         Self-adaptive physics-informed neural networks using a soft attention mechanism~\citep{mcclenny2020self}& Introduces local weights an attention mechanisms & 2020 \\\hline
         On the eigenvector bias of Fourier feature networks: From regression to solving multi-scale PDEs with physics-informed neural networks~\citep{Wang2020_Fourier_nets}& Fourier features for multi-scale PDEs & 2020 \\\hline
         Separable Physics-Informed Neural Networks~\citep{cho2023separablephysicsinformedneuralnetworks}& Speeds up training up to 100 times & 2023 \\\hline
         Artificial to Spiking Neural Networks Conversion for Scientific Machine Learning~\citep{zhang2023artificial}& Extends PIML to Spiking NNs & 2023 \\\hline
         Residual-based attention and connection to information bottleneck theory in PINNs~\citep{anagnostopoulos2023residual}& Connection to information bottleneck theory & 2023 \\\hline
         Stacked networks improve physics-informed training: applications to neural networks and deep operator networks~\citep{howard2023stacked}& Stacked training improves performance & 2023 \\\hline
         Multi-stage Neural Networks: Function approximator of machine precision~\citep{wang2023multistageneuralnetworksfunction}& Multi-stage training for better performance & 2023 \\\hline
         Tackling the curse of dimensionality with Physics-Informed Neural Networks \citep{hu2024tackling}& Solves PDEs with up to 100,000 dimensions & 2023 \\\hline
         KAN: Kolmogorov-Arnold networks~\citep{liu2024kan}& Extension to KANs & 2024 \\\hline
 \end{tabular}
    \caption{Evolution of the algorithmic variants}
    \label{tab:Variants1}
\end{table}

\end{document}